\documentclass[journal]{IEEEtran}
\ifCLASSINFOpdf
\else
\fi

\usepackage{lineno}
\usepackage{amsmath,amssymb,amsfonts}
\usepackage{amsthm}
\usepackage{enumitem}
\usepackage{algorithm}
\usepackage{algorithmic}
\usepackage{textcomp}
\usepackage{multirow}
\usepackage{multicol}
\usepackage{longtable}
\usepackage{lipsum}
\usepackage{float}
\usepackage{dblfloatfix}
\usepackage{mathtools}
\usepackage{xcolor}
\usepackage{array}
\usepackage{graphicx}
\usepackage{bbold}

\newtheorem{lemma}{Lemma}

\newtheorem{theorem}{Theorem}

\newtheorem{definition}{Definition}

\newtheorem*{remark}{Remark}

\newcommand{\R}{\mathbb{R}}

\ifCLASSOPTIONcompsoc
\usepackage[caption=false,font=normalsize,labelfon
t=sf,textfont=sf]{subfig}
\else
\usepackage[caption=false,font=footnotesize]{subfig}
\fi

\newcommand{\PreserveBackslash}[1]{\let\temp=\\#1\let\\=\temp}
\newcolumntype{C}[1]{>{\PreserveBackslash\centering}p{#1}}
\newcolumntype{R}[1]{>{\PreserveBackslash\raggedleft}p{#1}}
\newcolumntype{L}[1]{>{\PreserveBackslash\raggedright}p{#1}}

\usepackage{scalerel}
\usepackage{tikz}
\usetikzlibrary{svg.path}

\definecolor{orcidlogocol}{HTML}{A6CE39}
\tikzset{
	orcidlogo/.pic={
		\fill[orcidlogocol] svg{M256,128c0,70.7-57.3,128-128,128C57.3,256,0,198.7,0,128C0,57.3,57.3,0,128,0C198.7,0,256,57.3,256,128z};
		\fill[white] svg{M86.3,186.2H70.9V79.1h15.4v48.4V186.2z}
		svg{M108.9,79.1h41.6c39.6,0,57,28.3,57,53.6c0,27.5-21.5,53.6-56.8,53.6h-41.8V79.1z M124.3,172.4h24.5c34.9,0,42.9-26.5,42.9-39.7c0-21.5-13.7-39.7-43.7-39.7h-23.7V172.4z}
		svg{M88.7,56.8c0,5.5-4.5,10.1-10.1,10.1c-5.6,0-10.1-4.6-10.1-10.1c0-5.6,4.5-10.1,10.1-10.1C84.2,46.7,88.7,51.3,88.7,56.8z};
	}
}

\newcommand\orcidicon[1]{\href{https://orcid.org/#1}{\mbox{\scalerel*{
				\begin{tikzpicture}[yscale=-1,transform shape]
				\pic{orcidlogo};
				\end{tikzpicture}
			}{|}}}}
		
\usepackage[hidelinks]{hyperref}

\begin{document}
%
\title{Random Hyperboxes}
%
%
%

\author{Thanh Tung Khuat$^{\textsuperscript{\orcidicon{0000-0002-6456-8530}}}$,~\IEEEmembership{Student Member,~IEEE},~and~Bogdan Gabrys$^{\textsuperscript{\orcidicon{0000-0002-0790-2846}}}$,~\IEEEmembership{Senior Member,~IEEE}

\thanks{T.T. Khuat (email: thanhtung.khuat@student.uts.edu.au) and B. Gabrys (email: Bogdan.Gabrys@uts.edu.au) are with Advanced Analytics Institute, Faculty of Engineering and Information Technology, University of Technology Sydney, Ultimo, NSW 2007, Australia.
}}

%
%

\markboth{}
{Khuat and Gabrys: Random Hyperboxes}

%



\maketitle

\begin{abstract}
This paper proposes a simple yet powerful ensemble classifier, called Random Hyperboxes, constructed from individual hyperbox-based classifiers trained on the random subsets of sample and feature spaces of the training set. We also show a generalization error bound of the proposed classifier based on the strength of the individual hyperbox-based classifiers as well as the correlation among them. The effectiveness of the proposed classifier is analyzed using a carefully selected illustrative example and compared empirically with other popular single and ensemble classifiers via 20 datasets using statistical testing methods. The experimental results confirmed that our proposed method outperformed other fuzzy min-max neural networks, popular learning algorithms, and is competitive with other ensemble methods. Finally, we indentify the existing issues related to the generalization error bounds of the real datasets and inform the potential research directions.
\end{abstract}


%
\IEEEpeerreviewmaketitle

%
%
%
%

\section{Introduction}
\label{intro}
\IEEEPARstart{A} random hyperboxes (RH) classifier encompasses many individual hyperbox-based learners, e.g., fuzzy min-max neural networks (FMNNs) \cite{Gabrys2000}. One of the key characteristics of hyperbox-based classifiers is the single-pass through the training data learning ability. Based on this incremental learning ability, new data and classes can be added to the model without retraining the whole network. Another interesting characteristic of hyperbox-based models is their interpretability thanks to the human understandable rule sets which can be extracted directly or indirectly from hyperboxes. Interpretability is one of the key requirements when applying machine learning algorithms to high-stakes applications such as medical diagnostics, financial investment, self-driving systems, and criminal justice \cite{Rudin2019}.

The random hyperboxes model can be categorized into the family of ensemble classifiers, which build many base estimators and then combine them to create a final model. It is well-known that ensemble models are usually much more accurate than their base learners \cite{Biau2008}. There are two main methods to construct an ensemble model when using resampling methods and the same type of base learners. The first one aims to build many independent or low correlation individual estimators and combining their predictive outputs using majority voting or averaging approach. The representative models for this group include Bagging \cite{Breiman1996} and Random Forests \cite{Breiman2001}. The second paradigm consists of algorithms building base estimators in a sequential manner, where the newly added learner tries to correct errors generated by previous classifiers. Adaptive boosting (Adaboost) \cite{Freund1997} and Gradient Boosting Machines \cite{Friedman2001} are typical algorithms under the boosting framework. Extreme Gradient Boosting (XGBoost) \cite{Chen2016} and LightGBM \cite{Ke2017} are two recent effective and scalable implementations of the gradient boosting algorithm.

Our random hyperboxes classifier belongs to the first group because it shares the same principle with the bagging, i.e., using individual hyperbox-based learners with low correlation and combining their outputs by the majority voting. As shown in a recent study on hyperbox-based machine learning algorithms \cite{Khuat2019}, there is only one study \cite{Gabrys2002a} related to the use of bagging techniques with hyperbox-based models as base learners and another one which is concerned with method independent learning approaches for constructing either ensembles or individual hyperbox-based classifiers \cite{Gabrys04}. In their work, after training individual hyperbox-based estimators on different subsets of the training sets, the resulting base learners are combined at the decision level using the majority voting or averaging of membership values or combined at the model level into a single model. However, as it has been frequently shown resampling methods used with bagging like algorithms operating only in the sample space can generate a limited level of diversity amongst the base classifiers trained in this way. As the diversity amongst the base learners is of key importance \cite{ruta2005classifier}, there is another mechanism needed for making the resulting ensembles more effective and well performing. Based on Lemma \ref{lemma1}, adapted from \cite{Hastie2009}, it can be seen that the high correlation between base learners leads to a high testing error for the average classifier. To cope with this problem, we will lower the correlation but without significantly increasing the variance $\sigma$ of individual hyperbox-based learners by using only a subset of features when building base estimators. This fact can be achieved by utilizing feature subsets selected randomly for training each base classifier besides the subsets of samples. The use of a subsampling technique for both sample and feature spaces to construct the ensemble model constitutes the core principle of the random hyperboxes classifier. From surveys on hyperbox-based machine learning algorithms \cite{Khuat2019} and fuzzy min-max neural networks \cite{Sayaydeh2019}, it can be observed that this paper is the first study using randomized hyperbox estimators trained on subsets of both samples and features to construct an ensemble model.

\begin{lemma} \label{lemma1}
	Given $m$ identically distributed random variables (not necessarily independent) with the variance of each variable $\sigma^2$ and positive pairwise correlation $\mathbf{\rho}$, the variance of the average random variable is: \\
	\begin{equation}
	\rho \cdot \sigma^2 + \cfrac{1 - \rho}{m} \cdot \sigma^2
	\end{equation}
\end{lemma}
\begin{proof}
	See Appendix \ref{prooflemma1}.
\end{proof}

The use of subsets of features in building classifiers results in many effective models such as randomized trees on geometric feature selection \cite{Amit1997}, the random subspace-based decision forests \cite{Ho1998}, and random forests \cite{Breiman2001}. Recently, there have been several studies focusing on employing random projections of the feature vectors into a lower-dimensional space to form training data for classifiers such as Fisher's linear discriminant \cite{Durrant2015}, random projection neural network \cite{Andras2018}, or a general framework of random-projection based ensemble models \cite{Cannings2017}. These results have provided further motivation for the proposed random hyperboxes classifier.

One of the interesting characteristics of the proposed classifier is that it is easy to scale with large-sized training sets because each base learner can be constructed independently, so the learning process may be parallelised easily. Our contributions in this paper can be summarized as follows:

\begin{itemize}
	\item We propose a new ensemble classifier built from individual hyperbox-based learners using random subsets of both sample and feature spaces.
	\item We derive a generalization error bound of the RH classifier based on the strength and correlation between base learners.
	\item We analyze the effectiveness of the RH classifier in comparison to its base learners concerning the decrease in the variance of the ensemble model and the increase in the accuracy. We have also conducted extensive experiments on 20 datasets to compare the performance of the proposed method to other FMNNs as well as popular single and ensemble classifiers.
	\item We discuss the generalization error bounds on the real dataset and inform the open research directions.
\end{itemize}

The rest of this paper is structured as follows. Section \ref{preliminary} presents the general fuzzy min-max neural network (GFMMNN) and its learning algorithms used for base learners. In section \ref{proposed_method}, the formal description of the proposed method is provided and the generalization error bounds are derived. Section \ref{experiment} is devoted to experimental results. We discuss several issues concerning the generalization error bounds on the real datasets and identify the open problems in Section \ref{discussion_bound}. Section \ref{conclusion} concludes the findings and proposes directions for the future work.

\section{Preliminaries} \label{preliminary}
The RH classifier is constructed from base learners which can deploy any hyperbox-based machine learning algorithms. However, in this paper, we use the GFMMNN as base learners to assess the efficiency of the proposed method. Therefore, this part provides the readers with some basic knowledge of the GFMMNN and its learning algorithms.

The GFMMNN \cite{Gabrys2000} is a generalized version of FMNNs for classification \cite{Simpson1992} and clustering \cite{Simpson1993}. Its structure includes three layers, in which the input layer can accept both crisp and fuzzy data. Therefore, the input layer contains $2p$ nodes corresponding to $p$ features of the input data which can be represented in the form of lower and upper bounds (i.e. as a real interval). The second layer consists of hyperboxes dynamically created during the learning process. The connection weights between the first and the second layers are the minimum points \textbf{V} and the maximum points \textbf{W} of hyperboxes, which are adjusted in the learning process. The connection between the hyperbox $B_i$ in the second layer and an output node $c_i$ in the third layer $u_{ij}$ is stored in the matrix \textbf{U} such that:
\begin{equation}
u_{ij} = \begin{cases}
1, \mbox{if } class(B_i) = c_j \\
0, \mbox{otherwise}
\end{cases}
\end{equation}

In the GFMMNN, the degree of fit of each hyperbox $B_i = [V_i, W_i]$, where minimum point $V_i = [v_{i1},\ldots,v_{ip}]$ and maximum point $W_i = [w_{i1},\ldots,w_{ip}]$, with respect to each input pattern $\mathbf{x} = [\mathbf{x}^l, \mathbf{x}^u]$ is computed using a membership function as Eq. \eqref{eq_mem}.
\begin{equation}
\label{eq_mem}
\begin{split}
b_i(\mathbf{x}, B_i) = \min \limits_{j = 1}^{p} (\min(&[1 - f(x_{j}^u - w_{ij}, \gamma_j)], \\
&[1 - f(v_{ij} - x_{j}^l, \gamma_j)]))
\end{split}
\end{equation}
where $ f(\xi, \gamma) $ is two-parameter ramp function described in Eq. \eqref{eq_ramp}, $ \gamma = (\gamma_1, \gamma_2,..., \gamma_p) $ contains the sensitivity parameters regulating the decreasing speed of the membership values, and $ 0 \leq b_i(\mathbf{x}, B_i) \leq 1 $.
\begin{equation}
\label{eq_ramp}
f(\xi, \gamma) = \begin{cases}
1, & \mbox{if } \xi \cdot \gamma > 1 \\
\xi \cdot \gamma, & \mbox{if } 0 \leq \xi \cdot \gamma \leq 1 \\
0, & \mbox{if } \xi \cdot \gamma < 0
\end{cases}
\end{equation}

In the classification phase, assuming that the membership value between the input $\mathbf{x}$ and the hypberbox $B_i$ is the highest compared to other existing hyperboxes, the predictive class of the model for the input $\mathbf{x}$ is the class of $B_i$.

Given a training set, there are two types of learning algorithms used to train the GFMM classifier, i.e., the incremental (online) learning \cite{Gabrys2000} and agglomerative (batch) learning \cite{Gabrys2002b}. The batch learning algorithm starts with all of the training samples and then repeatedly merging hyperboxes with the same class satisfying the maximum hyperbox size ($\theta$), minimum similarity threshold ($\sigma_s$), and no generation of overlapping regions with hyperboxes of other classes. The training time of this algorithm is long because of the iterative computation of membership and similarity values between all pairs of existing hyperboxes. In contrast, the online learning algorithm is much faster since it uses a single pass mechanism through learning samples to build and adjust hyperboxes. However, the hyperbox contraction process to resolve hyperbox overlapping areas causes a decrease in predictive accuracy \cite{Bargiela2004}. In a recent study, an improved online learning algorithm of GFMMNN, called IOL-GFMM, has been proposed to combine the strong points of both incremental and batch learning algorithms. Therefore, in this paper, the IOL-GFMM will be used to build base hyperbox classifiers. We would like to refer the readers to \cite{Khuat2020iol} for more details of the IOL-GFMM algorithm.

\section{Proposed Method} \label{proposed_method}
\subsection{Formal Description}
Let us denote by $\mathcal{T}_n = \{(\mathbf{x}_i, c_i)\}_{i = 1}^n$ a training data where $x_i \in \mathbf{X} \subset \R^p$ is a $p$-dimensional vector of observations (i.e. features) and $c_i \in \mathcal{C}$, $\mathcal{C}$ is a set of categorical variables denoting classes to which the observations fall. Given an input $\mathbf{x}$,  our goal is to build an ensemble classifier which predicts class $c$ from $\mathbf{x}$ using the training data $\mathcal{T}_n$. 

Please note that for the theoretical considerations of the proposed algorithm covered in this section and the discussion of the convergence properties and the derivation of generalisation error bounds presented in Section \ref{Properties}, an assumption is made that the observations are independent and identically distributed (i.i.d.) random variables. 

A random hyperboxes model with $m$ hyperbox-based learners is a classifier including a set of randomized base hyperbox models $h(\mathbf{x}, \Phi_1), \ldots, h(\mathbf{x}, \Phi_m)$, where $\Phi_1, \ldots, \Phi_m$ are i.i.d. random vectors of a randomizing vector $\Phi$, independent conditionally on $\mathbf{X}, \mathcal{C}$, and $\mathcal{T}_n$. Each individual hyperbox-based learner $h(\mathbf{x}, \Phi_i)$ is constructed using the training set $\mathcal{T}_n$ and a random vector $\Phi_i$. $\Phi_i$ introduces the randomness to the building process of hyperbox-based learners including the determining of a subset $\mathcal{T}_{\Phi_i}$ of the full training data $\mathcal{T}_n$ as well as determining a subset of features $\mathbf{x}_{\Phi_i}$ used. After a large number of hyperbox-based learners genereated, the random hyperboxes estimator takes the class with most votes among base learners as its predictive result. Formally, the definition of the random hyperboxes classifier can be stated as follows:

\begin{definition}
	A random hyperboxes model is a classifier including a set of hyperbox-based learners $\{h(\mathbf{x}, \Phi_i): i = 1, \ldots, m\}$, where $\{\Phi_i\}$ are independent and identically distributed random vectors of a model random vector $\Phi$ independent conditionally on sample space $(\mathbf{X}, \mathcal{C})$ and the training set $\mathcal{T}_n$. Each hyperbox-based learner gives a unit vote based on the class of the hyperbox with the maximum membership degree with respect to the input pattern $\mathbf{x}$. The predictive result of the random hyperboxes model is the aggregation of predictive results from its base learners using a majority voting method.
\end{definition}

In particular, the predictive class $(c_k \in \mathcal{C})$ with respect to input data $\mathbf{x}$ of a random hyperboxes classifier including $m$ base learners (let $\Phi^{(m)} = \{\Phi_1,\ldots,\Phi_m \}$) can be shown as follows:
\begin{equation*}
h(\mathbf{x}, \Phi^{(m)}) = arg \max \limits_{c_k \in \mathcal{C}} \cfrac{1}{m} \sum \limits_{i = 1}^m \mathbb{1} (h(\mathbf{x}, \Phi_i) = c_k)
\end{equation*}
where $\mathbb{1}(\cdot)$ is the indicator function. According to the strong law of large numbers, when the number of base learners increases, we almost surely obtain $\lim \limits_{m \rightarrow \infty} h(\mathbf{x}, \Phi^{(m)}) = \overline{h}(\mathbf{x}, \Phi)$, where $\overline{h}(\mathbf{x}, \Phi) = arg \max \limits_{c_k \in \mathcal{C}} \mathbb{E}_{\Phi}[\mathbb{1} (h(\mathbf{x}, \Phi) = c_k)]$ (Here $\mathbb{E}_\Phi$ denotes the expectation with regard to the random variable $\Phi$).

\begin{algorithm}[!ht]
	\scriptsize{
		\caption{Training algorithm of the Random hyperboxes}
		\label{alg_rh}
		\begin{algorithmic}
			\STATE {\bfseries Input:} training set $\mathcal{T}$, sampling rate for samples $r_s$, maximum number of used features $m_f$, number of base estimators $m$, maximum hyperbox size $\theta$, sensitivity parameter $\gamma$
			\STATE {\bfseries Output:} A random Hyperboxes model $\mathbf{H}$
			\item[]
			\STATE $i = 1; \mathbf{H} \leftarrow \varnothing$
			\FOR{$i \leq m$}
			\STATE $T_i \leftarrow$ Perform subsampling on $\mathcal{T}$ with rate $r_s$
			\STATE $d \leftarrow $ Generate a uniform random number in the range of $[1, m_f]$
			\STATE $T_i^d \leftarrow $ Random sampling $d$ features of $T_i$
			\STATE $h_i \leftarrow $ \textbf{IOL-GFMM}($T_i^d, \gamma, \theta$)
			\STATE $\mathbf{H} \leftarrow \mathbf{H} \cup h_i$
			\STATE $ i = i + 1$
			\ENDFOR
			\STATE {\bfseries return} $\mathbf{H}$
		\end{algorithmic}
	}
\end{algorithm}

Each random hyperbox-based learner $h(\mathbf{x}, \Phi)$ is formed as follows. We select randomly a subset $\mathcal{T}_l$ including $l < n$ samples from the full training data $\mathcal{T}_n$ using subsampling method without replacement under weak assumptions $l \rightarrow 0$ and $r_s = l/n \rightarrow 0$ as $n \rightarrow \infty$. According to \cite{politis1999}, under the weak convergence hypothesis, the sampling distributions of $\mathcal{T}_l$ and $\mathcal{T}_n$ should be close, and they will converge to the true unknown distribution of whole sample space. After that, we will select at uniformly random $d$ ($1 \leq d \leq m_f \leq p$) features from $p$ features of $\mathcal{T}_l$ to form a training set $\mathcal{T}_l^{(d)}$ for $h(\mathbf{x}, \Phi)$, where $m_f$ is the maximum features used for each base learner. There are many learning algorithms which could be used to train the base hyperbox-based classifier $h(\mathbf{x}, \Phi)$ on $T_l^{(d)}$. This study uses the IOL-GFMM \cite{Khuat2020iol} to build the base estimators. This is a new online learning algorithm of GFMM which integrates the advantages of the incremental learning and batch learning algorithms for the building process of a GFMMNN. It is noted that the base model $h(\mathbf{x}, \Phi)$ is trained on only $d$ features of $\mathcal{T}_n$, so in the classification step, $h(\mathbf{x}, \Phi)$ only makes prediction using the same $d$ features with respect to the unseen sample $\mathbf{x}$. The learning and classification steps for each base learner are kept the same as in the IOL-GFMM algorithm.

The basic steps of the building process of the random hyperboxes classifier are shown in Algorithm \ref{alg_rh}.

\subsection{Time Complexity}
Based on Algorithm \ref{alg_rh}, it is easily observed that the time complexity of a random hyperboxes model depends mainly on the time complexity of the training process for each base learner. As discussed in \cite{Khuat2020acc}, the time complexity of the IOL-GFMM algorithm trained on a dataset containing $n$ samples with $p$ features is $\mathcal{O}(n \cdot \mathcal{K} \cdot \mathcal{R} \cdot p)$, where $\mathcal{K}$ is the average number of expandable hyperbox candidates and $\mathcal{R}$ is the average number of hyperboxes representing classes different from the input pattern class for each iteration in the training process. For the random hyperboxes model, each base learner is trained on only $l < n$ samples with the maximum $m_f < p$ features. Therefore, the time complexity of each base learner in the worst case is $\mathcal{O}(l \cdot \mathcal{K} \cdot \mathcal{R} \cdot m_f)$. We need to build $m$ base learners for a random hyperboxes classifier. As a result, if the base learners are sequentially constructed, the time complexity of training a random hyperboxes model in the worst case is $\mathcal{O}(m \cdot l \cdot \mathcal{K} \cdot \mathcal{R} \cdot m_f)$.

\subsection{Properties of the Random Hyperboxes} \label{Properties}
\subsubsection{The Convergence of the Random Hyperboxes Model} \hfill

Let $\mathbf{x}$ be a random sample, drawn from the sample space, to be classified with true class $c$. Let $\mathcal{T}_n$ be a random training set drawn i.i.d. from the true distribution of sample space $(\mathbf{X}, \mathcal{C})$. Given an ensemble of $m$ base learners $h_1(\mathbf{x}), \ldots, h_m(\mathbf{x})$, where $h_i(\mathbf{x}) \equiv h(\mathbf{x}, \Phi_i)$, we can define a margin function of a random hyperboxes model with $m$ base estimators for an input sample $\mathbf{x}$ as Eq. \eqref{margin_func}:
\begin{equation}
\label{margin_func}
\mathcal{M}(\mathbf{x}, c) = \cfrac{1}{m} \sum_{i=1}^m{\mathbb{1}(h_i(\mathbf{x}) = c)} - \max_{j \ne c}\cfrac{1}{m} \sum_{i=1}^m{\mathbb{1}(h_i(\mathbf{x}) = j)}
\end{equation}
where $\mathbb{1}(\cdot)$ is the indicator function.

\begin{remark}
	The margin can be considered as a confidence measure with respect to the classification result of the random hyperboxes model. A large margin increases the confidence in predictive results for observations and vice versa.
\end{remark}

Based on the above margin function, the generation error of the random hyperboxes model is defined as follows:

\begin{definition}
	\label{def1}
	The generalization error is the probability $\mathbf{P}_{\mathbf{X}, \mathcal{C}}$ measured in the sample space $(\mathbf{X}, \mathcal{C})$ that gives a negative margin: $\mathcal{E} = \mathbf{P}_{\mathbf{X}, \mathcal{C}}(\mathcal{M}(\mathbf{x}, c) < 0)$
\end{definition}

\begin{lemma}
	\label{lemma2}
	When the number of base estimators increases $(m \rightarrow \infty)$ and base estimators are independent, for almost surely all i.i.d. random vectors $\Phi_1, \Phi_2, \ldots$, the margin function for a random hyperboxes model $\mathcal{M}(\mathbf{x}, c)$ at each input $\mathbf{x}$ converges to:
	\begin{equation}
	\mathcal{M}^*(\mathbf{x}, c) =  \mathbf{P}_\Phi(h(\mathbf{x}, \Phi) = c) - \max_{j \ne c} \mathbf{P}_\Phi(h(\mathbf{x}, \Phi) = j)
	\end{equation}
\end{lemma}
\begin{proof}
	See Appendix \ref{prooflemma2}.
\end{proof}

From definition \ref{def1} and lemma \ref{lemma2}, we achieve the following theorem for the convergence of generalization error:

\begin{theorem}
	When the number of base learners increases $(m \rightarrow \infty)$, for almost surely all random vectors $\Phi_1, \Phi_2, \ldots$, the generalization error $\mathcal{E}$ converges to: $\mathcal{E}^* = \mathbf{P}_{\mathbf{X}, \mathcal{C}}\left[\mathcal{M}^*(\mathbf{x}, c) < 0\right]$ 
\end{theorem}

This theorem explains that the random hyperboxes model does not overfit when more base learners are added to the model if hyperbox-based learners are independent and under the i.i.d. assumption. In the next subsection, the upper bound of the generalization error will be derived.

\subsubsection{Generalization Error Bound} \hfill

Based on Lemma \ref{lemma1}, we can observe that to decrease the variance of the average classifier, we need to reduce the correlation of base learners. However, if the correlation decreases, the variance of base learners usually increases, and it makes the reduction of the prediction error harder. The correlation among base learners can be easily decreased by increasing base models' randomness. However, in this way the variance of the base learners will also be increased. Therefore, we should not let the variance increase too fast. To cope with this issue, we can inspect and monitor the change in the generalization error bound.

Instead of having a fixed number of base estimators $m$, let us assume that we have a fixed probability distribution for the random vector $\Phi$ from which base models are constructed. Similarly to random forests \cite{Breiman2001}, we can define the strength of the random hyperbox model based on the limit of the margin function as follows:

\begin{definition}
	The strength of the random hyperboxes model is defined as:
	\begin{equation}
	\mathcal{S} = \mathbb{E}_{\mathbf{X}, \mathcal{C}} \mathcal{M}^*(\mathbf{x}, c)
	\end{equation}
\end{definition}
where $\mathbb{E}_{\mathbf{X}, \mathcal{C}}$ is the expectation through the $(\mathbf{X}, \mathcal{C})$ space.

Assuming that $\mathcal{S} > 0$, according to Chebyshev's inequality, we have:
\begin{equation*}
\begin{split}
\mathcal{E}^* &= \mathbf{P}_{\mathbf{X}, \mathcal{C}}\left[\mathcal{M}^*(\mathbf{x}, c) < 0\right] \leq \mathbf{P}_{\mathbf{X}, \mathcal{C}}\left[\mathcal{S} - \mathcal{M}^*(\mathbf{x}, c) \geq \mathcal{S}\right]\\
&= \mathbf{P}_{\mathbf{X}, \mathcal{C}}\left[|\mathcal{M}^*(\mathbf{x}, c) - \mathcal{S}| \geq \mathcal{S}\right] \leq \cfrac{\mathtt{Var}_{\mathbf{X}, \mathcal{C}}(\mathcal{M}^*(\mathbf{x}, c))}{\mathcal{S}^2}
\end{split}
\end{equation*}
This is a weak upper bound of the generalization error, and it indicates that the prediction error is always lower than an explicit but unknown limit. The value of $\mathcal{S}$ can be estimated over the training set $\mathcal{T}_n$ as follows:
\begin{equation*}
\begin{split}
\overline{\mathcal{S}} &= \cfrac{1}{n} \sum \limits_{i = 1}^n \mathcal{M}(\mathbf{x}_i, c_i) \\
&= \cfrac{1}{nm} \sum \limits_{i=1}^n \Bigl( \sum \limits_{k=1}^m \mathbb{1}(h_k(\mathbf{x}_i) = c_i) -
\max \limits_{j \ne c_i} \sum \limits_{k=1}^m \mathbb{1}(h_k(\mathbf{x}_i) = j) \Bigr)
\end{split}
\end{equation*}

Let $J(\mathbf{x}, c) = arg \max\limits_{j \ne c} \mathbf{P}_{\Phi}(h(\mathbf{x}, \Phi) = j)$ be the class $j$ leading to the most incorrect classification of base learners with respect to the input $\mathbf{x}$. Then, we can define a raw margin function for each base learner at each input $\mathbf{x}$ as follows:

\begin{definition}
	The raw margin function is defined by:
	\begin{equation}
	\begin{split}
	\mathcal{R}(\Phi) = \mathcal{R}(\mathbf{x}, c, \Phi) = &\mathbb{1}(h(\mathbf{x}, \Phi) = c) \\
	&- \mathbb{1}(h(\mathbf{x}, \Phi) = J(\mathbf{x}, c))
	\end{split}
	\end{equation}
\end{definition}
Following from the above definition, \begin{equation*}\begin{split}
\mathcal{M}^*(\mathbf{x}, c) &= \mathbf{P}_{\Phi}(h(\mathbf{x}, \Phi) = c) - \mathbf{P}_{\Phi}(h(\mathbf{x}, \Phi) = J(\mathbf{x}, c)) \\
&= \mathbb{E}_{\Phi} \left[\mathbb{1}(h(\mathbf{x}, \Phi) = c) - \mathbb{1}(h(\mathbf{x}, \Phi) = J(\mathbf{x}, c)) \right] \\
&= \mathbb{E}_{\Phi} \mathcal{R}(\Phi)
\end{split}\end{equation*}
It means that the limit of the margin values is the expectation of raw margin values computed over all realizations of $\Phi$.

From the above raw margin function, we now can define the correlation between two hyperbox-based learners $h(\mathbf{x}, \Phi_i)$ and $h(\mathbf{x}, \Phi_j)$ generated from two i.i.d. random vectors $\Phi_i$ and $\Phi_j$ as follows:
\begin{definition}
	The correlation between two hyperbox-based learners $h(\mathbf{x}, \Phi_i)$ and $h(\mathbf{x}, \Phi_j)$ of a random hyperboxes model can be calculated from the raw margin function through all observations as follows:
	\begin{equation}
	\rho_{\mathbf{X}, \mathcal{C}}(\Phi_i, \Phi_j) = \cfrac{\mathtt{Cov}_{\mathbf{X}, \mathcal{C}}(\mathcal{R}(\Phi_i), \mathcal{R}(\Phi_j))}{\sigma_{\mathbf{X}, \mathcal{C}} (\mathcal{R}(\Phi_i)) \sigma_{\mathbf{X}, \mathcal{C}} (\mathcal{R}(\Phi_j))}
	\end{equation}
\end{definition} 
where $\mathtt{Cov}$ is the covariance, $\sigma_{\mathbf{X}, \mathcal{C}} (\mathcal{R}(\Phi_i))$ denotes the standard deviation of $\mathcal{R}(\Phi_i)$, holding $\Phi_i$ fixed, computed over observations.

Generally, the average correlation between base learners in the random hyperboxes models is computed through all pairs of two i.i.d. random vectors $\Phi$ and $\Phi'$ as follows:
\begin{equation}
\overline{\rho} = \mathbb{E}_{\Phi, \Phi'} [\rho_{\mathbf{X}, \mathcal{C}}(\Phi, \Phi')]
\end{equation}

From the average correlation between base learners and the strength $\mathcal{S}$, we have the following theorem for the upper bound of the generalization error:
\begin{theorem}
	An upper bound of the generalization error for the random hyperboxes model can be estimated from the strength of base learners and correlation between base learners as follows:
	\begin{equation}
	\mathcal{E}^* \leq \overline{\rho} \; \Bigl(\cfrac{1}{\mathcal{S}^2} - 1 \Bigr)
	\end{equation}
\end{theorem}
\begin{proof}
	See Appendix \ref{prooftheorem2}.
\end{proof}

\section{Experimental Results}\label{experiment}

It is noted that the derivations and proofs in the previous section have been carried out under the i.i.d. assumption which in practice is difficult to verify and is very often not satisfied. In this section and the appendices we are, therefore, conducting extensive benchmarking and experimental evaluation of the proposed method to also verify its practical characteristics and performance.
\subsection{Analyzing the Random Hyperboxes Classifier}
\subsubsection{The Decrease in the Variance Compared to Base Learners}\label{variance} \hfill

To conduct this experiment, we used six datasets with diversity in the numbers of samples, features, and classes. All of the experimental results are shown in Appendix \ref{sup_variance}. This section only illustrates the results for a dataset of the one-hundred plant species leaves for margin \cite{Mallah2013}. This dataset includes 1600 samples with 64 features and 100 classes. We performed 10 times repeated 4-fold cross-validation to evaluate the ensemble model with 100 base learners. Therefore, there are 4000 base learners using the IOL-GFMM algorithm and 40 random hyperboxes models generated. The variance values in terms of weighted-F1 scores of base learners and the random hyperboxes models are shown in Fig. \ref{fig1}. The variance values of other datasets are shown in Fig. \ref{Fig_S1} in Appendix \ref{sup_variance}. These results confirmed that the variance of random hyperboxes models using simple majority voting is significantly reduced compared to their base learners, so its classification accuracy is also higher than that of base estimators.
\begin{figure}[!ht]
	\centering
	\includegraphics[width=0.32\textwidth]{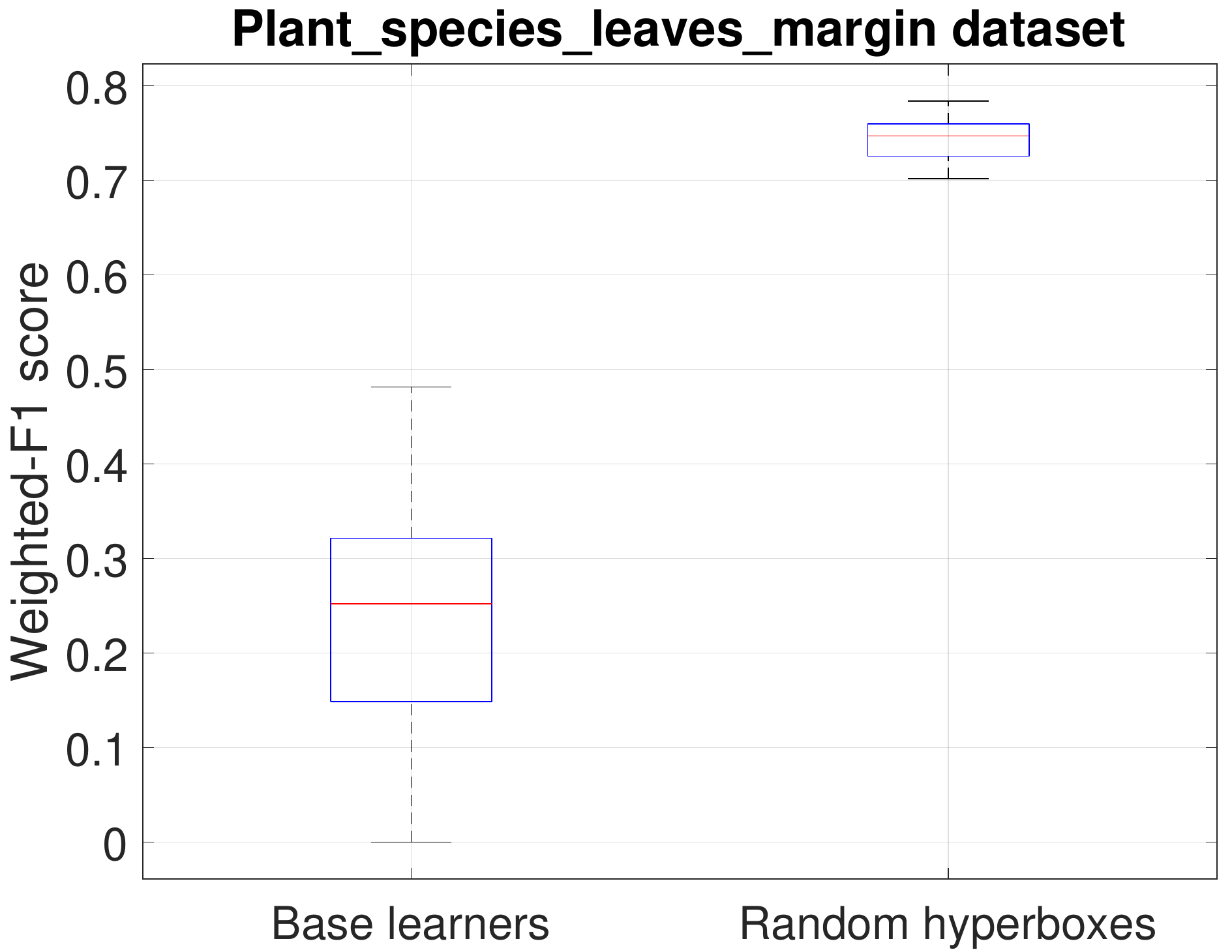}
	\caption{The variances of RH models and their base learners (\textit{plant\_species\_leaves\_margin} dataset).}
	\label{fig1}
\end{figure}

In this experiment, we set the maximum number of used features $m_f = 2 \sqrt{p} = 16$ (for the \textit{plant\_species\_leaves\_margin} dataset) and 50\% of the training data samples were randomly selected to train each base learner. The probability of the number of features, $d$, used to build the 4000 base learners is shown in Fig. \ref{Fig_S2} in Appendix \ref{sup_variance}. The importance scores of features through all base learners can be identified using the used probability of each feature, as shown in Fig. \ref{Fig_S3} in Appendix \ref{sup_variance}.

\begin{figure}[!ht]
	\centering
	\includegraphics[width=0.34\textwidth]{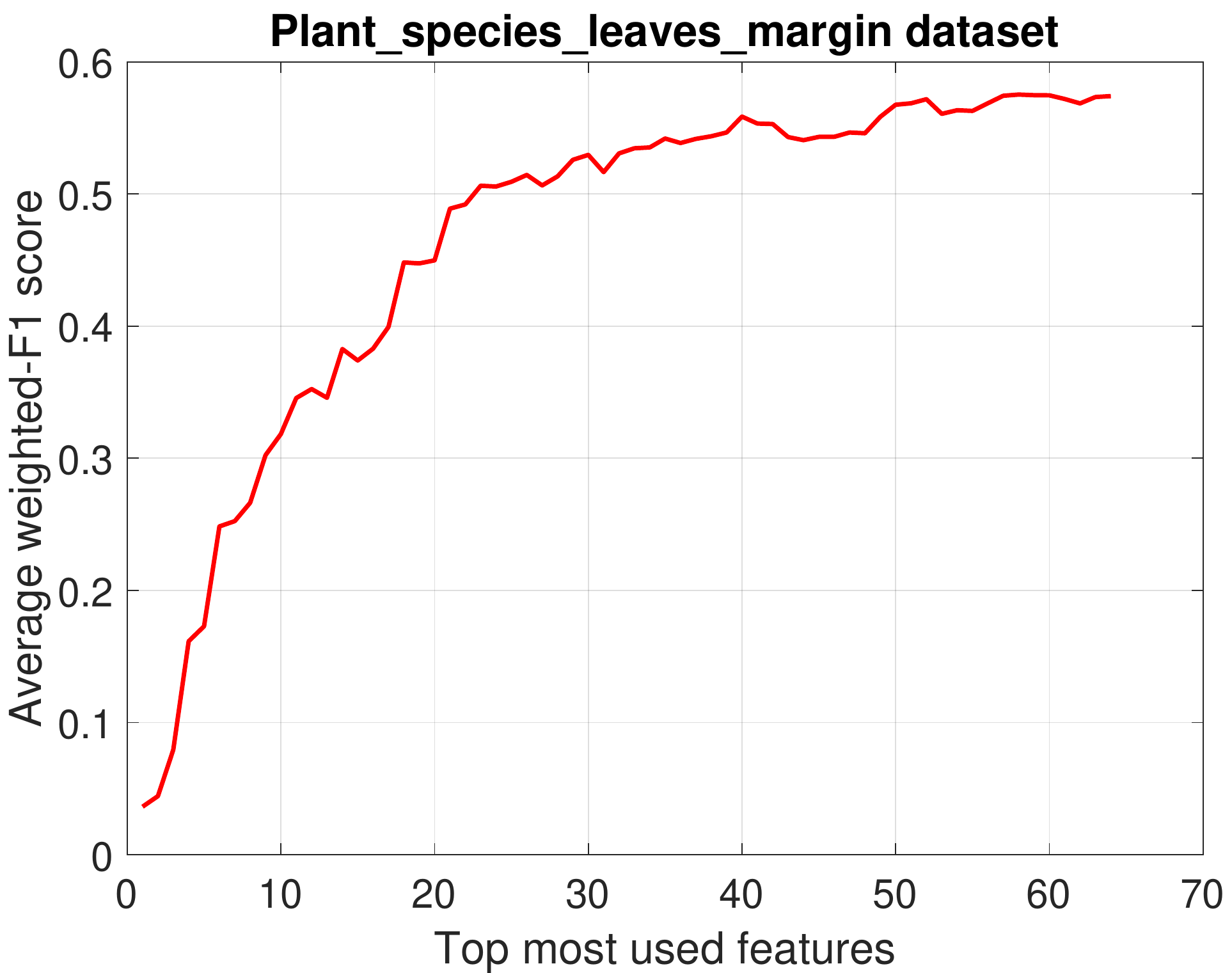}
	\caption{Average weighted-F1 scores through 40 testing folds of a single model using training sets with top-k most used features (\textit{plant\_species\_leaves\_margin} dataset).}
	\label{fig2}
\end{figure}

Based on the probability that each feature is used in 4000 base learners, we can determine the contribution of the combination of features to the performance of each classifier. Therefore, we have trained a single model using the IOL-GFMM algorithm using top-K most used features $(K = 1,\ldots,p)$ ($p = 64$ for the \textit{plant\_species\_leaves\_margin} dataset) in each iteration. Fig. \ref{fig2} shows the average weighted-F1 scores for 40 testing folds (10 times repeated 4-fold cross-validation) for each top-K of the most often used features in the \textit{plant\_species\_leaves\_margin} dataset. The results for the other datasets can be found in Fig. \ref{Fig_S4} in Appendix \ref{sup_variance}. It can be seen that the single model usually achieves the best performance if it is trained on all features. However, by using the random hyperboxes method with base learners trained on only a maximum of $m_f$ features, we can obtain a higher accuracy than the single model trained on all features. Furthermore, in several datasets such as \textit{ringnorm} and \textit{connectionist\_bench\_sonar}, the best performance is often obtained when using a subset of the most crucial features. It is due to the fact that the redundant features can prevent the single GFMM model from learning the true distribution of the underlying data with a given finite number of training samples. Therefore, the use of the random hyperboxes model of which base learners are trained on a subset of features can capture the data distribution more effectively and achieve better classification performance compared to the case of employing of a single GFMM model.

In general, the RH classifier can achieve much better performance compared to the single IOL-GFMM classifier trained on full feature space, especially for very high dimensional datasets. These results are shown in Appendix \ref{sup_high_dim}.

\subsubsection{The Roles of the Number of Base Learners and Maximum Number of Used Features} \hfill

\begin{figure}[!ht]
	\centering
	\includegraphics[width=0.32\textwidth]{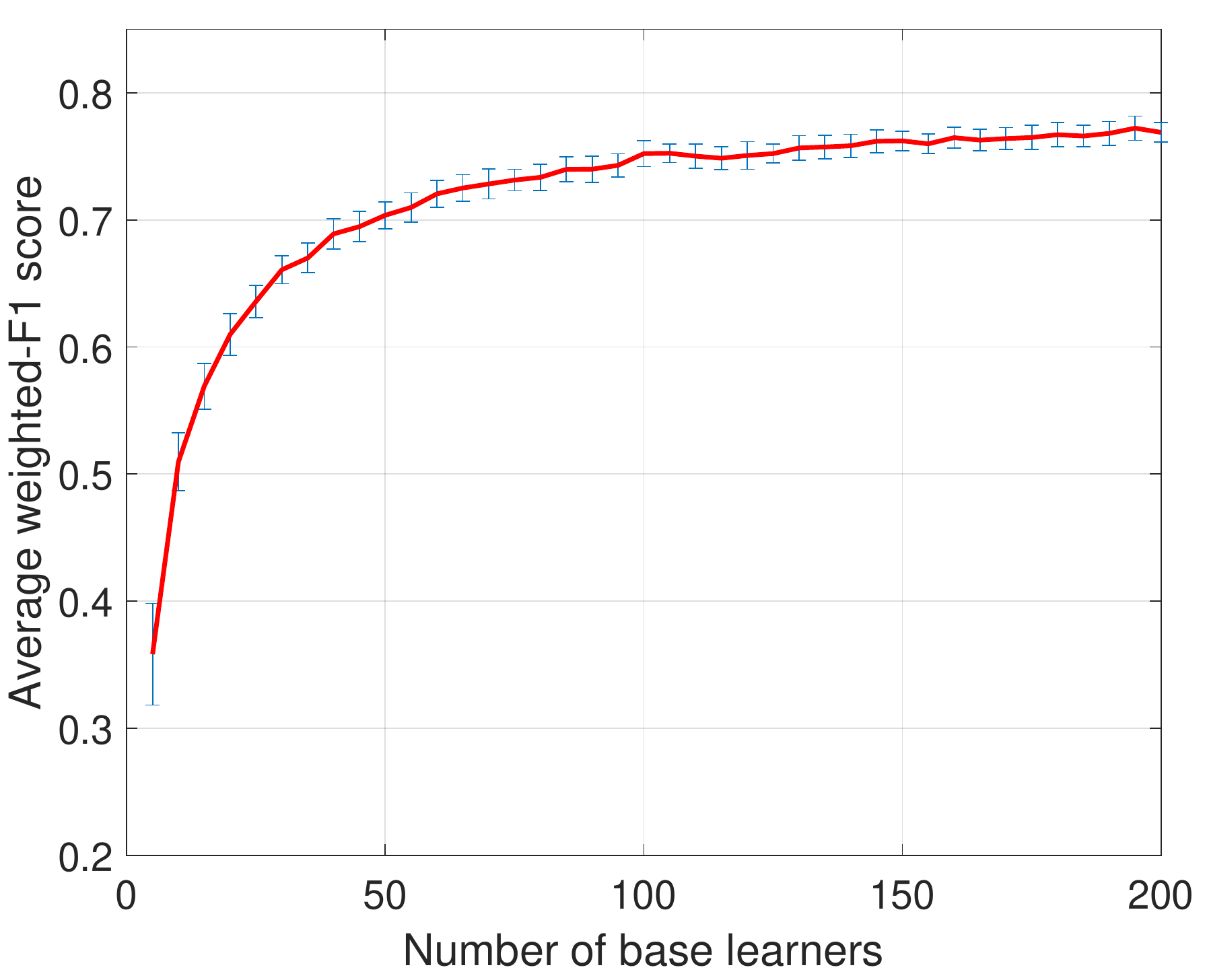}
	\caption{The change in the average weighted-F1 scores when increasing the number of base learners (\textit{plant\_species\_leaves\_margin} dataset).}
	\label{fig3}
\end{figure}

This experiment is to assess the sensitivity of hyper-parameters such as the number of base learners and the maximum number of used features on the performance of the random hyperboxes model. We used eight datasets with diversity in the numbers of samples, classes, and features for this purpose. All of the empirical results can be found in Appendix \ref{sup_sen_params}. This section only illustrates the outcomes of the same dataset used in subsection \ref{variance}. To evaluate the impact of the number of base learners on the performance of the random hyperboxes model, we kept the maximum number of used features $m_f = 2 \cdot \sqrt{p}$ ($m_f = 16$ in this case), the maximum hyperbox size of each base learner $\theta = 0.1$, and 50\% of samples were randomly selected to train each base estimator. The number of base learners is set from 5 to 200 with step 5. Fig. \ref{fig3} shows the average weighted-F1 scores over 10 times repeated 4-fold cross-validation at each threshold for the \textit{plant\_species\_leaves\_margin} dataset. The results for the other datasets can be found in Fig. \ref{Fig_S7} in Appendix \ref{sup_sen_params}. It can be observed that the performance of the random hyperboxes classifier is not reduced as more base learners are added. These figures confirm that the random hyperboxes classifier does not overfit when adding more base learners.

\begin{figure}[!ht]
	\centering
	\includegraphics[width=0.32\textwidth]{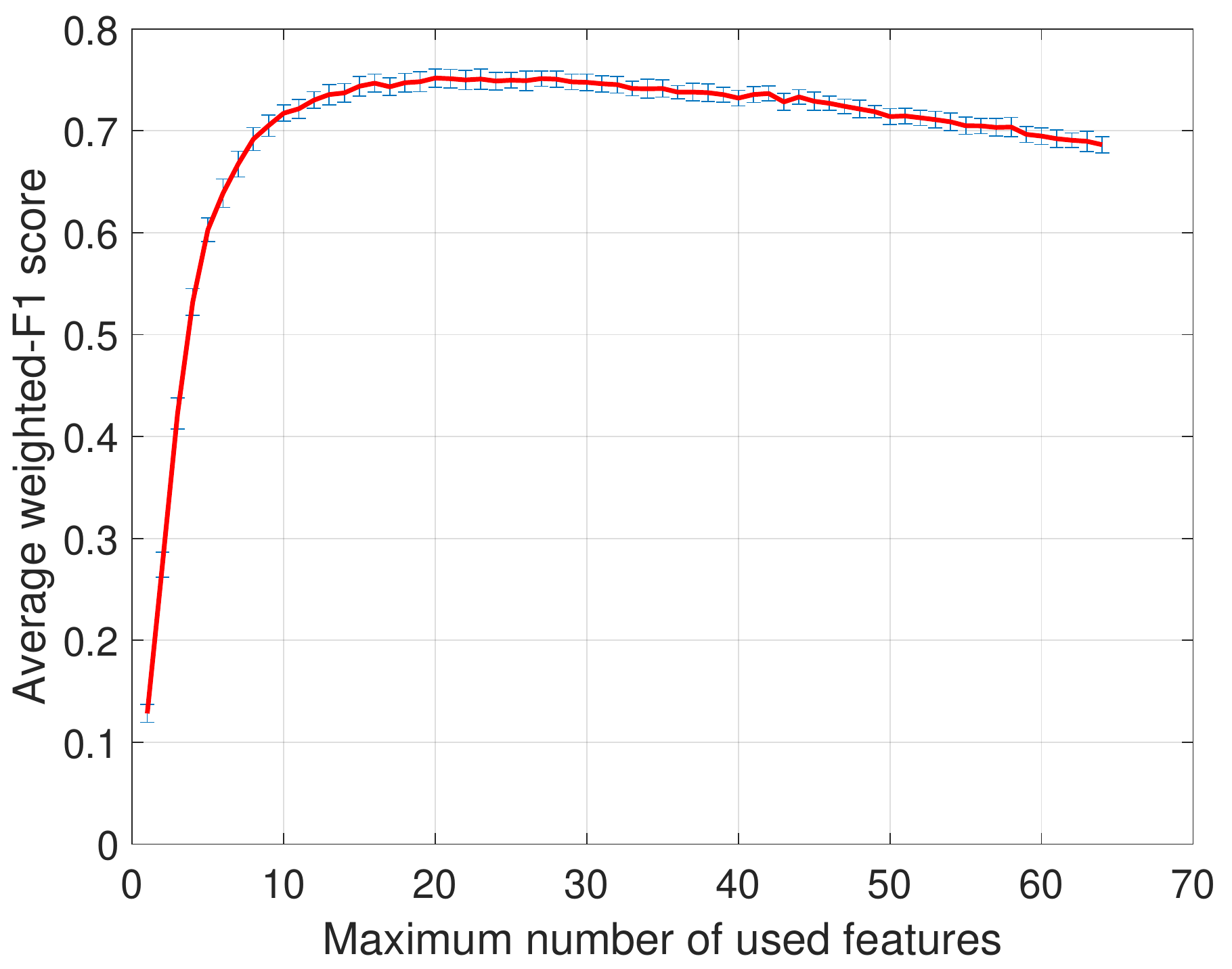}
	\caption{The change in the average weighted-F1 scores when increasing the maximum number of used dimensions (\textit{plant\_species\_leaves\_margin} dataset).}
	\label{fig4}
\end{figure}

To assess the influence of the maximum number of used features $m_f$, we kept the number of base learners $ m = 100$, $\theta = 0.1$, $r_s = 0.5$, and changed the maximum numbers of used features from 1 to $p$ ($p = 64$ in this case). Fig. \ref{fig4} depicts the average weighted-F1 scores for 10 times repeated 4-fold cross-validation at each value of the maximum number of used features for the \textit{plant\_species\_leaves\_margin} dataset. The outcomes for the remaining datasets are shown in Fig. \ref{Fig_S8} in Appendix \ref{sup_sen_params}.

It can be easily observed that the overall trend when increasing the maximum number of used features is that the accuracy of the random hyperboxes classifier only increases to a certain threshold, and then its accuracy will decrease. It is due to the fact that the correlation between base learners will be higher when we use too many features for each base learner. In contrast, if too few features are used, the strength of each base learner gets a low value, so the error of the ensemble model will increase. This fact confirms that the maximum number of used features is an important parameter, which needs to be carefully selected to achieve the high accuracy for the random hyperboxes classifier.

\subsection{Comparing the Performace of the Random Hyperboxes to Other Classifiers}
The datasets used and parameter settings for models are presented in Appendix \ref{sup_dataset}. The following results are the average weighted-F1 scores using 10 times repeated 4-fold cross-validation. In this study, we consider the multi-class classification problem, so the weighted-F1 measure is more suitable and less biased than the often used classification accuracy. Weighted F1-score is the average F1-score of each class weighted by the support which is the number of patterns of each class. In each iteration, three folds were used for training and one remaining fold was used as a testing set.

\subsubsection{A Comparison of the Random Hyperboxes With Other FMNNs} \hfill

This experiment compares the RH model with FMNN \cite{Simpson1992}, online learning version of GFMMNN (Onln-GFMM) \cite{Gabrys2000}, agglomerative learning algorithm version 2 of GFMMNN (AGGLO-2) \cite{Gabrys2002b}, combination of Onln-GFMM at $\theta=0.05$ and AGGLO-2 \cite{Gabrys2002b}, IOL-GFMM \cite{Khuat2020iol}, enhanced fuzzy min-max neural network (EFMNN) \cite{Mohammed2015}, enhanced fuzzy min-max neural network with k-nearest hyperbox selection rules (KNEFMNN) \cite{Mohammed2017}, and refined fuzzy min-max neural network (RFMNN) \cite{Sayaydeh2020}. The classification accuracy results of fuzzy min-max neural networks at low values of $\theta$ are usually better than those at high values of $\theta$ \cite{Khuat2020}. Therefore, in this experiment, we will compare the RH model with other FMNNs using $\theta = 0.1$ and $\theta = 0.7$. The average weighted-F1 scores of classifiers, as well as their ranks, are shown in Tables from \ref{cmp_other_fmnn_01} to \ref{cmp_other_fmnn_07_ranking} in Appendix \ref{sup_cmp_other_fmnn}. Fig. \ref{fig5} summarizes these results by comparing the results of the RH classifier with the best values of other FMNNs. We can see that in both subplots most points locate above the diagonal line, these figures illustrate the efficiency and robustness of the random hyperboxes for both low and high thresholds of $\theta$.

\begin{figure}[!ht]
	\begin{subfloat}[$\theta=0.1$]{
			\includegraphics[width=0.23\textwidth]{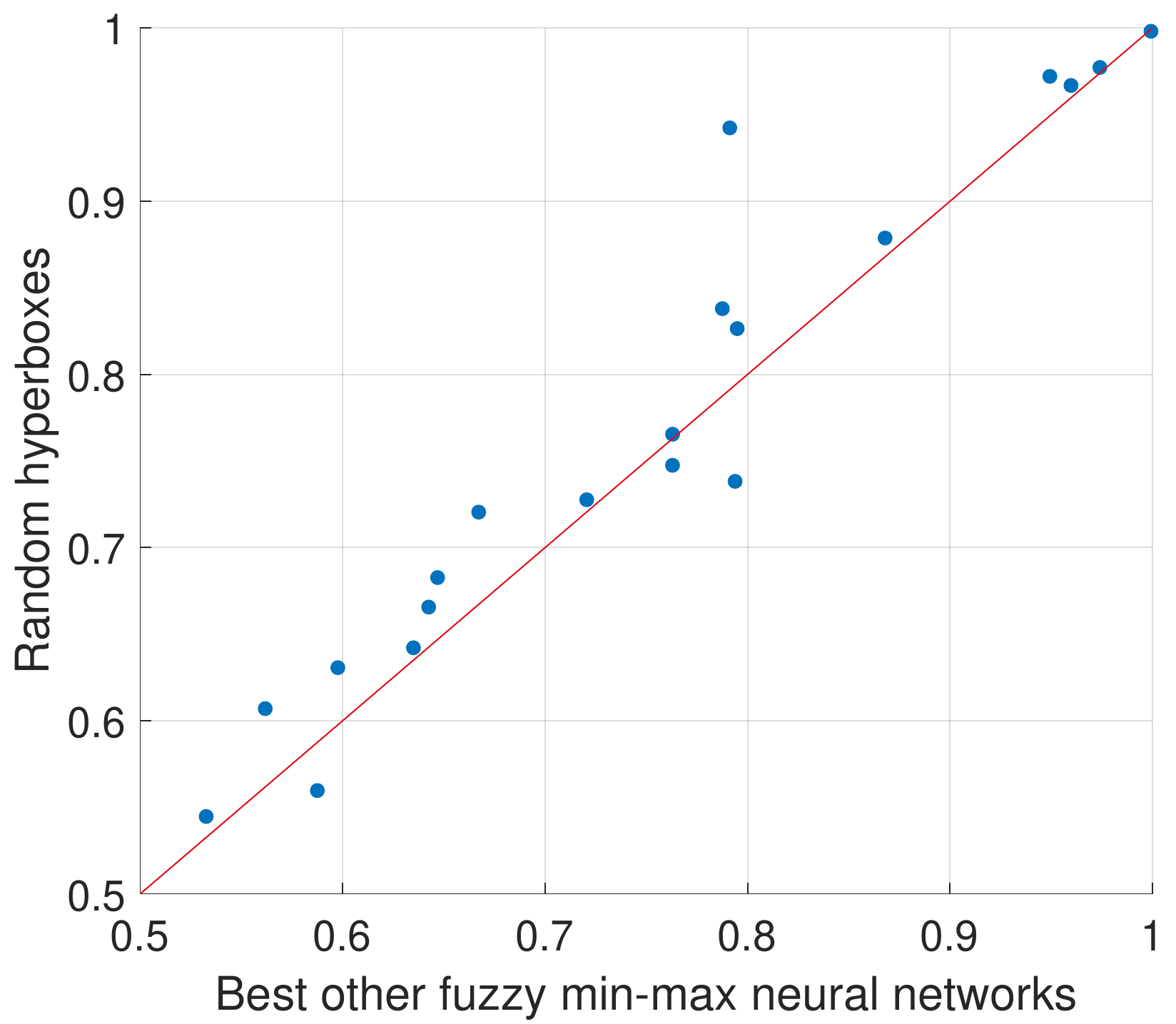}}
	\end{subfloat}
	\hfill%
	\begin{subfloat}[$\theta=0.7$]{
			\includegraphics[width=0.23\textwidth]{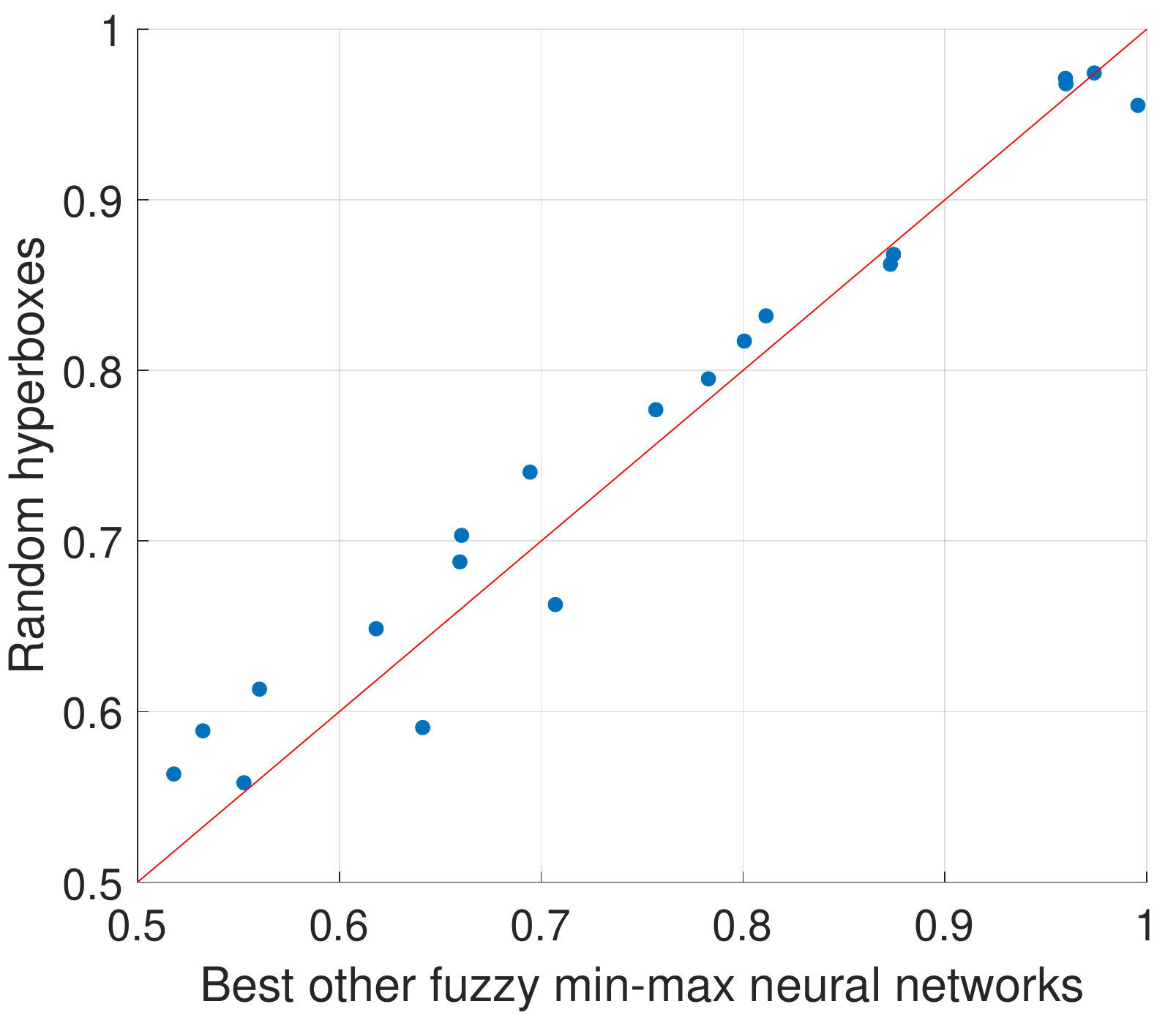}}
	\end{subfloat}
	\caption{Comparison of average weighted-F1 scores of the random hyperboxes and the best value from single FMNNs.}
	\label{fig5}
\end{figure}

\begin{figure}[!ht]
	\centering
	\includegraphics[width=0.4\textwidth]{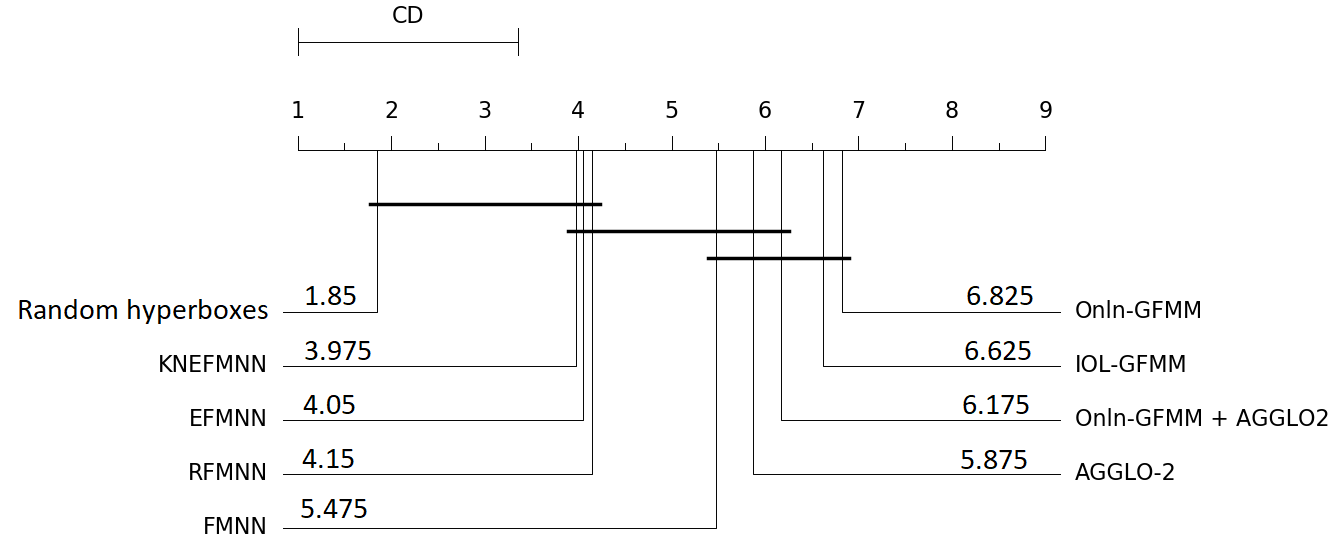}
	\caption{Critical difference diagram for the performance of the RH classifier and other FMNNs ($\theta=0.1$).}
	\label{fig6}
\end{figure}

Using the Friedman rank-sum test \cite{Friedman1940}, we can compute the F-distribution value $F_F = 10.1868$ from the average ranks of models at $\theta = 0.1$. Since the critical value of $F(8, 152)$ for the significance level $\alpha = 0.05$ is 1.9998, the null hypothesis is rejected. It means that there are significant differences between the average weighted-F1 scores of these models. To further compare the peformance of the RH model to other FMNNs at $\theta = 0.1$, the Critical Difference (CD) diagram with Bonferroni-Dunn test \cite{Demsar2006} for $\alpha = 0.05$ is computed and shown in Fig. \ref{fig6}.

\begin{figure}[!ht]
	\centering
	\includegraphics[width=0.38\textwidth]{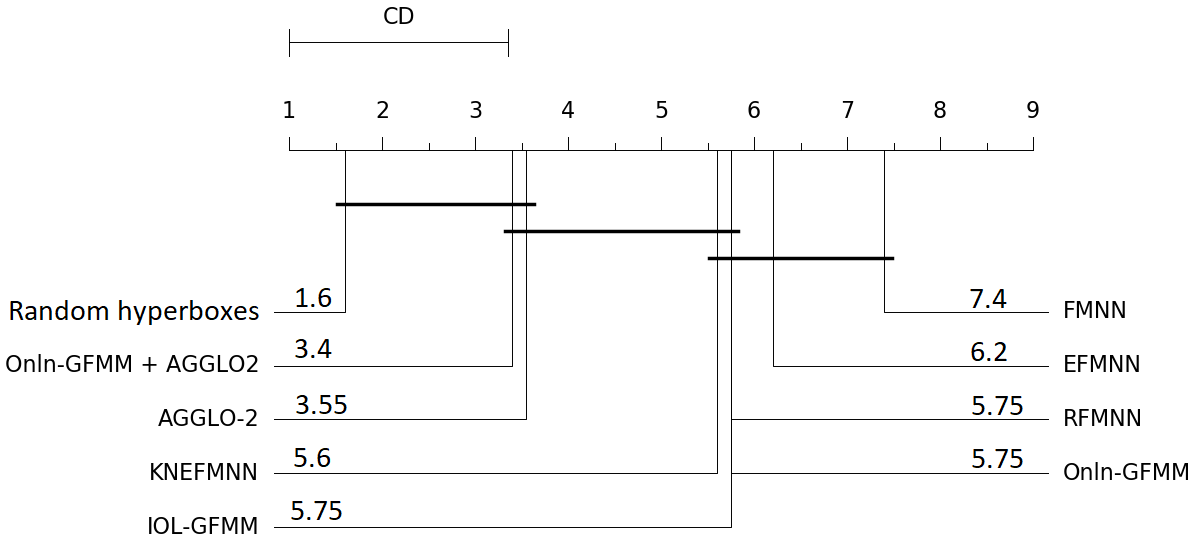}
	\caption{Critical difference diagram for the performance of the RH classifier and other FMNNs ($\theta=0.7$).}
	\label{fig7}
\end{figure}

Similarly, with results of average ranks at $\theta = 0.7$, we can calculate the F-distribution value using the Friedman test $F_F = 14.0148 > F(8, 152) = 1.9998$. Therefore, there are significant differences among models using $\theta = 0.7$. By applying the Bonferroni-Dunn test, we can draw the CD diagram shown in Fig. \ref{fig7}.

It can be seen that at the low value of $\theta$, the RH classifier is significantly better than Onln-GFMM, IOL-GFMM, FMNN, AGGLO-2, and Onln-GFMM + AGGLO2 in terms of average weighted-F1 score. However, its performance still has no significant difference compared to EFMNN, KNEFMNN, and RFMNN, although the average ranking of RH classifier is lowest among nine fuzzy min-max models over 20 considered datasets. With a high value of $\theta$, the RH model is significantly better than KNEFMNN, IOL-GFMM, Onln-GFMM, RFMNN, EFMNN, and FMNN. In this case, however, there is no statistical difference in the accuracy among the RH model, Onln-GFMM + AGGLO2 and AGGLO-2, although the performance of the RH classifier outperforms those of Onln-GFMM + AGGLO2 and AGGLO-2.

\subsubsection{A Comparison of the Random Hyperboxes With Other Ensemble Classifiers} \hfill

This experiment compares the perfomance of the random hyperboxes classifier with other prevalent ensemble models including Random Forest \cite{Breiman2001}, Rotation Forest \cite{Rodriguez2006}, XGBoost \cite{Chen2016}, LightGBM \cite{Ke2017}, Gradient Boosting \cite{Friedman2001}, and ensemble of base IOL-GFMM classifiers at the decision level (Ens-IOL-GFMM (DL)) and at the model level (Ens-IOL-GFMM (ML)) \cite{Gabrys2002a}.

\begin{figure}[!ht]
	\centering
	\includegraphics[width=0.45\textwidth]{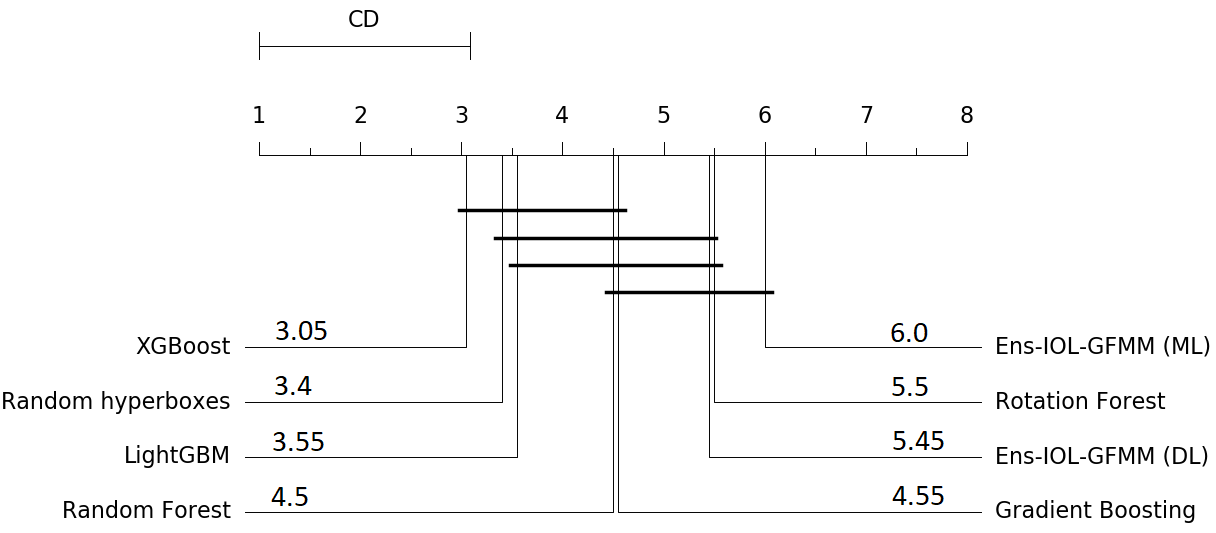}
	\caption{Critical difference diagram for the performance of the RH classifier and other ensemble models.}
	\label{fig8}
\end{figure}

The average weighted-F1 scores of classifiers through 10 times repeated 4-fold cross-validation and their ranking are given in Tables \ref{cmp_other_ensemble} and \ref{cmp_other_ensemble_ranking} in Appendix \ref{sup_cmp_other_ensemble}. Based on their average rank for 20 datasets, we can apply Friedman rank-sum test to calculate the F-distribution value $F_F = 4.7288 > F(7, 133) = 2.0791$. Therefore, there are differences in the performance of classifiers. Using the Bonferroni-Dunn test, we have the CD diagram of the RH model and other ensemble classifiers as Fig. \ref{fig8}.  

Although the average rank of over 20 datasets of the RH model is higher than XGBoost, there are no significant differences in the accuracy values among XGBoost, LightGBM, Random Forest, and Gradient Boosting. In contrast, the RH classifier is statistically better than Rotation Forest and ensemble methods of IOL-GFMM base learners using full features on 20 considered datasets.

\subsubsection{A Comparison of the Random Hyperboxes With Other Machine Learning Algorithms} \hfill

This experiment compares the RH classifier with other popular machine learning algorithms including Decision Tree \cite{Breiman1984}, Naive Bayes \cite{Zhang2004}, support vector machine (SVM) \cite{Suykens1999}, K-nearest neighbors (KNN) \cite{Altman1992}, and Linear Discriminant Analysis (LDA) \cite{Ye2007}. The experimental results of classifiers and their ranking are shown in Tables \ref{cmp_other_ml} and \ref{cmp_other_ml_rank} in Appendix \ref{sup_cmp_other_ml}.

\begin{figure}[!ht]
	\centering
	\includegraphics[width=0.4\textwidth]{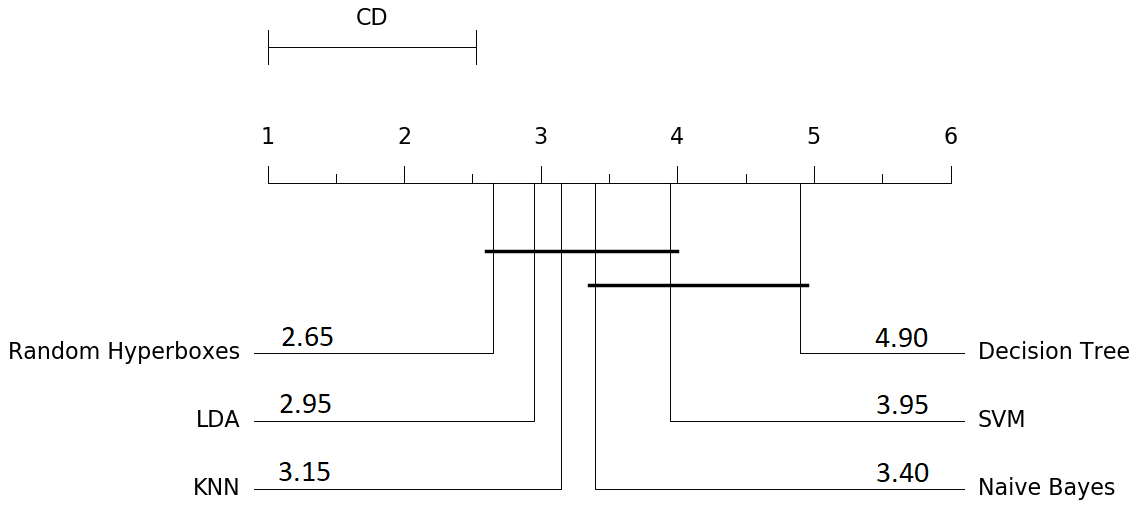}
	\caption{Critical difference diagram for the performance of the RH classifier and other popular learning algorithms.}
	\label{fig9}
\end{figure}

Using Friedman rank-sum test, we get the F-distribution value $F_F = 4.4485 > F(5, 95) = 2.3102$. Hence, there are statistical differences in the performance of classifiers. Similarly, using the Bonferroni-Dunn test, we obtain the CD diagram in this case as Fig. \ref{fig9}.

Although the average rank of the RH classifier over 20 datasets is lowest among methods, there is no significant difference in the performance of the RH compared to LDA, KNN, SVM, and Naive Bayes. However, the RH classifier is much better than the decision tree.

\begin{table*}
	\centering
	\caption{Estimated Upper Generalization Error Bounds, Real Testing Error, and Their Standard Deviations Computed From Different Assessment Methods}\label{estbound}
	\begin{scriptsize}
		\begin{tabular}{|c|l|c|c|c|c|}
			\hline
			\multirow{2}{*}{\textbf{ID}} & \multirow{2}{*}{\textbf{Dataset}} & \multicolumn{2}{c|}{\textbf{10 times repeated 4-fold cross-validation}} & \multicolumn{2}{c|}{\textbf{4-DPS fold cross-validation}} \\ \cline{3-6} 
			&                                   & \textbf{Testing error}     & \textbf{Estimated upper bound}    & \textbf{Testing error}  & \textbf{Estimated upper bound}  \\ \hline \hline
			1                            & Balance\_scale                    & 0.225205 $\pm$ 0.08439         & 0.47831 $\pm$ 0.042218                & 0.113598 $\pm$ 0.010918     & 0.406031 $\pm$ 0.040286             \\ \hline
			2                            & banknote\_authentication          & 0.001821 $\pm$ 0.001832        & 0.024073 $\pm$ 0.003918               & 0.001458 $\pm$ 0.002915     & 0.021189 $\pm$ 0.005162             \\ \hline
			3                            & blood\_transfusion                & 0.269997 $\pm$ 0.041152        & 0.882506 $\pm$ 0.086337               & 0.215241 $\pm$ 0.014064     & 0.88731 $\pm$ 0.092977              \\ \hline
			4                            & breast\_cancer\_wisconsin         & 0.033258 $\pm$ 0.018076        & 0.108871 $\pm$ 0.016842               & 0.028604 $\pm$ 0.00805      & 0.109176 $\pm$ 0.0185528            \\ \hline
			5                            & BreastCancerCoimbra               & 0.323276 $\pm$ 0.088239        & 0.099026 $\pm$ 0.011883               & 0.241379 $\pm$ 0.116086     & 0.117675 $\pm$ 0.022436             \\ \hline
			6                            & connectionist\_bench\_sonar       & 0.443689 $\pm$ 0.107638        & 0.069858 $\pm$ 0.011676               & 0.125 $\pm$ 0.036824        & 0.073405 $\pm$ 0.009768             \\ \hline
			7                            & haberman                          & 0.354808 $\pm$ 0.074583        & 0.601802 $\pm$ 0.064146               & 0.251581 $\pm$ 0.015168     & 0.529377 $\pm$ 0.071196             \\ \hline
			8      & heart                             & 0.170758 $\pm$ 0.024509        & 0.199591 $\pm$ 0.021899               & 0.174056 $\pm$ 0.024932     & 0.185977 $\pm$ 0.016658             \\ \hline
			9                            & movement\_libras                  & 0.415556 $\pm$ 0.097833        & 0.102645 $\pm$ 0.0173813              & 0.136111 $\pm$ 0.042913     & 0.147793 $\pm$ 0.022053             \\ \hline
			10                           & pima\_diabetes                    & 0.257552 $\pm$ 0.02993         & 0.269897 $\pm$ 0.023865               & 0.239583 $\pm$ 0.020395     & 0.260163 $\pm$ 0.019503             \\ \hline
			11                           & plant\_species\_leaves\_margin    & 0.242875 $\pm$ 0.018245        & 0.128801 $\pm$ 0.010176               & 0.226875 $\pm$ 0.031516     & 0.118236 $\pm$ 0.008702             \\ \hline
			12                           & plant\_species\_leaves\_shape     & 0.37025 $\pm$ 0.02875          & 0.171654 $\pm$ 0.007884               & 0.34125 $\pm$ 0.040337      & 0.184498 $\pm$ 0.005001             \\ \hline
			13                           & ringnorm                          & 0.059649 $\pm$ 0.006073        & 0.048538 $\pm$ 0.002107               & 0.073514 $\pm$ 0.005635     & 0.05311 $\pm$ 0.001684              \\ \hline
			14                           & landsat\_satelite                 & 0.116943 $\pm$ 0.006915        & 0.181342 $\pm$ 0.015345               & 0.104273 $\pm$ 0.004502     & 0.181983 $\pm$ 0.019838             \\ \hline
			15                           & twonorm                           & 0.027892 $\pm$ 0.003177        & 0.037681 $\pm$ 0.000687               & 0.029189 $\pm$ 0.002457     & 0.038032 $\pm$ 0.001046             \\ \hline
			16                           & vehicle\_silhouettes              & 0.267981 $\pm$ 0.028846        & 0.206567 $\pm$ 0.013313               & 0.251716 $\pm$ 0.028308     & 0.205494 $\pm$ 0.008012             \\ \hline
			17                           & vertebral\_column                 & 0.229125 $\pm$ 0.0483          & 0.125849 $\pm$ 0.009933               & 0.209665 $\pm$ 0.051705     & 0.147310 $\pm$ 0.014369             \\ \hline
			18                           & vowel                             & 0.363582 $\pm$ 0.06567         & 0.047757 $\pm$ 0.002076               & 0.023247 $\pm$ 0.011655     & 0.054667 $\pm$ 0.001788             \\ \hline
			19                           & waveform                          & 0.158041 $\pm$ 0.006471        & 0.087998 $\pm$ 0.003282               & 0.1636 $\pm$ 0.006804       & 0.089598 $\pm$ 0.005328             \\ \hline
			20                           & wireless\_indoor\_localization    & 0.02275 $\pm$ 0.008566         & 0.098253 $\pm$ 0.014660               & 0.0155 $\pm$ 0.004435       & 0.089935 $\pm$ 0.003942             \\ \hline
		\end{tabular}
	\end{scriptsize}
\end{table*}

\section{On the Estimation of Generalization Error Bounds and Open Problems}\label{discussion_bound}
The upper generalization error bound of the random hyperboxes model is computed based on the i.i.d. assumption of samples in both training and testing sets. However, in practice, this assumption is usually violated for the real world datasets. This means that it is very difficult to obtain the training and testing sets which are representatives of a true distribution of the sample space. In this section, we will estimate the upper generalization error bounds of datasets used for the experiments in section \ref{experiment}. The purpose of this section is to identify the effectiveness of the upper generalization error bound on real datasets and the existing problems when applying a strong assumption from the theoretical derivations to the practical issues. The upper bound values were estimated from the training set and 100 base learners trained by the IOL-GFMM algorithm with $\theta = 0.1$. The estimated results of the upper generalization error bound are the average values from 40 iterations (10 times repeated 4-fold cross-validation). To strengthen the comparison and conclusion, we also estimated the upper generalization error bounds from the base learners trained in turn on each of four folds generated by using the density preserving sampling (DPS) method \cite{Budka2013}. The DPS method aims to preserve the data density and the classes shapes when splitting an original dataset into many folds, so it is possible to create the testing sets which are representatives for the training data. Hence, the testing errors on the DPS folds are usually smaller than those calculated from folds of the cross-validation method. This fact is confirmed with the results shown in Table \ref{estbound}. This table presents the real average testing errors of 4-DPS fold cross-validation and 10 times repeated 4-fold cross-validation as well as their upper generalization error bounds estimated from corresponding training sets. 

In general, we have ten datasets in which the estimated upper bounds are higher than real testing errors. Among them, there are a number of datasets with real errors close to the estimated upper bounds, such as $heart$, $pima\_diabetes$, $landsat\_satelite$, and $twonorm$. One explanation for these good estimations is that the training sets and testing sets are good representatives of each other and the whole sample space. It can be seen that, for these datasets, the real testing errors of 10 times repeated 4-fold cross-validation and 4-DPS fold cross-validation are relatively close to each other.

In the ten remaining datasets, the estimated values of upper bounds are much lower than the real testing errors when applying the 10 times repeated 4-fold cross-validation method. The same behavior but with a smaller error can be found with the 4-DPS fold cross-validation method on eight datasets. Interestingly, there are two datasets, $wovel$ and $movement\_libras$, in which the estimated values are very bad when using 10 times repeated 4-fold cross-validation, but we can obtain very good estimated upper bounds when deploying the 4-DPS fold cross-validation. This fact indicates that if the representativeness of training sets with regard to the whole sample space is good, we can achieve a much better estimation of the upper generalization error bounds which is close to the testing error on unseen data with the same distribution.

\begin{figure}[!ht]
	\centering
	\includegraphics[width=0.4\textwidth]{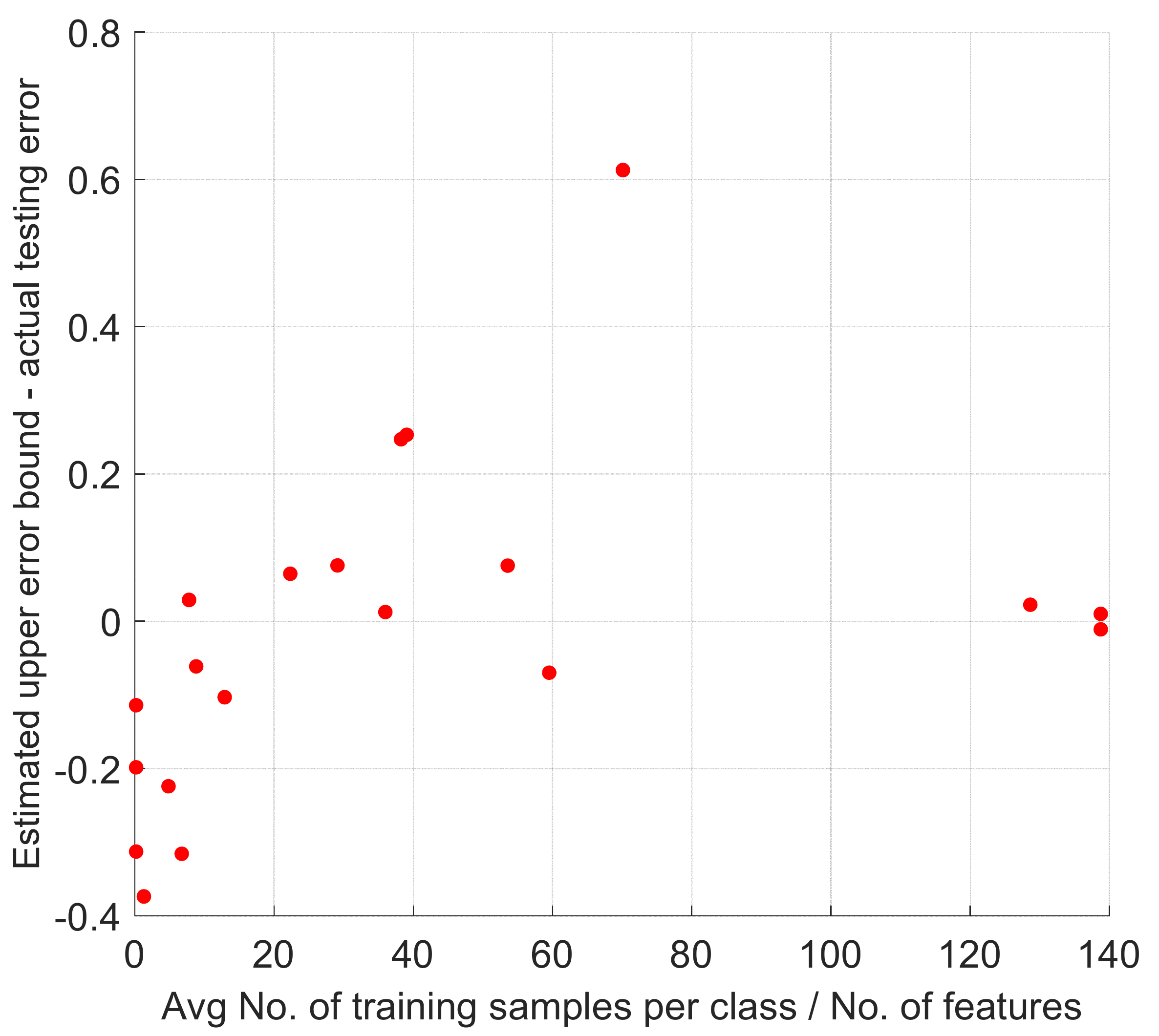}
	\caption{The corelation of the difference in the estimated upper error bound and actual testing error with respect to the ratio of the average number of training samples per class and the number of features.}
	\label{fig10}
\end{figure}

One general characteristic of datasets resulting in the poor estimated upper bounds is their sparsity with regard to a small number of samples and a relatively high number of dimensions. For these datasets, we do not have sufficient number of samples to accurately enough capture the underlying distribution of the whole sample space. As a result, the base estimators overfit with their training data, and the estimated values of the upper error bounds are usually small. Meanwhile, the testing errors on unseen data are fairly high. Here, one open problem identified is the relationship between the number of samples, classes, and dimensions so that we can obtain a good estimation of the generalization error bounds from the training data. This is a critical issue that needs to be tackled in future work. As an example demonstration for this issue, Fig. \ref{fig10} shows the correlation of the difference in the estimated upper error bound and actual testing error to the ratio of the average training samples per class and the number of features for 20 datasets used in this experiment. We can see that a good estimation of the upper error bound can be obtained if the ratio of the average training samples per class and the number of features is larger than 20. If this ratio is higher than 120, it is more likely to achieve an estimated upper error bound close to the actual testing error.

In summary, the i.i.d. assumption of training and testing sets is usually not met in practical datasets. Therefore, to reduce the classification error on unseen data, we need to use several methods to guarantee the representativeness of the training and testing sets when assessing the performance of models. Moreover, identification of the relationship between the numbers of samples, classes, and features is crucial to building a representative training set.

One of the strong points of the general fuzzy min-max neural network is the interpretability. However, the significantly improved predictive accuracy of the proposed random hyperboxes method comes at a price of loss of interpretability as is common with other ensemble methods. As previously shown in \cite{Gabrys2002a}, hyperbox representation allows for combination at the model level rather than the decision level and therefore retaining the interpretability of the final model. Nonetheless, the combination of the individual hyperbox-based learners which are built from different random subspaces of features is not a trivial problem. Therefore, the future study should focus on building interpretable random hyperboxes models.

\section{Conclusion and Future Work} \label{conclusion}
This paper proposed a novel random hyperboxes classifier, discussed its properties and provided derivations of its generalization error bounds. The experimental results confirmed the efficiency of the proposed method in comparison to other single fuzzy min-max neural networks as well as single learning algorithms. The random hyperboxes model is also competitive with other popular ensemble methods. Furthermore, we provided several discussion on the estimation of the upper generalization error bounds for real-world datasets, and identified some open issues for future work.

There are still many opportunities for improvement of the proposed classifier. The relationship between correlation and variance between base learners as well as the trade-off between variance and bias of the random hyperboxes model need to be analyzed in more details. In addition, the influence of hyperparameters of the random hyperboxes model should be assessed by a comparative study. In this paper, we assumed that the strength $\mathcal{S} > 0$ when analyzing the generalization error bound. In the case of highly imbalanced classes, this assumption may be false because the strength usually focuses on the majority class. Therefore, the efficiency of the random hyperboxes classifier and its theoretical results should be investigated and extended for imbalanced datasets.

\section*{Acknowledgment}
T.T. Khuat acknowledges FEIT-UTS for awarding his PhD scholarships (IRS and FEIT scholarships).


\onecolumn
\newpage

\appendices
\section{Proof of Lemma 1}
\label{prooflemma1}
This section provides the readers with the proof of Lemma 1 in the main paper.

\begin{lemma}
	Given $m$ identically distributed random variables (not necessarily independent) with the variance of each variable $\sigma^2$ and positive pairwise correlation $\mathbf{\rho}$, the variance of the average random variable is: \\
	\begin{equation*}
	\rho \cdot \sigma^2 + \cfrac{1 - \rho}{m} \cdot \sigma^2
	\end{equation*}
\end{lemma}
\begin{proof}
	Supposing that $\Phi = (\Phi_1,\ldots, \Phi_m)$ is a set of $m$ random variables with given covariances $\sigma_{ij} = \mathtt{Cov}(\Phi_i, \Phi_j) $, we need to find variance of an average variable $\mathcal{L}(\Phi_1,\ldots,\Phi_m)$ obtained as a linear combination of $m$ random variables, i.e., $$\mathcal{L}(\Phi_1,\ldots,\Phi_m) = \sum_{i=1}^m(\lambda_i \cdot \Phi_i)$$ We can rewrite this formula in a compact way using matrix and vector notations as follows: $$\mathcal{L}(\Phi) = \mathbf{\Lambda}^T \cdot \Phi $$ where $\mathbf{\Lambda}^T = (\lambda_1,\ldots,\lambda_m)$. And then, we have the expected value: $$\mathbb{E}(\mathcal{L}(\Phi)) = \mathbb{E}(\mathbf{\Lambda}^T \cdot \Phi) = \mathbf{\Lambda}^T \cdot \mathbb{E}(\Phi)  $$
	and the variance:
	\begin{align*} \mathtt{Var}(\mathcal{L}(\Phi))
	&= \mathbb{E}(\mathcal{L}^2(\Phi)) - [\mathbb{E}(\mathcal{L}(\Phi))]^2 \\
	&=\mathbb{E}(\mathbf{\Lambda}^T \Phi \Phi^T \mathbf{\Lambda}) - \mathbb{E}(\mathbf{\Lambda}^T \Phi) [\mathbb{E}(\mathbf{\Lambda}^T \Phi)]^T \\
	&= \mathbf{\Lambda}^T \mathbb{E}(\Phi \Phi^T) \mathbf{\Lambda} - \mathbf{\Lambda}^T \mathbb{E}(\Phi) (\mathbb{E}(\Phi))^T \mathbf{\Lambda} \\
	&=\mathbf{\Lambda}^T [\mathbb{E}(\Phi \Phi^T) - \mathbb{E}(\Phi) (\mathbb{E}(\Phi))^T] \mathbf{\Lambda} \\
	&= \mathbf{\Lambda}^T \mathtt{Cov}(\Phi) \mathbf{\Lambda}\\
	&= \mathbf{\Lambda}^T \Sigma \mathbf{\Lambda}
	\end{align*}
	where $\Sigma = (\sigma_{ij})$ is the covariance of $\Phi$
	
	In this lemma, $\sigma_{ij} = \rho \cdot \sigma^2$ when $i \ne j$. We also have $\sigma_{ii} = \mathtt{Cov}(\Phi_i, \Phi_i) = \sigma^2 = [\rho + (1-\rho)] \sigma^2$. Hence, we may decompose the covariance matrix $\Sigma$ into the sum of two matrices, i.e., one includes $\rho$ in every entry and the other includes $(1 - \rho)$ on the main diagonal and zeros for the rest. Formally, we achieve:
	$$\Sigma = \sigma^2 [\rho \mathbf{1}_m \mathbf{1}_m^T + (1 - \rho) \mathbf{I}_m]$$ \\
	where $\mathbf{1}_m$ is a column vector containing $m$ 1's and $\mathbf{I}_m$ is an identity matrix with size $m \times m$. Then we get:
	\begin{align*}
	\mathtt{Var}(\mathcal{L}(\Phi)) &= \mathbf{\Lambda}^T \sigma^2 [\rho \mathbf{1}_m \mathbf{1}_m^T + (1 - \rho) \mathbf{I}_m] \mathbf{\Lambda} \\
	&= (\mathbf{\Lambda}^T \mathbf{1}_m \mathbf{1}_m^T \mathbf{\Lambda}) \rho \sigma^2 + (\mathbf{\Lambda}^T \mathbf{I}_m \mathbf{\Lambda}) (1 - \rho) \sigma^2
	\end{align*}
	For $\mathbf{\Lambda}^T = (1/m,\ldots,1/m)$, we get:
	$$\mathbf{\Lambda}^T \mathbf{1}_m \mathbf{1}_m^T \mathbf{\Lambda} = (\mathbf{\Lambda}^T \mathbf{1}_m)^2 = (m \cdot 1/m)^2 = 1$$
	and \\
	$$\mathbf{\Lambda}^T \mathbf{I}_m \mathbf{\Lambda} = 1/m^2 + \ldots + 1/m^2 = m \cdot 1/m^2 = 1/m$$
	Therefore, $$\mathtt{Var}(\mathcal{L}(\Phi)) = \rho \sigma^2 + \cfrac{1 - \rho}{m} \sigma^2$$
	
	The lemma is proved.
\end{proof}

\section{Proof of Lemma 2}\label{prooflemma2}
This section provides the proof of Lemma 2 in the main paper.
\begin{lemma}
	When the number of base estimators increases $(m \rightarrow \infty)$ and base estimators are independent, for almost surely all i.i.d. random vectors $\Phi_1, \Phi_2, \ldots$, the margin function for a random hyperboxes model $\mathcal{M}(\mathbf{x}, c)$ at each input $\mathbf{x}$ converges to:
	\begin{equation*}
	\mathcal{M}^*(\mathbf{x}, c) =  \mathbf{P}_\Phi(h(\mathbf{x}, \Phi) = c) - \max_{j \ne c} \mathbf{P}_\Phi(h(\mathbf{x}, \Phi) = j)
	\end{equation*}
\end{lemma}
\begin{proof}
	We have the margin function of the random hyperboxes model with $m$ base learners at each input sample $\mathbf{x}$ as follows:
	\begin{equation*}
	\mathcal{M}(\mathbf{x}, c) = \cfrac{1}{m} \sum_{i=1}^m{\mathbb{1}(h_i(\mathbf{x}) = c)} - \max_{j \ne c}\cfrac{1}{m} \sum_{i=1}^m{\mathbb{1}(h_i(\mathbf{x}) = j)}
	\end{equation*}
	
	For random vectors $\Phi_1, \Phi_2, \ldots$ and for all input vectors $\mathbf{x}$, to prove Lemma \ref{lemma2}, it suffices to show
	\begin{equation*}
	\cfrac{1}{m} \sum_{i=1}^m{\mathbb{1}(h_i(\mathbf{x}) = j)} \xrightarrow{m \rightarrow \infty} \mathbf{P}_{\Phi}(h(\mathbf{x}, \Phi) = j)
	\end{equation*}
	where $h_i(\mathbf{x}) \equiv h(\mathbf{x}, \Phi_i)$, and $\mathbb{1}(\cdot)$ is an indicator function.
	
	For each hyperbox-based learner, $h(\mathbf{x}, \Phi_i) = j$ is union of hyerboxes with class $j$ and their neighborhood regions which generate the maximum membership value from these hyperboxes to an input $\mathbf{x}$ in comparison to hyperboxes representing other classes. Assuming a finite number of random vectors $\Phi$ (the finite number of sample subsets and finite number of feature subsets) from which any hyperbox-based learner $h(\mathbf{x}, \Phi_i)$ ($\Phi_i \subset \Phi$) is constructed, then there exists a finite number $K$ of such unions of hyperboxes and neighbourhood regions, called $S_1, \ldots, S_K$.
	
	Let define:
	\begin{equation*}
	\varphi(\Phi) = k \mbox{ if } \{\mathbf{x}: h(\mathbf{x}, \Phi) = j \} = S_k
	\end{equation*}
	
	Let $N_k$ be the number of times that $\varphi(\Phi_i) = k$ in the first $m$ trials, then we obtain:
	\begin{equation*}
	\cfrac{1}{m} \sum_{i=1}^m{\mathbb{1}(h(\mathbf{x}, \Phi_i) = j)} = \cfrac{1}{m} \sum_{k} N_k \mathbb{1}(\mathbf{x} \in S_k)
	\end{equation*}
	
	According to the strong law of large numbers when $m$ increases, $$N_k = \cfrac{1}{m} \sum_{i=1}^m \mathbb{1}(\varphi(\Phi_i) = k)$$ converges almost surely (a.s.) with probability 1 to $$\mathbb{E}_\Phi [\mathbb{1}(\varphi(\Phi) = k)] = \mathbf{P}_\Phi(\varphi(\Phi) = k)$$ Therefore,
	\begin{equation*}
	\begin{split}
	\cfrac{1}{m} \sum_{i=1}^m{\mathbb{1}(h(\mathbf{x}, \Phi_i) = j)} \xrightarrow{a.s.} &\sum_{k} \mathbf{P}_{\Phi}(\varphi(\Phi) = k) \mathbb{1}(\mathbf{x} \in S_k) \\
	& = \mathbf{P}_{\Phi}(h(\mathbf{x}, \Phi) = j)
	\end{split}
	\end{equation*}
	
	The lemma is proved.
\end{proof}

\section{Proof of Theorem 2}\label{prooftheorem2}
This section shows the proof for Theorem 2 from the main paper.
\setcounter{theorem}{1}
\begin{theorem}
	An upper bound of the generalization error for the random hyperboxes model can be estimated from the strength of base learners and correlation between base learners as follows:
	\begin{equation*}
	\mathcal{E}^* \leq \overline{\rho} \; \Bigl(\cfrac{1}{\mathcal{S}^2} - 1 \Bigr)
	\end{equation*}
\end{theorem}
\begin{proof}
	From lemma \ref{lemma2}, we have:
	\begin{equation*}
	\mathcal{M}^*(\mathbf{x}, c) =  \mathbf{P}_\Phi(h(\mathbf{x}, \Phi) = c) - \max_{j \ne c} \mathbf{P}_\Phi(h(\mathbf{x}, \Phi) = j)
	\end{equation*}
	With the assumption of the strength $\mathcal{S} = \mathbb{E}_{\mathbf{X}, \mathcal{C}} \mathcal{M}^*(\mathbf{x}, c) > 0$, according to Chebyshev's inequality, we have:
	\begin{equation*}
	\begin{split}
	\mathcal{E}^* &= \mathbf{P}_{\mathbf{X}, \mathcal{C}}\left[\mathcal{M}^*(\mathbf{x}, c) < 0\right] \leq \mathbf{P}_{\mathbf{X}, \mathcal{C}}\left[\mathcal{S} - \mathcal{M}^*(\mathbf{x}, c) \geq \mathcal{S}\right]\\
	&= \mathbf{P}_{\mathbf{X}, \mathcal{C}}\left[|\mathcal{M}^*(\mathbf{x}, c) - \mathcal{S}| \geq \mathcal{S}\right] \leq \cfrac{\mathtt{Var}_{\mathbf{X}, \mathcal{C}}(\mathcal{M}^*(\mathbf{x}, c))}{\mathcal{S}^2}
	\end{split}
	\end{equation*}
	
	For any function $f$ and two i.i.d. random variables $\Phi$ and $\Phi'$, we have:
	\begin{equation*}
	\mathbb{E}_\Phi [f(\Phi)]^2 = \mathbb{E}_{\Phi, \Phi'} [f(\Phi) f(\Phi')]
	\end{equation*}
	In the main paper, we get $\mathcal{M}^*(\mathbf{x}, c) = \mathbb{E}_\Phi \mathcal{R}(\Phi)$, thus
	\begin{equation*}
	[\mathcal{M}^*(\mathbf{x}, c)]^2 = \mathbb{E}_\Phi \mathcal{R}(\Phi)^2 = \mathbb{E}_{\Phi, \Phi'} [\mathcal{R}(\Phi) \mathcal{R}(\Phi')]
	\end{equation*}
	
	Now, we can compute $\mathtt{Var}_{\mathbf{X}, \mathcal{C}}(\mathcal{M}^*(\mathbf{x}, c))$ as follows:
	\begin{equation*}
	\begin{split}
	\mathtt{Var}_{\mathbf{X},\mathcal{C}}(\mathcal{M}^*(\mathbf{x}, c)) &= \mathbb{E}_{\mathbf{X},\mathcal{C}}([\mathcal{M}^*(\mathbf{x}, c)]^2) - \Bigl[\mathbb{E}_{\mathbf{X},\mathcal{C}}(\mathcal{M}^*(\mathbf{x}, c))\Bigr]^2   \\
	&= \mathbb{E}_{\mathbf{X},\mathcal{C}}\Bigl[ \mathbb{E}_{\Phi, \Phi'} [\mathcal{R}(\Phi) \mathcal{R}(\Phi')] \Bigr] - \Bigl[\mathbb{E}_{\mathbf{X},\mathcal{C}}(\mathbb{E}_\Phi \mathcal{R}(\Phi))\Bigr]^2 \\
	&= \mathbb{E}_{\Phi, \Phi'}\Bigl[ \mathbb{E}_{\mathbf{X},\mathcal{C}} [\mathcal{R}(\Phi) \mathcal{R}(\Phi')] \Bigr] - \Bigl[\mathbb{E}_{\Phi}(\mathbb{E}_{\mathbf{X},\mathcal{C}} \mathcal{R}(\Phi))\Bigr]^2\\
	&= \mathbb{E}_{\Phi, \Phi'}\Bigl[ \mathbb{E}_{\mathbf{X},\mathcal{C}} [\mathcal{R}(\Phi) \mathcal{R}(\Phi')] \Bigr] - \mathbb{E}_{\Phi, \Phi'}\Bigl[\mathbb{E}_{\mathbf{X},\mathcal{C}} \mathcal{R}(\Phi) \mathbb{E}_{\mathbf{X},\mathcal{C}} \mathcal{R}(\Phi')\Bigr] \\
	&= \mathbb{E}_{\Phi, \Phi'}\Bigl[ \mathbb{E}_{\mathbf{X},\mathcal{C}} [\mathcal{R}(\Phi) \mathcal{R}(\Phi')] - \mathbb{E}_{\mathbf{X},\mathcal{C}} \mathcal{R}(\Phi) \mathbb{E}_{\mathbf{X},\mathcal{C}} \mathcal{R}(\Phi')\Bigr] \\
	&= \mathbb{E}_{\Phi, \Phi'}\Bigl[ \mathtt{Cov}_{\mathbf{X}, \mathcal{C}} (\mathcal{R}(\Phi) \mathcal{R}(\Phi')) \Bigr]\\
	&= \mathbb{E}_{\Phi, \Phi'}\Bigl[ \rho_{\mathbf{X}, \mathcal{C}}(\Phi, \Phi') \sigma_{\mathbf{X}, \mathcal{C}}(\mathcal{R}(\Phi)) \sigma_{\mathbf{X}, \mathcal{C}}(\mathcal{R}(\Phi')) \Bigr]\\
	&= \overline{\rho} \Bigl[ \mathbb{E}_{\Phi}(\sigma_{\mathbf{X}, \mathcal{C}}(\mathcal{R}(\Phi))) \Bigr]^2
	\end{split}
	\end{equation*}
	where $\overline{\rho} = \mathbb{E}_{\Phi, \Phi'} [\rho_{\mathbf{X}, \mathcal{C}}(\Phi, \Phi')]$
	
	For any random variable $\mathbf{Z}$, $\mathtt{Var}(\mathbf{Z}) \geq 0 \Rightarrow \mathbb{E}(\mathbf{Z}^2) - \mathbb{E}(\mathbf{Z})^2 \geq 0 \Rightarrow \mathbb{E}(\mathbf{Z})^2 \leq \mathbb{E}(\mathbf{Z}^2)$. Therefore,
	\begin{equation*}
	\mathtt{Var}_{\mathbf{X},\mathcal{C}}(\mathcal{M}^*(\mathbf{x}, c)) = \overline{\rho} \Bigl[ \mathbb{E}_{\Phi}(\sigma_{\mathbf{X}, \mathcal{C}}(\mathcal{R}(\Phi))) \Bigr]^2 \leq \overline{\rho} \; \mathbb{E}_{\Phi}(\sigma_{\mathbf{X}, \mathcal{C}} (\mathcal{R}(\Phi))^2) = \overline{\rho} \; \mathbb{E}_{\Phi}(\mathtt{Var}_{\mathbf{X}, \mathcal{C}}(\mathcal{R}(\Phi)))
	\end{equation*}
	In addition, using the definition of the variance for a random variable and inequality $\mathbb{E}(\mathbf{Z})^2 \leq \mathbb{E}(\mathbf{Z}^2)$, we can write:
	\begin{equation*}
	\begin{split}
	\mathbb{E}_{\Phi}(\mathtt{Var}_{\mathbf{X}, \mathcal{C}}(\mathcal{R}(\Phi))) &= \mathbb{E}_{\Phi} \Bigl[ \mathbb{E}_{\mathbf{X}, \mathcal{C}}[\mathcal{R}(\Phi)^2] - \mathbb{E}_{\mathbf{X}, \mathcal{C}}[\mathcal{R}(\Phi)]^2 \Bigr] \\
	&= \mathbb{E}_{\Phi} \Bigl[ \mathbb{E}_{\mathbf{X}, \mathcal{C}}[\mathcal{R}(\Phi)^2] \Bigr] - \mathbb{E}_\Phi \Bigl[ [\mathbb{E}_{\mathbf{X}, \mathcal{C}}\mathcal{R}(\Phi)]^2 \Bigr] \\
	&\leq \mathbb{E}_{\Phi} \Bigl[ \mathbb{E}_{\mathbf{X}, \mathcal{C}}[\mathcal{R}(\Phi)^2] \Bigr] - \Bigl[ \mathbb{E}_\Phi( \mathbb{E}_{\mathbf{X}, \mathcal{C}}[\mathcal{R}(\Phi)]) \Bigr]^2 \\
	&= \mathbb{E}_{\Phi} \Bigl[ \mathbb{E}_{\mathbf{X}, \mathcal{C}}[\mathcal{R}(\Phi)^2] \Bigr] - \Bigl[ \mathbb{E}_{\mathbf{X}, \mathcal{C}}( \mathbb{E}_\Phi[\mathcal{R}(\Phi)]) \Bigr]^2 \\
	&= \mathbb{E}_{\Phi} \Bigl[ \mathbb{E}_{\mathbf{X}, \mathcal{C}}[\mathcal{R}(\Phi)^2] \Bigr] - \Bigl[ \mathbb{E}_{\mathbf{X}, \mathcal{C}} \mathcal{M}^*(\mathbf{x}, c) \Bigr]^2 \\
	&\leq 1 - \mathcal{S}^2
	\end{split}
	\end{equation*}
	due to $\mathcal{R}(\Phi) \leq 1$ and $\mathcal{S} = \mathbb{E}_{\mathbf{X}, \mathcal{C}} \mathcal{M}^*(\mathbf{x}, c)$. As a result,
	\begin{equation*}
	\mathcal{E}^* \leq \cfrac{\mathtt{Var}_{\mathbf{X}, \mathcal{C}}(\mathcal{M}^*(\mathbf{x}, c))}{\mathcal{S}^2} \leq \cfrac{\overline{\rho} \; \mathbb{E}_{\Phi}(\mathtt{Var}_{\mathbf{X}, \mathcal{C}}(\mathcal{R}(\Phi)))}{\mathcal{S}^2} \leq \cfrac{\overline{\rho} \; (1 - \mathcal{S}^2)}{\mathcal{S}^2} = \overline{\rho} \; \Bigl(\cfrac{1}{\mathcal{S}^2} - 1 \Bigr)
	\end{equation*}
	
	The theorem is proved.
\end{proof}

\section{Additional Experimental Results}
\subsection{Supplementary Part for Analyzing the Variance of the Random Hyperboxes Classifier}\label{sup_variance}
\begin{figure}[!ht]
	\begin{subfloat}[Plant\_species\_leaves\_margin]{
			\includegraphics[width=0.32\textwidth, height=0.18\textheight]{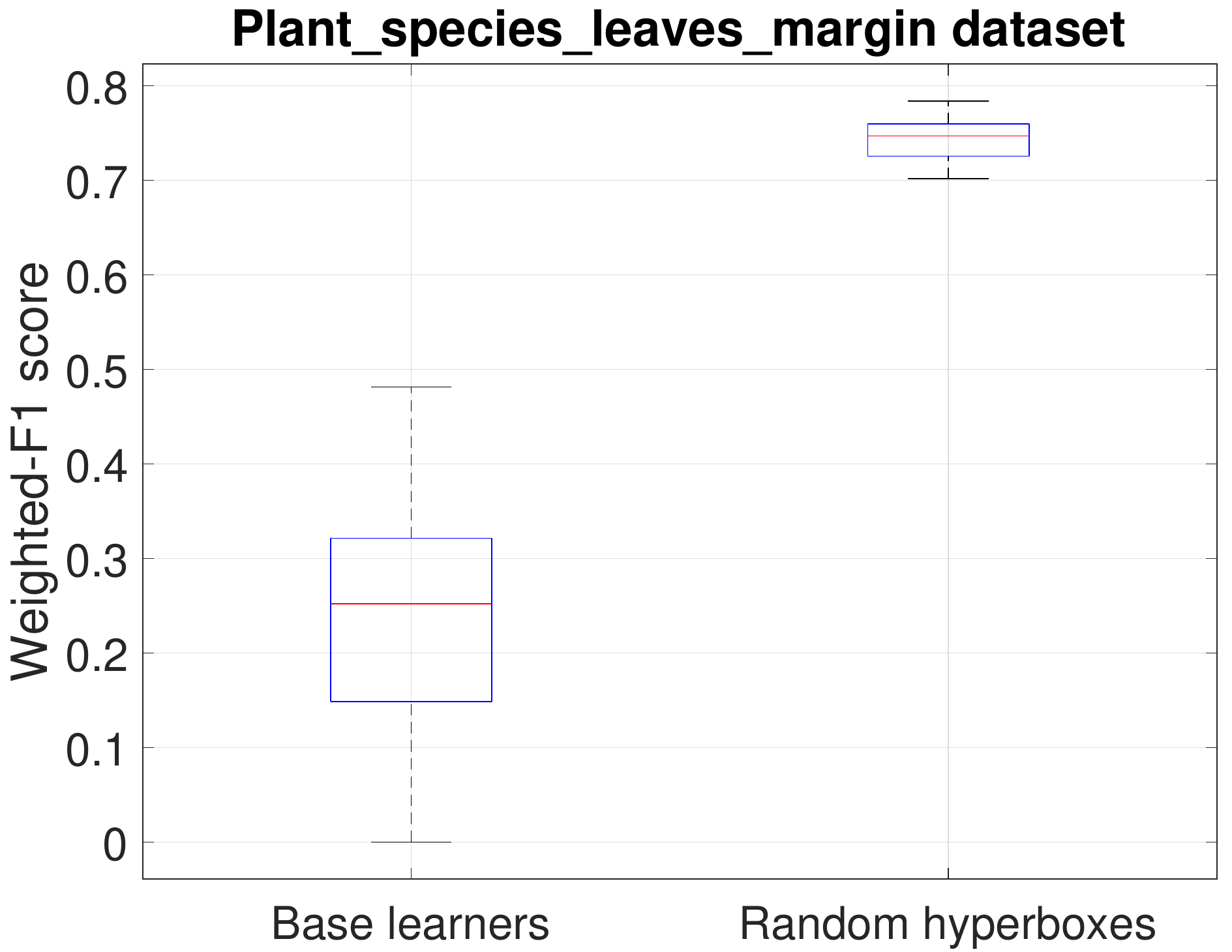}}
	\end{subfloat}
	\hfill%
	\begin{subfloat}[Plant\_species\_leaves\_shape]{
			\includegraphics[width=0.32\textwidth,height=0.18\textheight]{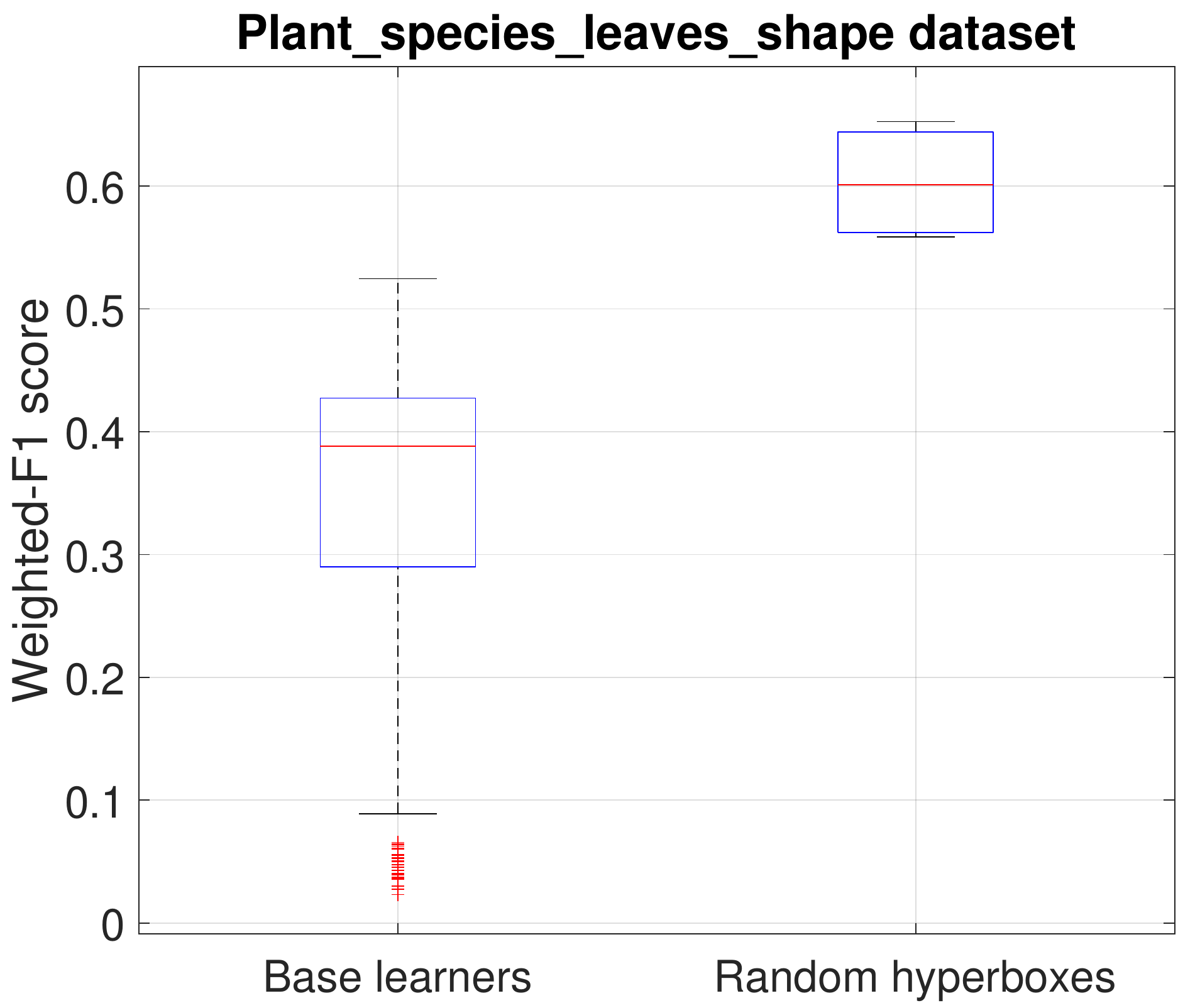}}
	\end{subfloat}
	\hfill%
	\begin{subfloat}[Heart]{
			\includegraphics[width=0.32\textwidth,height=0.18\textheight]{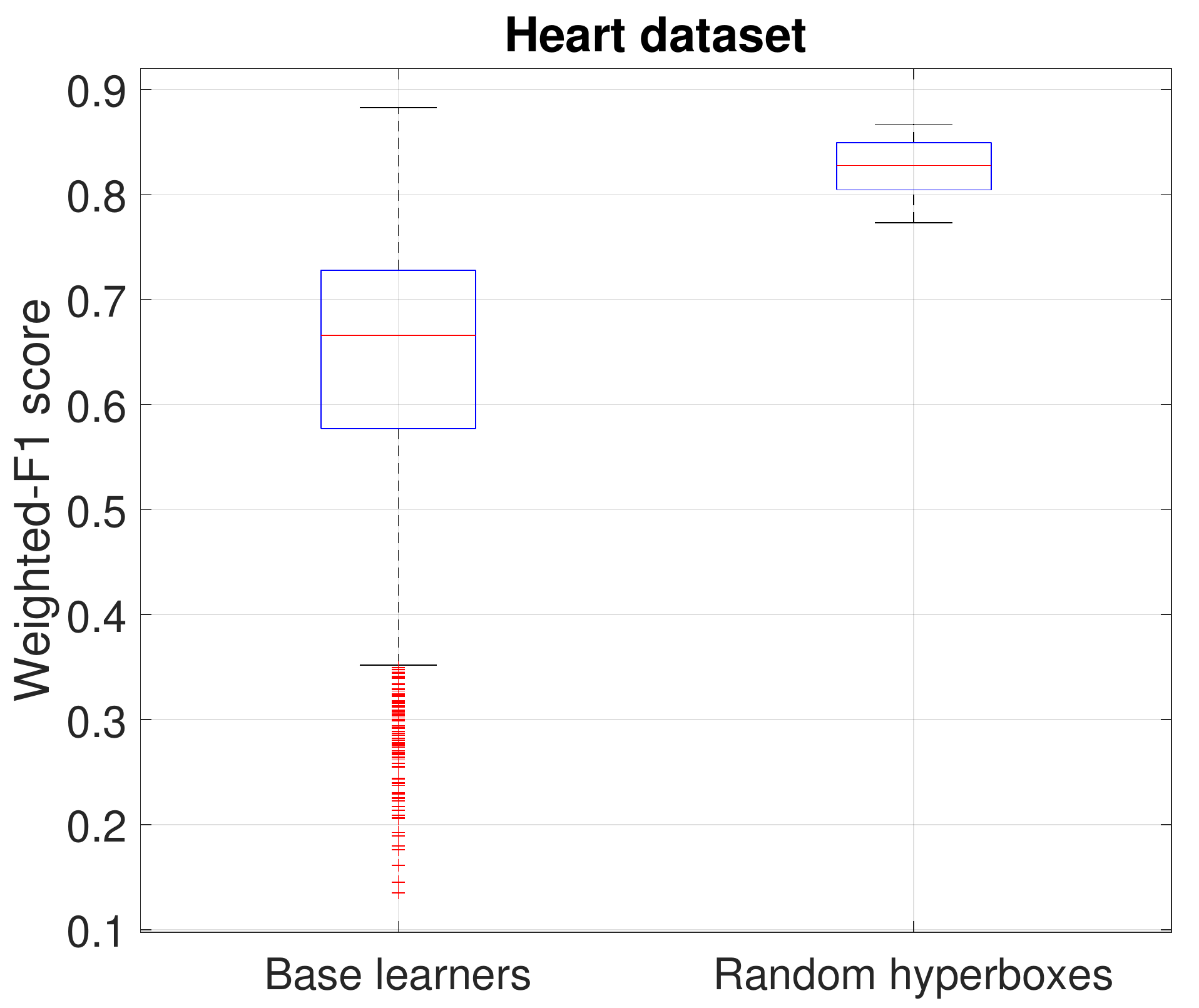}}
	\end{subfloat}
	\hfill%
	\begin{subfloat}[Vowel]{
			\includegraphics[width=0.32\textwidth,height=0.18\textheight]{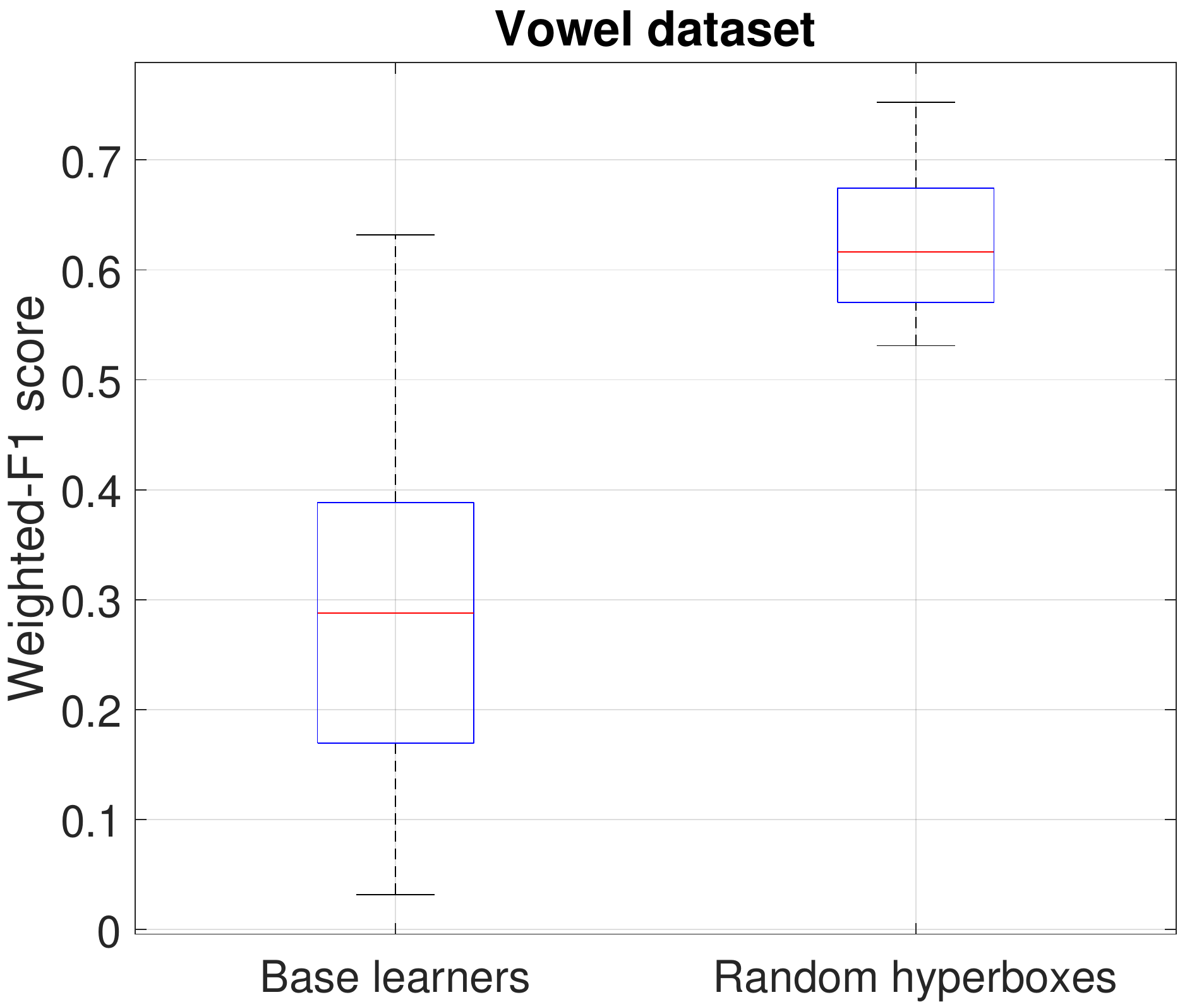}}
	\end{subfloat}
	\hfill%
	\begin{subfloat}[Ringnorm]{
			\includegraphics[width=0.32\textwidth,height=0.18\textheight]{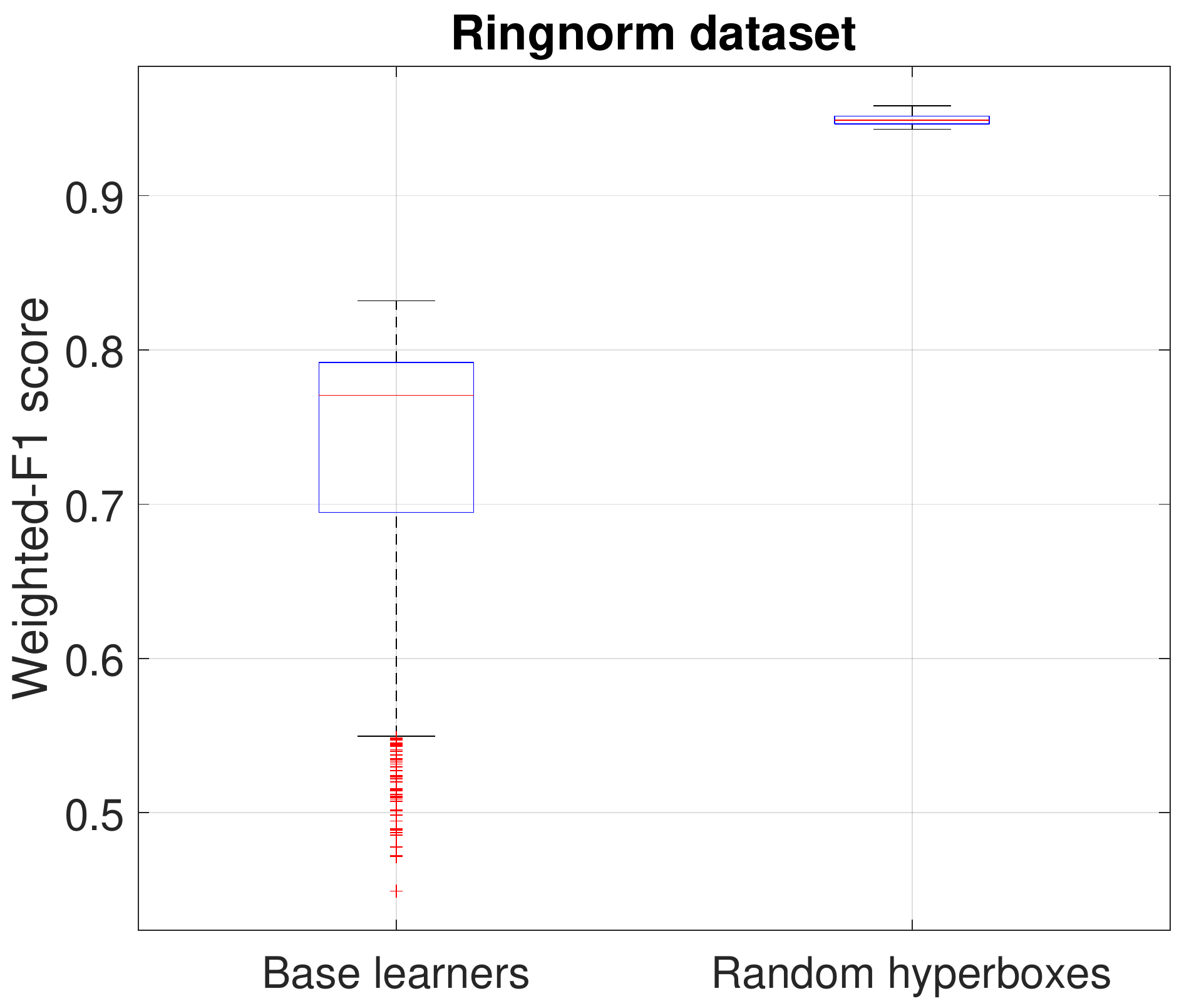}}
	\end{subfloat}
	\hfill%
	\begin{subfloat}[Connectionist\_bench\_sonar]{
			\includegraphics[width=0.32\textwidth,height=0.18\textheight]{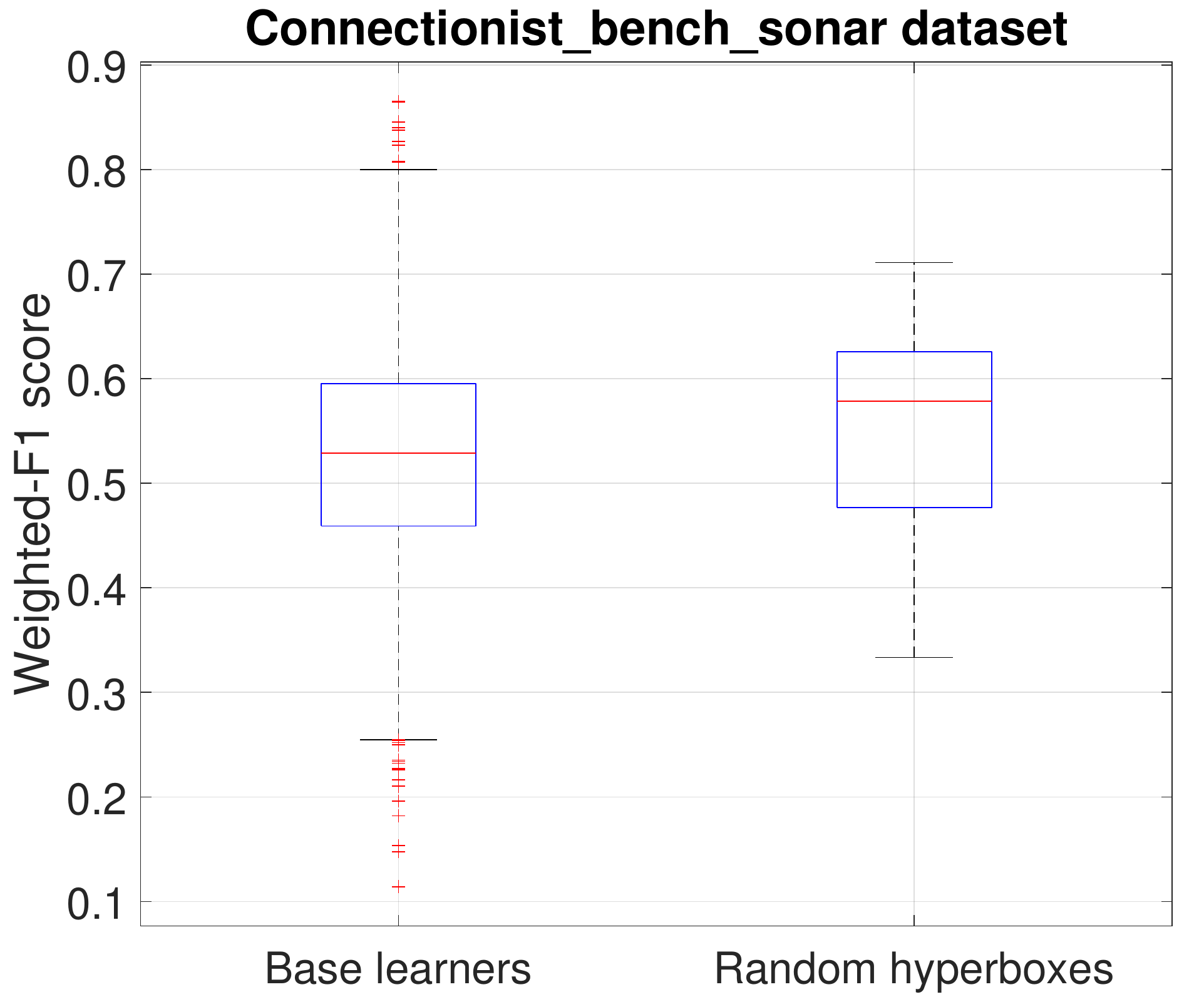}}
	\end{subfloat}
	\hfill
	\caption{The variances of the random hyperboxes models and their base learners for different datasets.}
	\label{Fig_S1}
\end{figure}

This part provides some supplementary figures for subsection IV.A.1 from the main paper. This experiment was performed on six datasets with diversity in the numbers of samples, features, and classes, i.e., \textit{plant\_species\_leaves\_margin}, \textit{plant\_species\_leaves\_shape}, \textit{heart}, \textit{vowel}, \textit{ringnorm}, and \textit{connectionist\_bench\_sonar}. Fig. \ref{Fig_S1} shows the variance values in terms of weighted-F1 scores using the 10 times repeated 4-fold cross-validation of base classifiers and the random hyperboxes models over different datasets. These results confirm that the random hyperboxes model is able to reduce the variance in its base learners, and so it can achieve better performance than its base models.

\begin{figure}[!ht]
	\begin{subfloat}[Plant\_species\_leaves\_margin]{
			\includegraphics[width=0.32\textwidth, height=0.18\textheight]{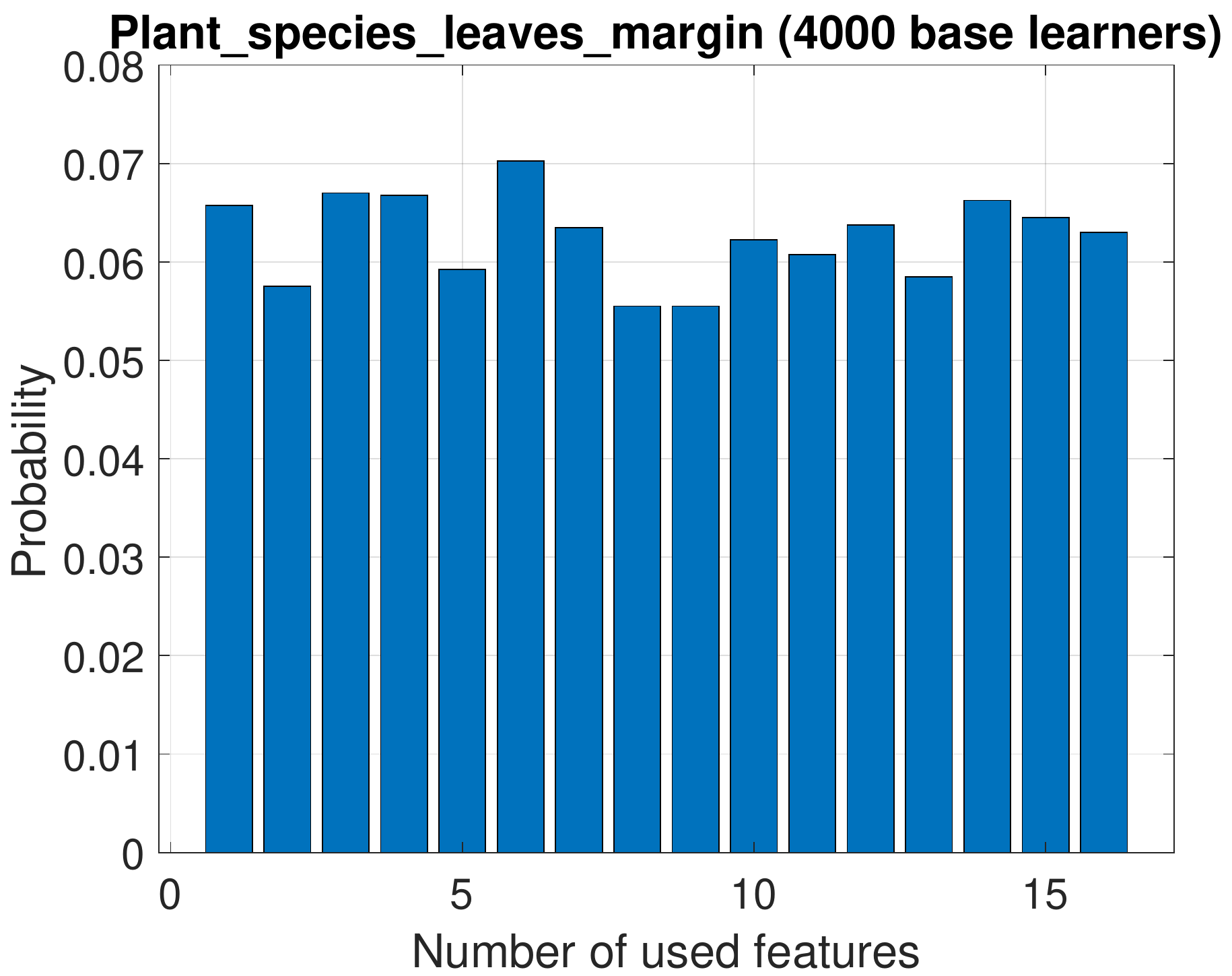}}
	\end{subfloat}
	\hfill%
	\begin{subfloat}[Plant\_species\_leaves\_shape]{
			\includegraphics[width=0.32\textwidth, height=0.18\textheight]{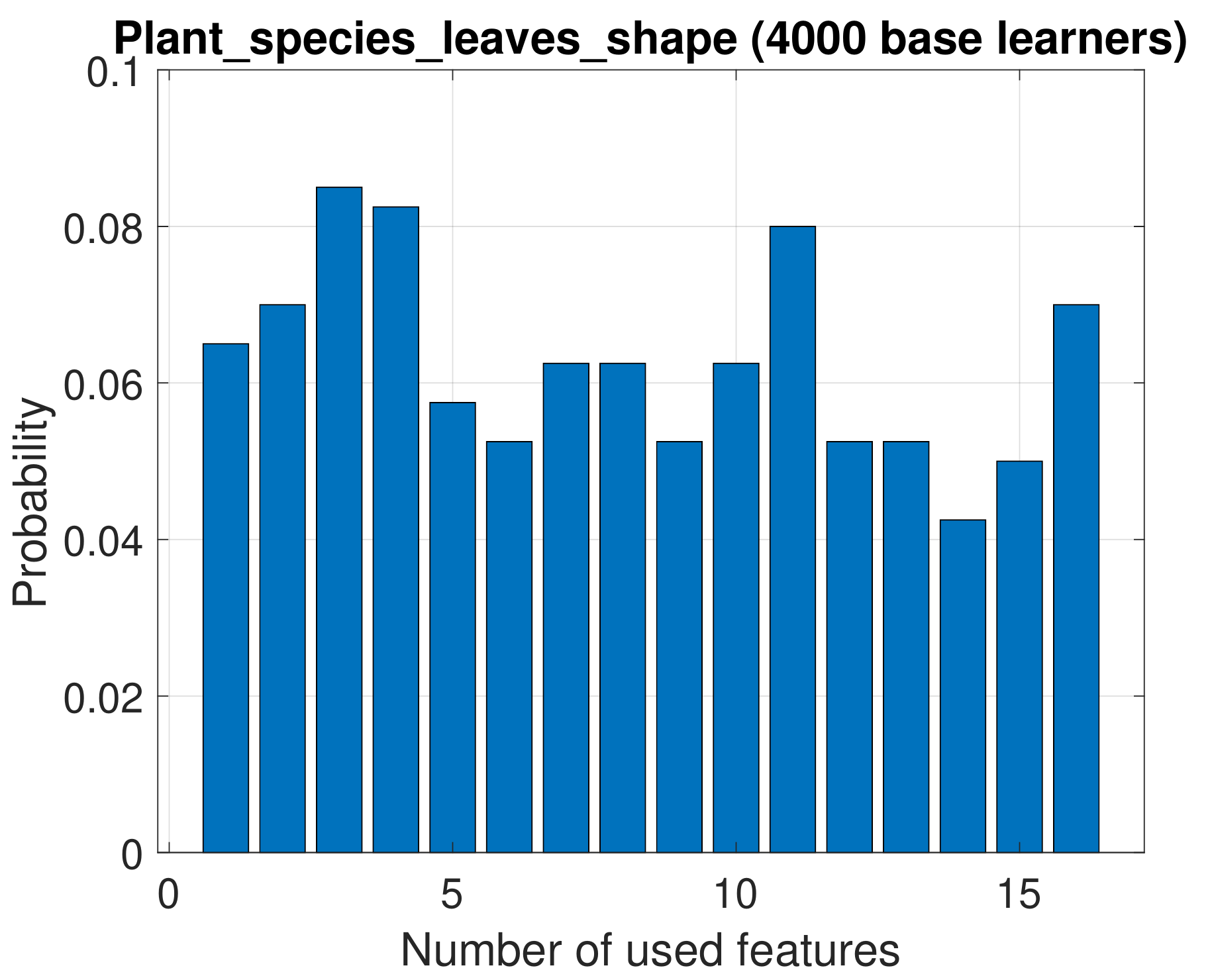}}
	\end{subfloat}
	\hfill%
	\begin{subfloat}[Heart]{
			\includegraphics[width=0.32\textwidth, height=0.18\textheight]{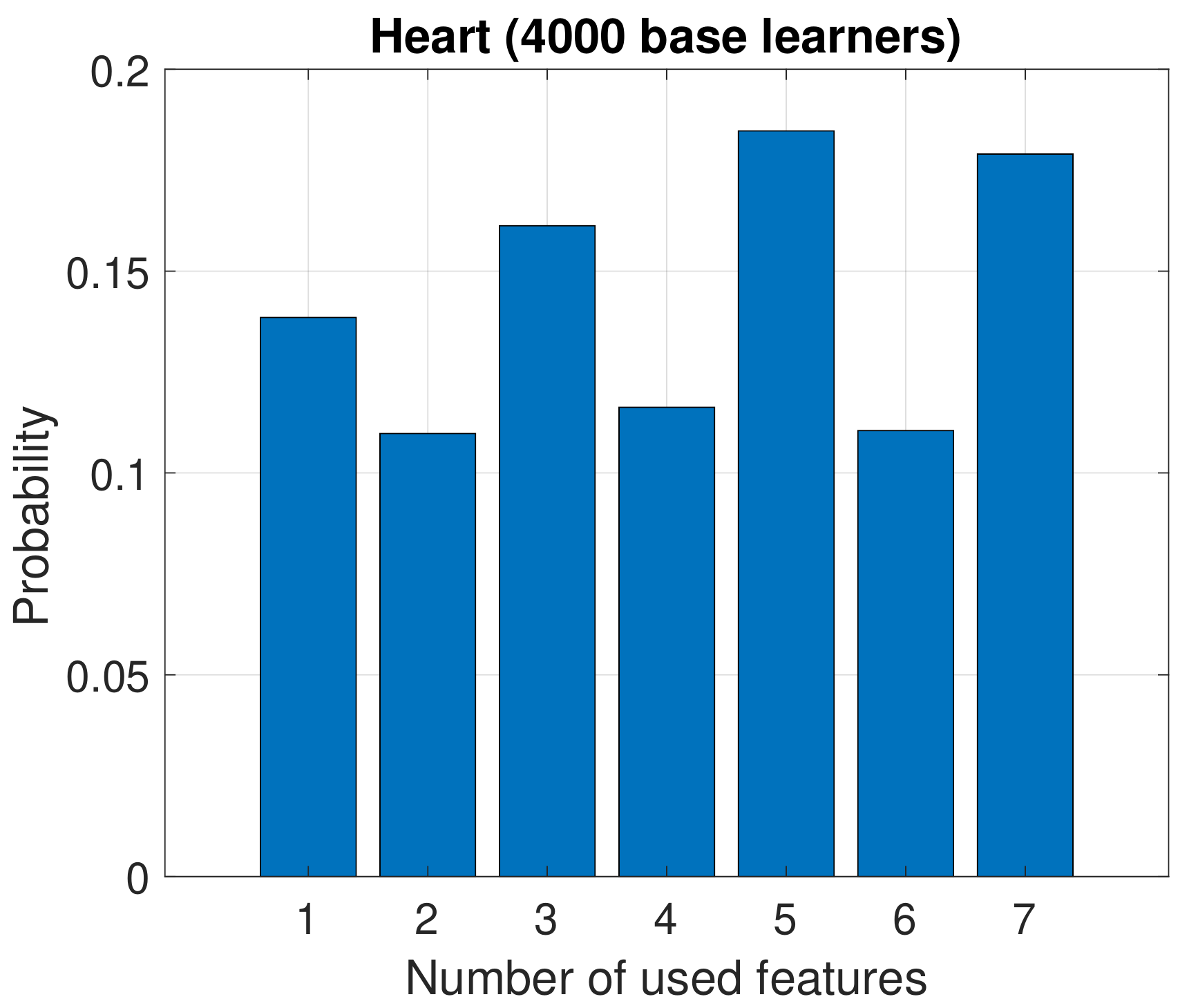}}
	\end{subfloat}
	\hfill%
	\begin{subfloat}[Vowel]{
			\includegraphics[width=0.32\textwidth, height=0.18\textheight]{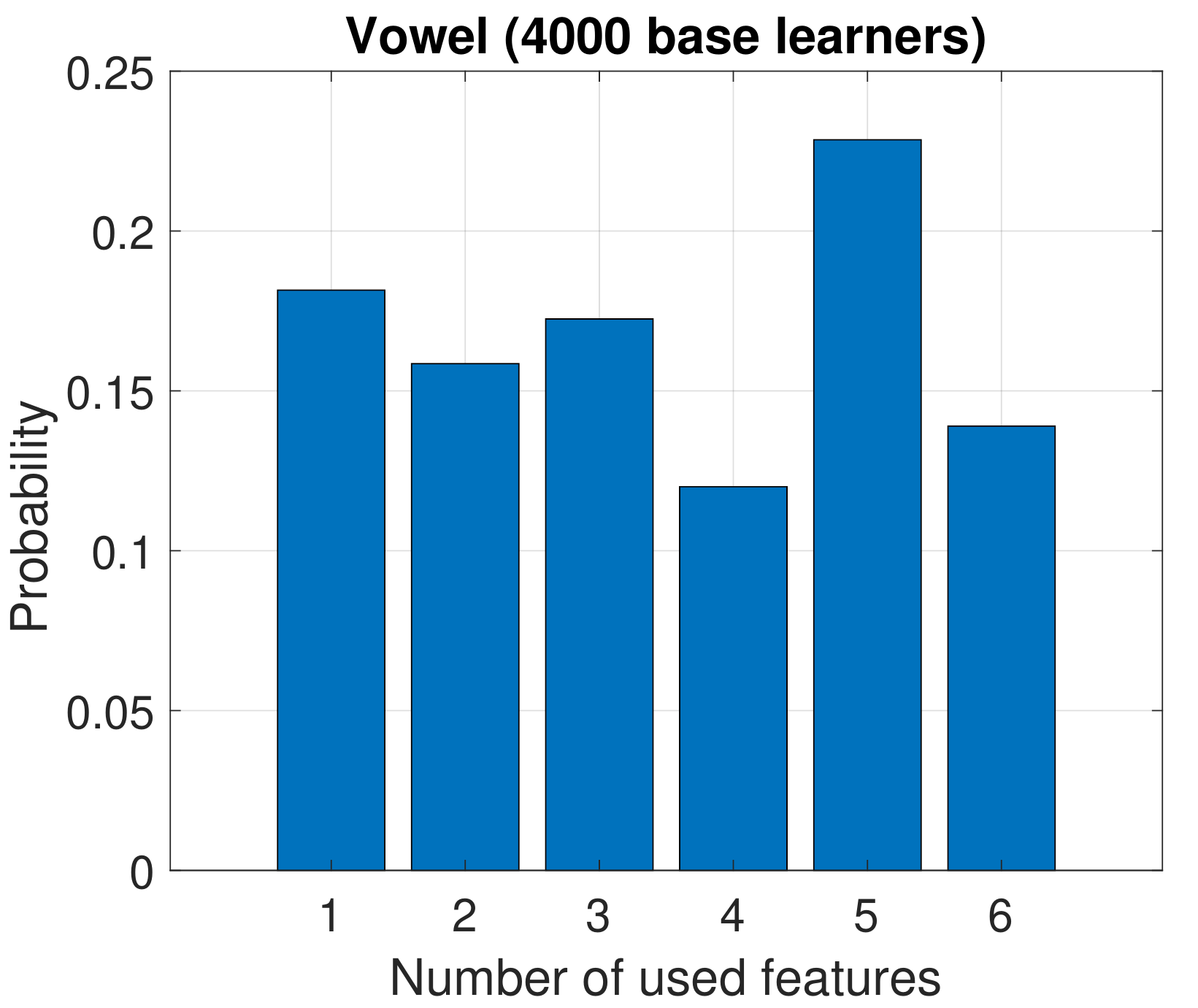}}
	\end{subfloat}
	\hfill%
	\begin{subfloat}[Ringnorm]{
			\includegraphics[width=0.32\textwidth, height=0.18\textheight]{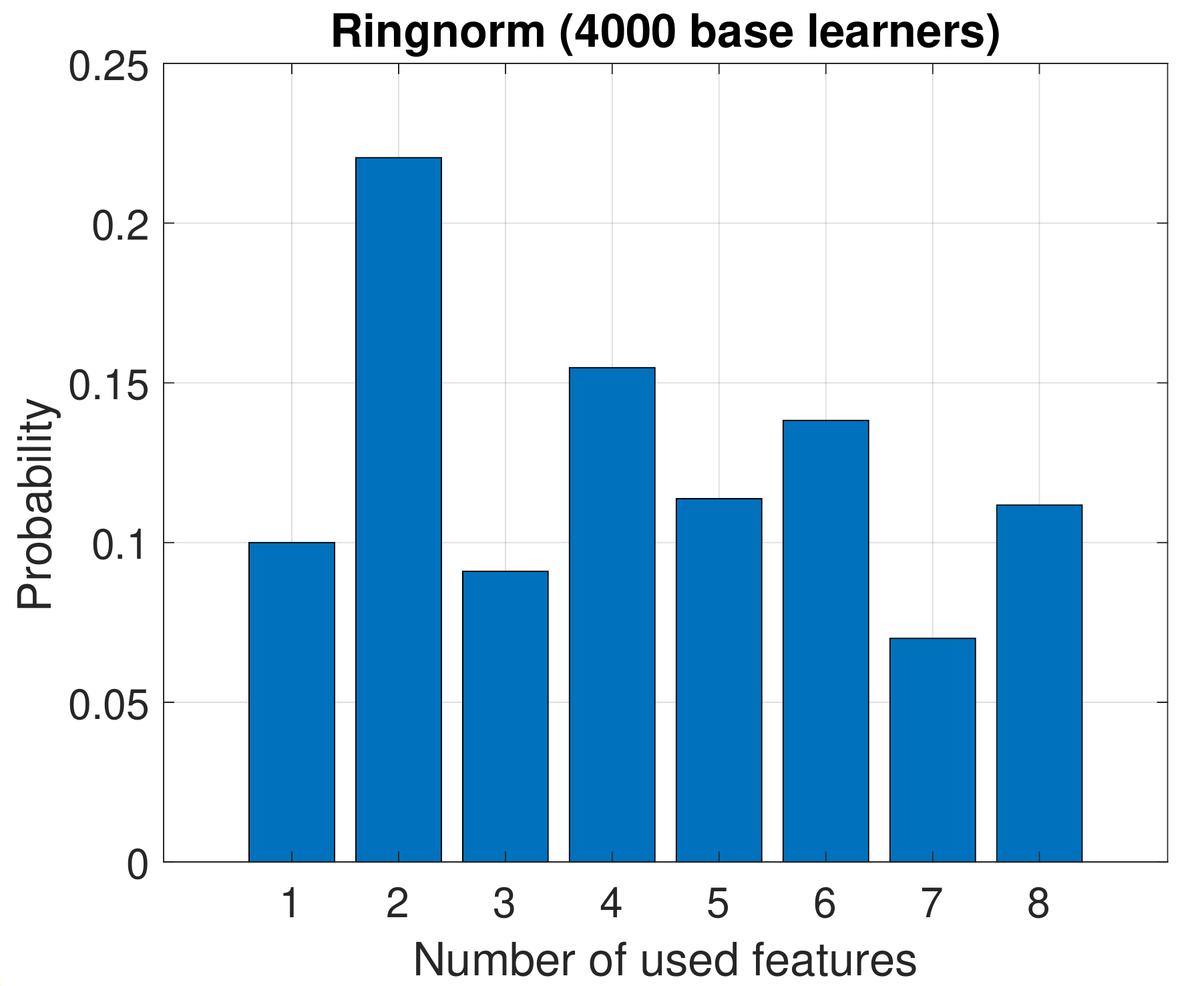}}
	\end{subfloat}
	\hfill%
	\begin{subfloat}[Connectionist\_bench\_sonar]{
			\includegraphics[width=0.32\textwidth, height=0.18\textheight]{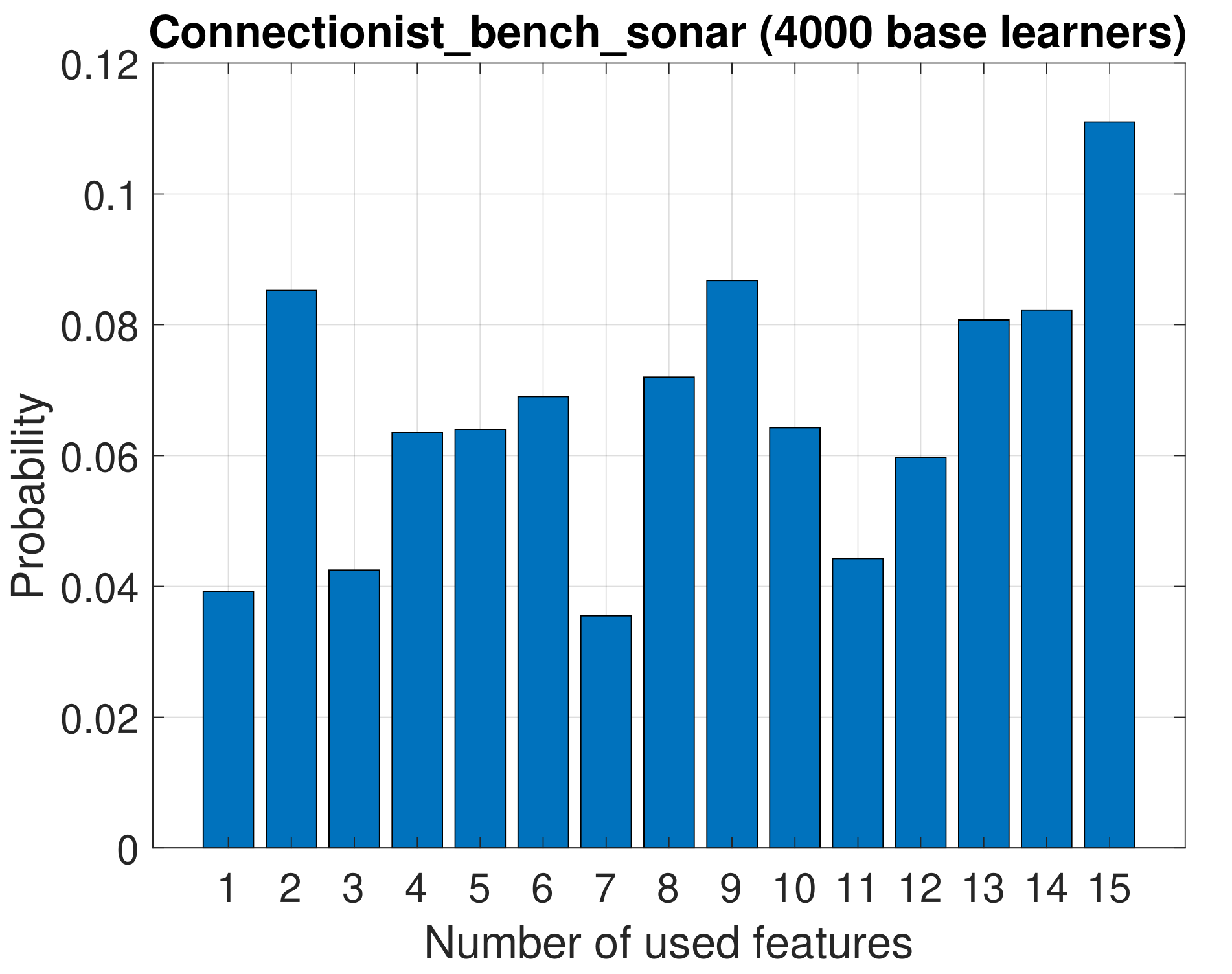}}
	\end{subfloat}
	\hfill
	\caption{The probability of the number of used features for all base learners over different datasets.}
	\label{Fig_S2}
\end{figure}

Fig. \ref{Fig_S2} shows the probability of the number of features, $d$, used to build the 4000 base learners for the experiment shown in subsection IV.A.1 from the main paper. It can be observed that the probability distribution of the number of used features is nearly uniform in all 4000 base learners.

\begin{figure}[!ht]
	\begin{subfloat}[Plant\_species\_leaves\_margin]{
			\includegraphics[width=0.32\textwidth, height=0.18\textheight]{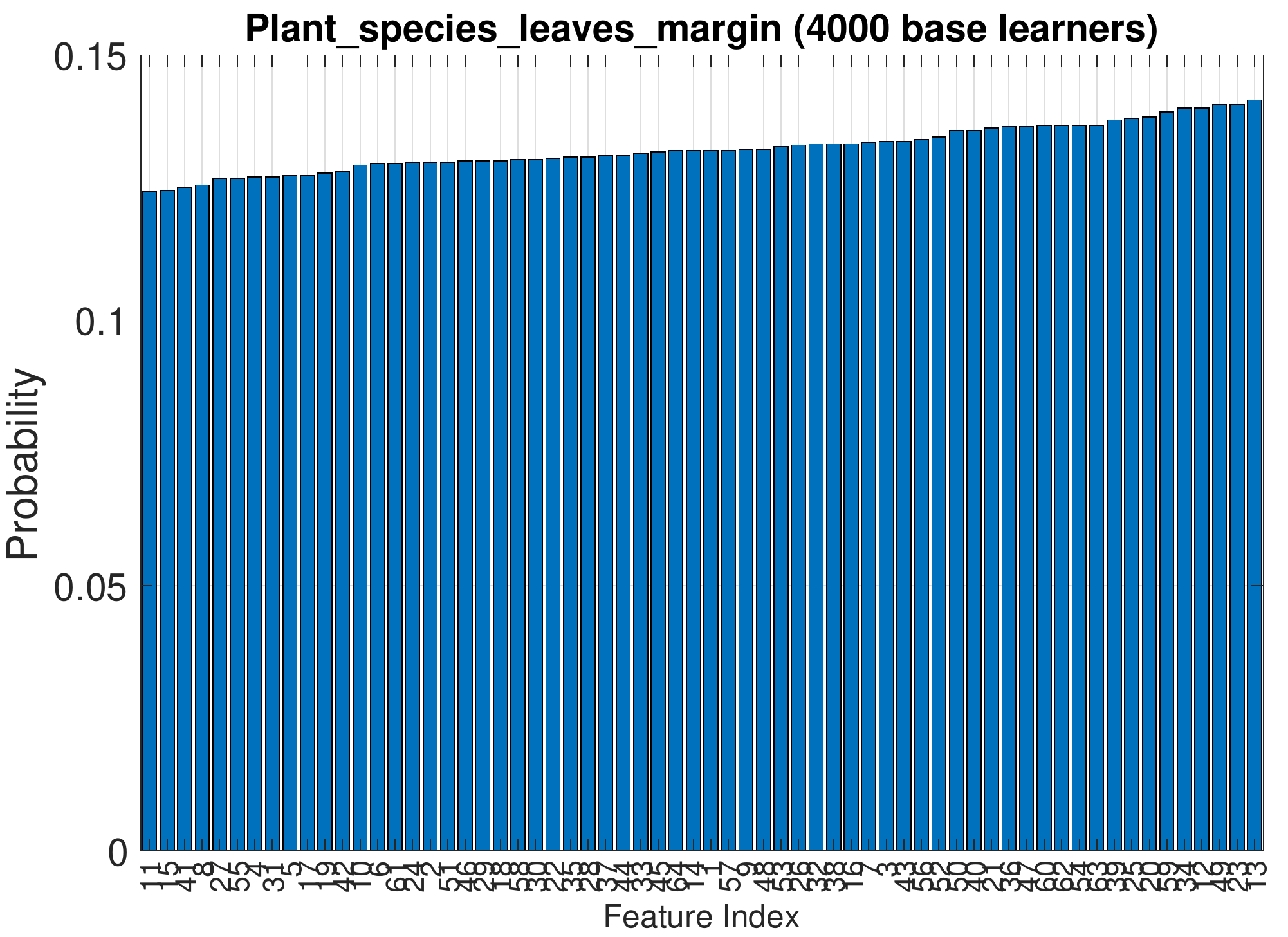}}
	\end{subfloat}
	\hfill%
	\begin{subfloat}[Plant\_species\_leaves\_shape]{
			\includegraphics[width=0.32\textwidth, height=0.18\textheight]{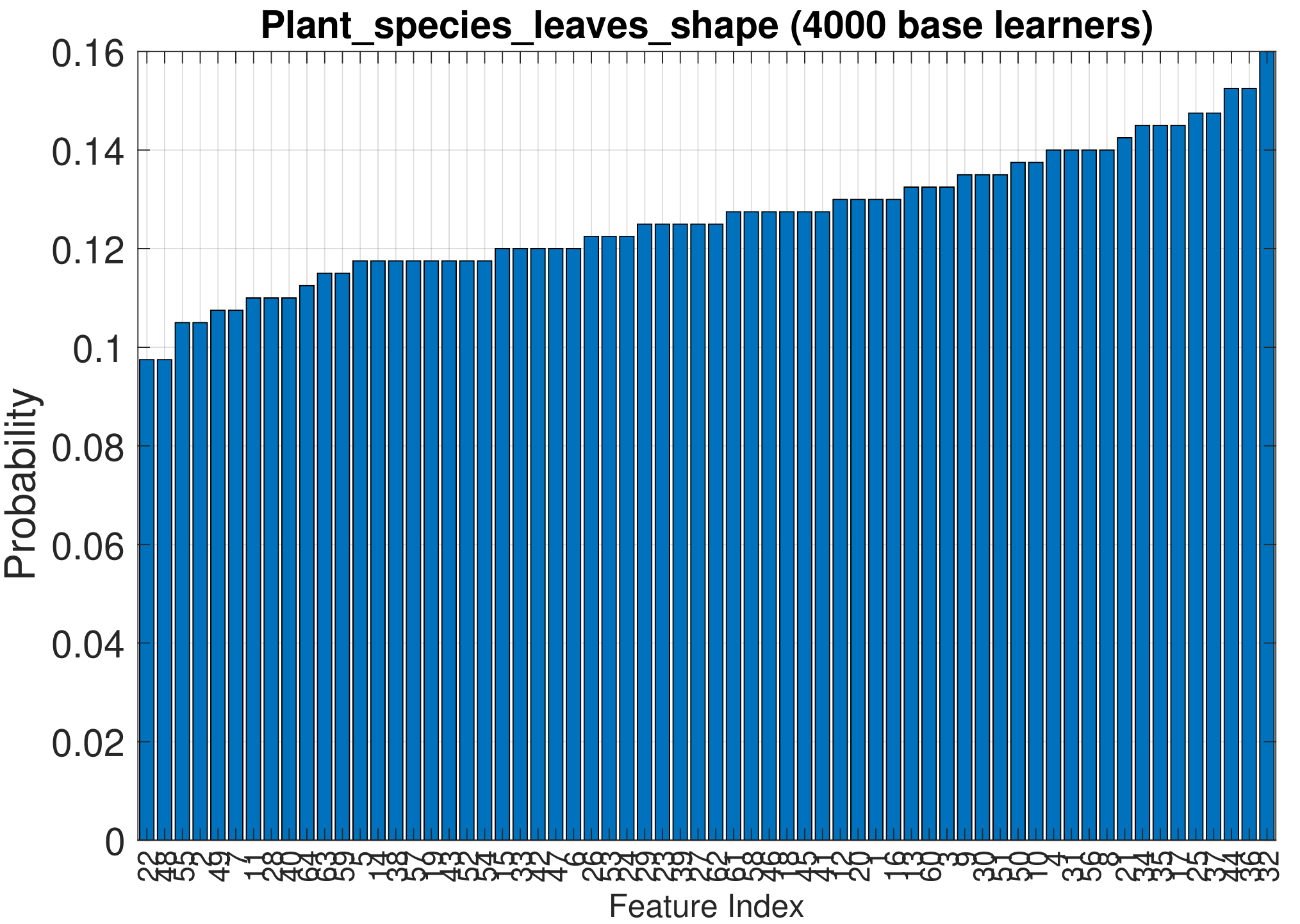}}
	\end{subfloat}
	\hfill%
	\begin{subfloat}[Heart]{
			\includegraphics[width=0.32\textwidth, height=0.19\textheight]{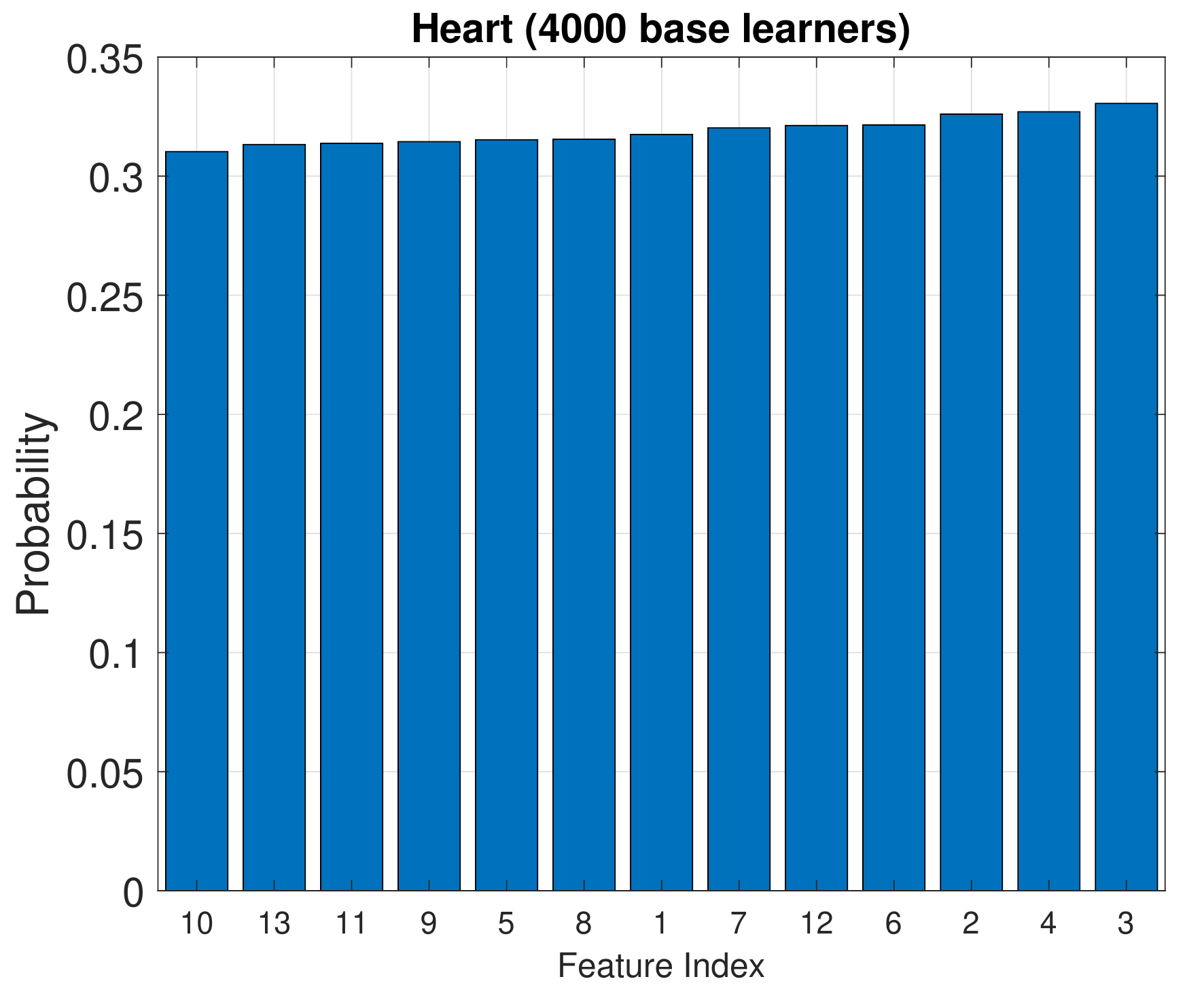}}
	\end{subfloat}
	\hfill%
	\begin{subfloat}[Vowel]{
			\includegraphics[width=0.32\textwidth, height=0.18\textheight]{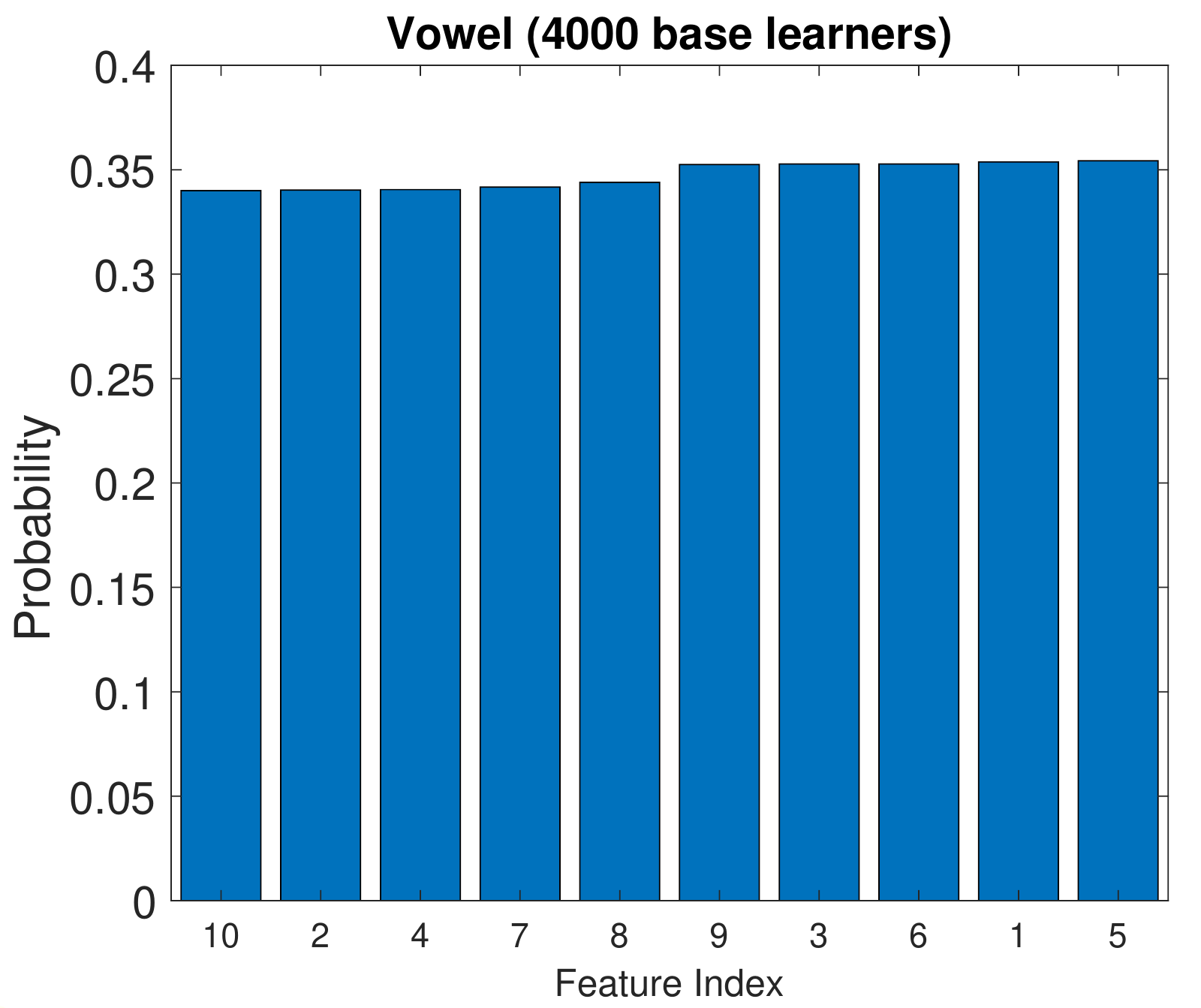}}
	\end{subfloat}
	\hfill%
	\begin{subfloat}[Ringnorm]{
			\includegraphics[width=0.32\textwidth, height=0.18\textheight]{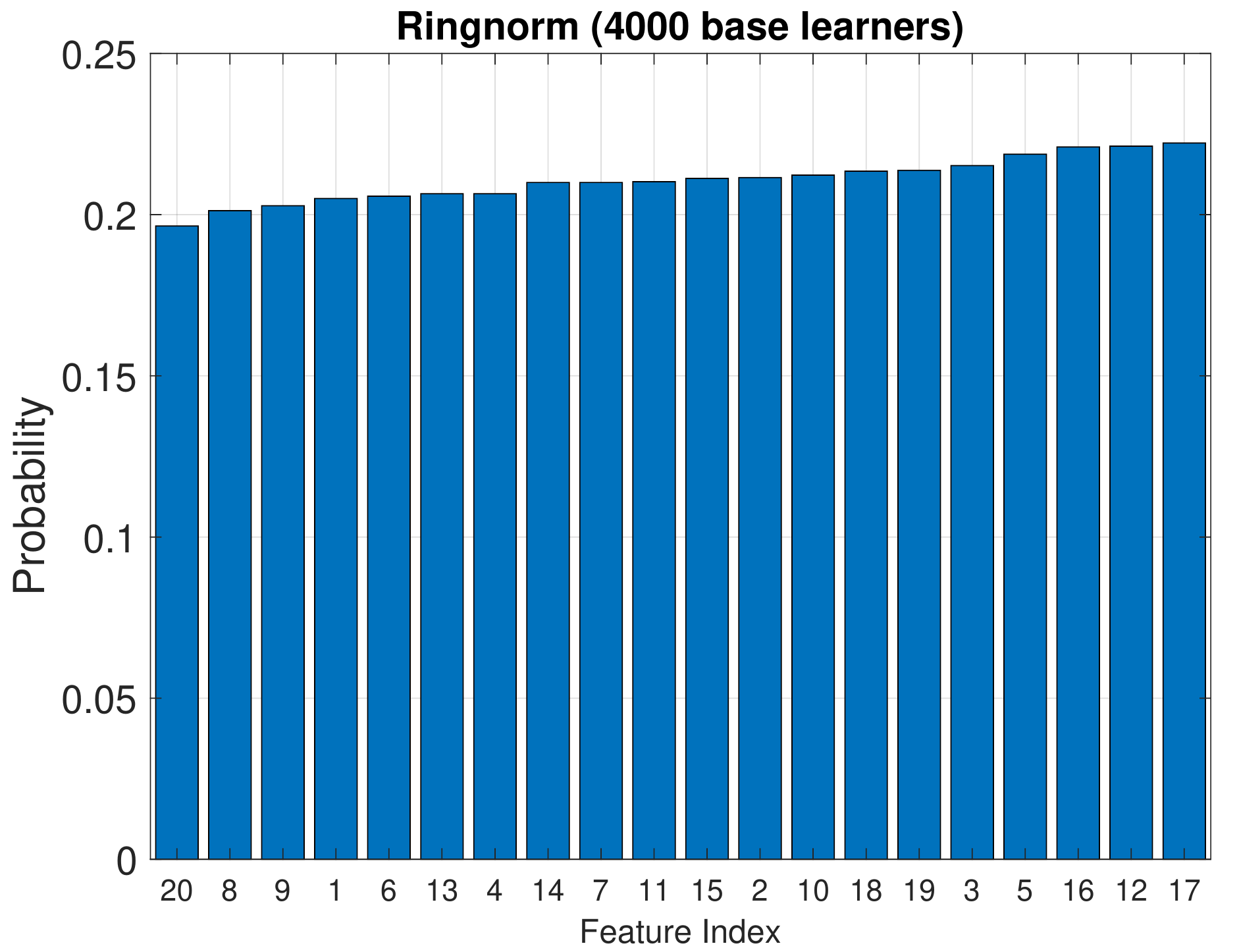}}
	\end{subfloat}
	\hfill%
	\begin{subfloat}[Connectionist\_bench\_sonar]{
			\includegraphics[width=0.32\textwidth, height=0.18\textheight]{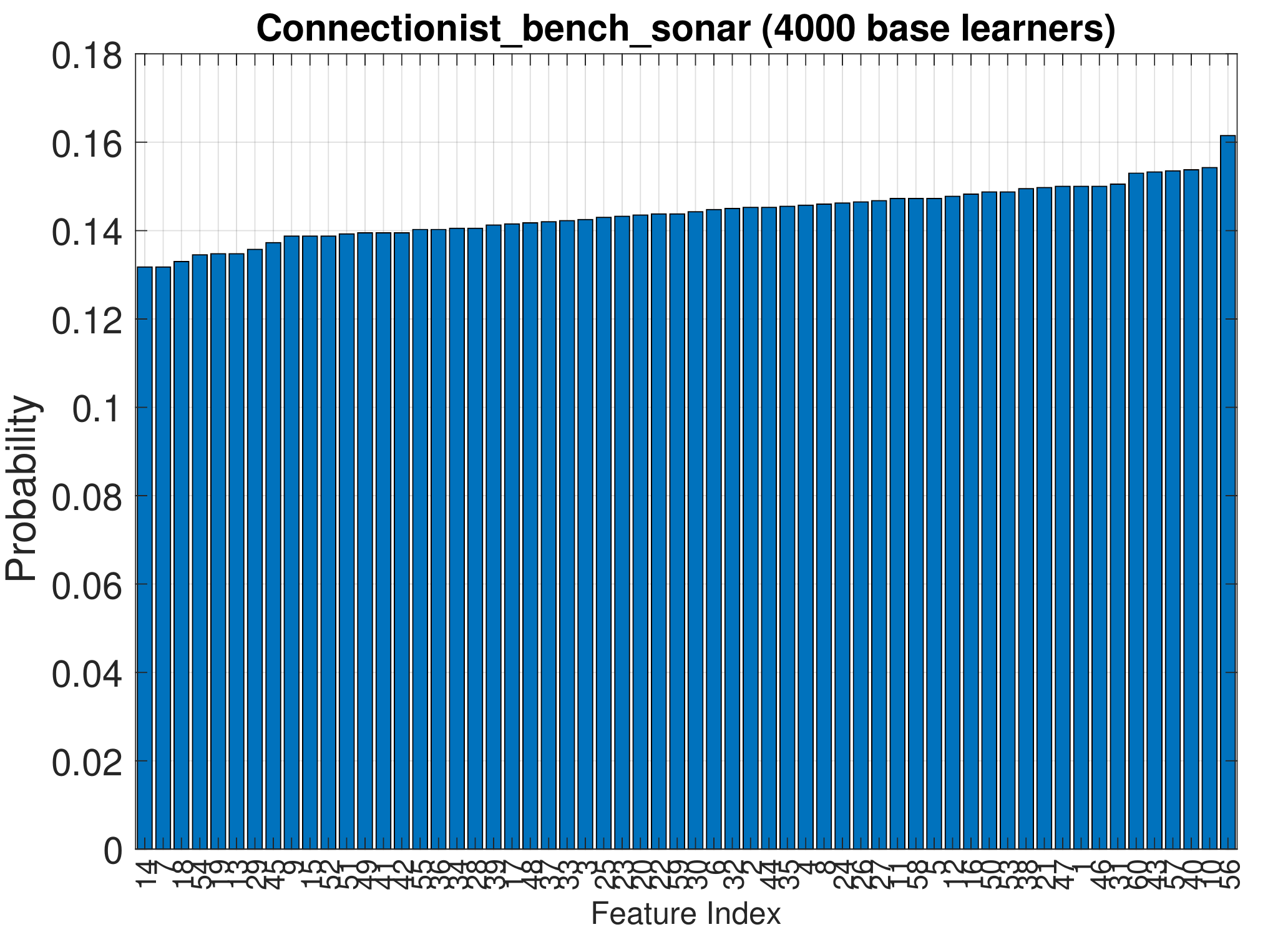}}
	\end{subfloat}
	\hfill
	\caption{The probability of each feature used for all base learners over different datasets.}
	\label{Fig_S3}
\end{figure}

\begin{figure}[!ht]
	\begin{subfloat}[Plant\_species\_leaves\_margin]{
			\includegraphics[width=0.32\textwidth, height=0.19\textheight]{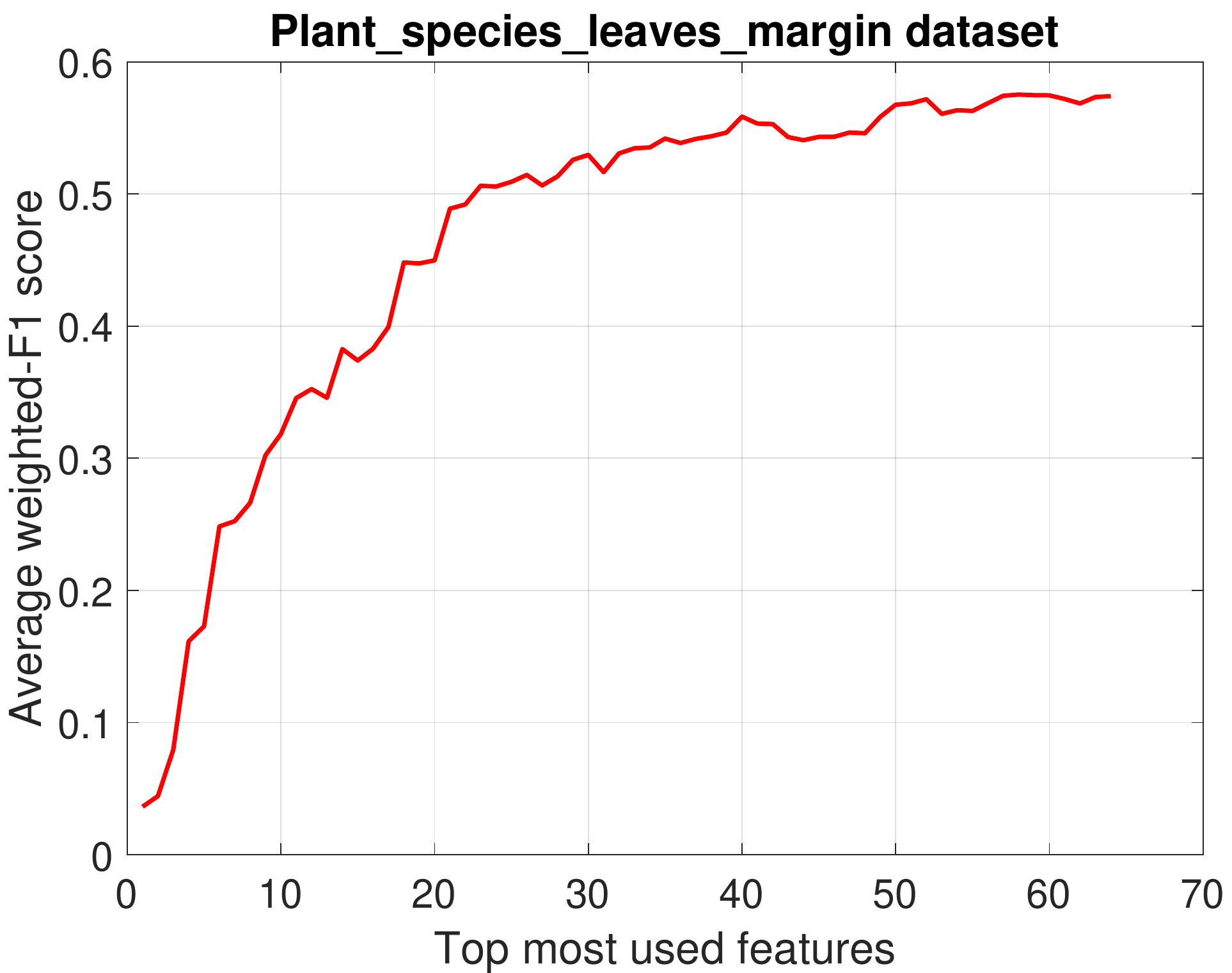}}
	\end{subfloat}
	\hfill%
	\begin{subfloat}[Plant\_species\_leaves\_shape]{
			\includegraphics[width=0.32\textwidth, height=0.19\textheight]{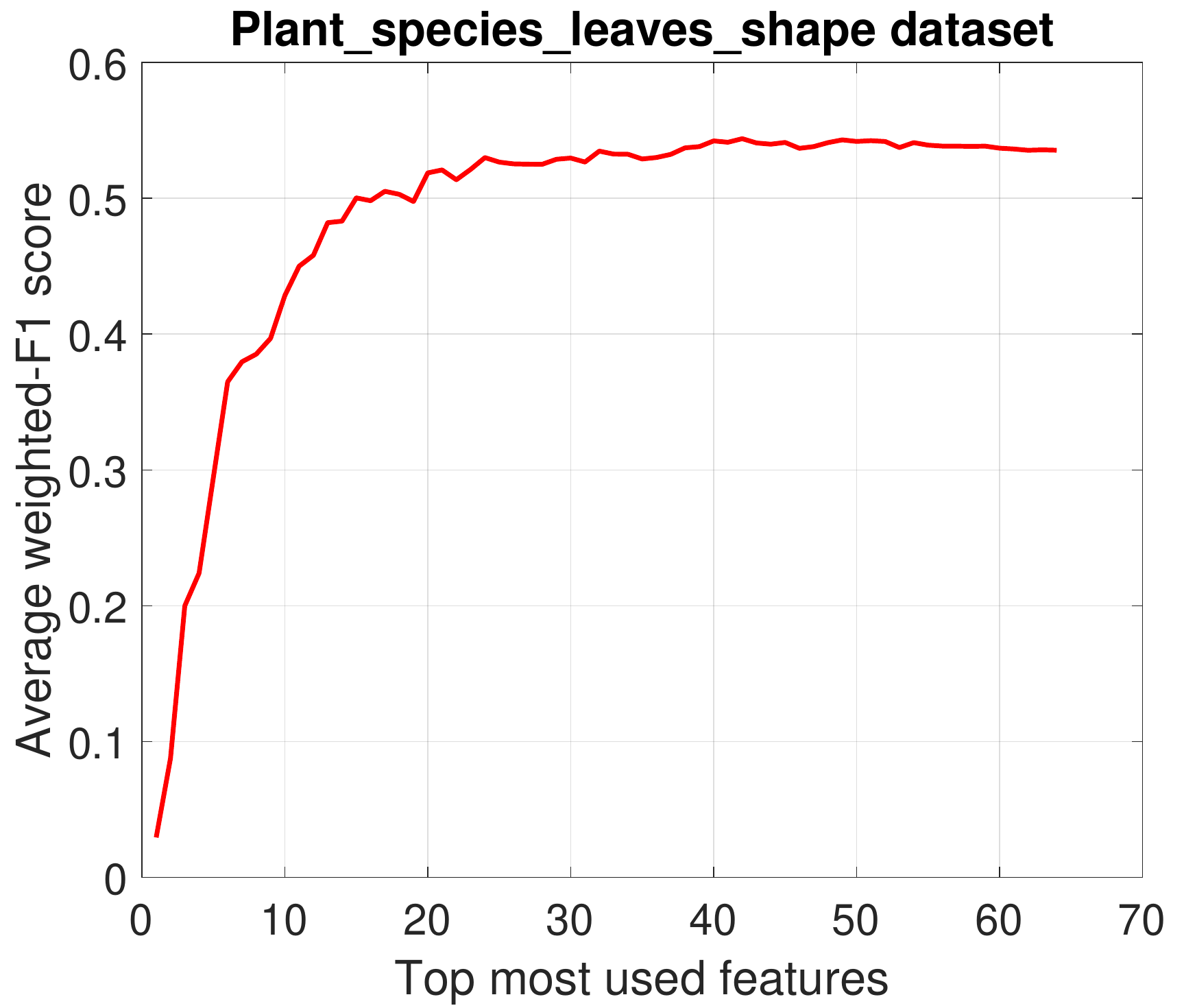}}
	\end{subfloat}
	\hfill%
	\begin{subfloat}[Heart]{
			\includegraphics[width=0.32\textwidth, height=0.19\textheight]{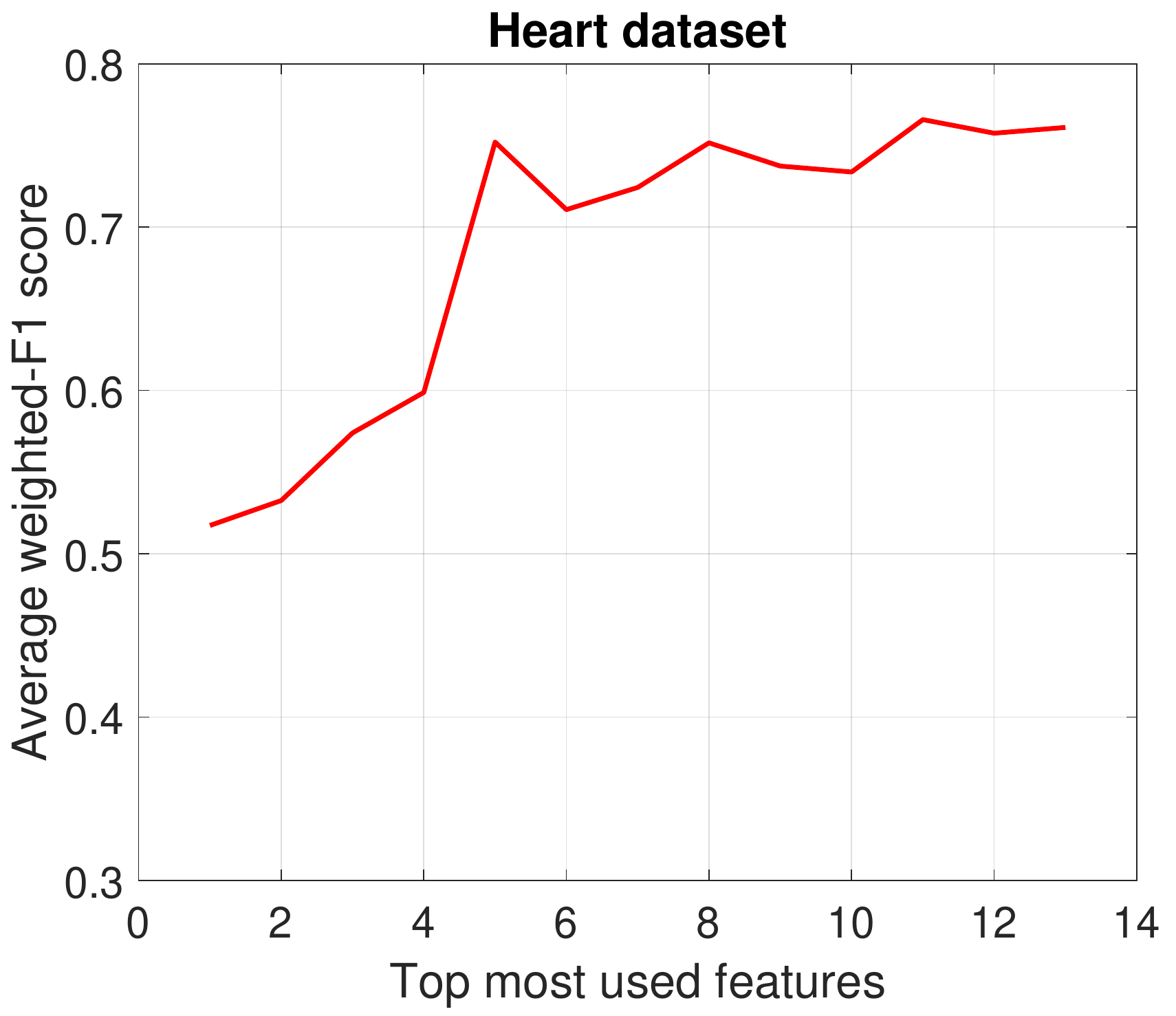}}
	\end{subfloat}
	\hfill%
	\begin{subfloat}[Vowel]{
			\includegraphics[width=0.32\textwidth, height=0.19\textheight]{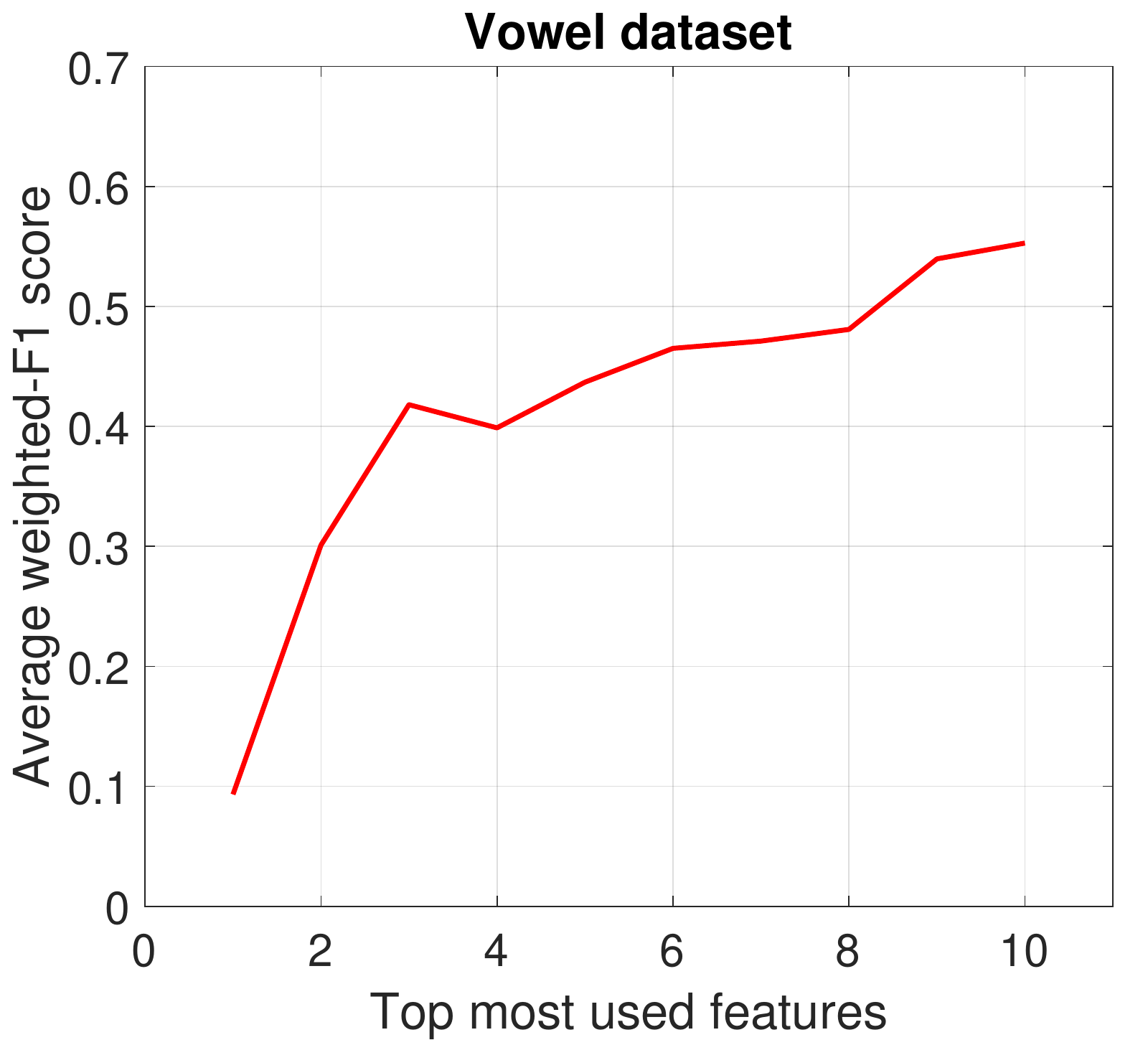}}
	\end{subfloat}
	\hfill%
	\begin{subfloat}[Ringnorm]{
			\includegraphics[width=0.32\textwidth, height=0.19\textheight]{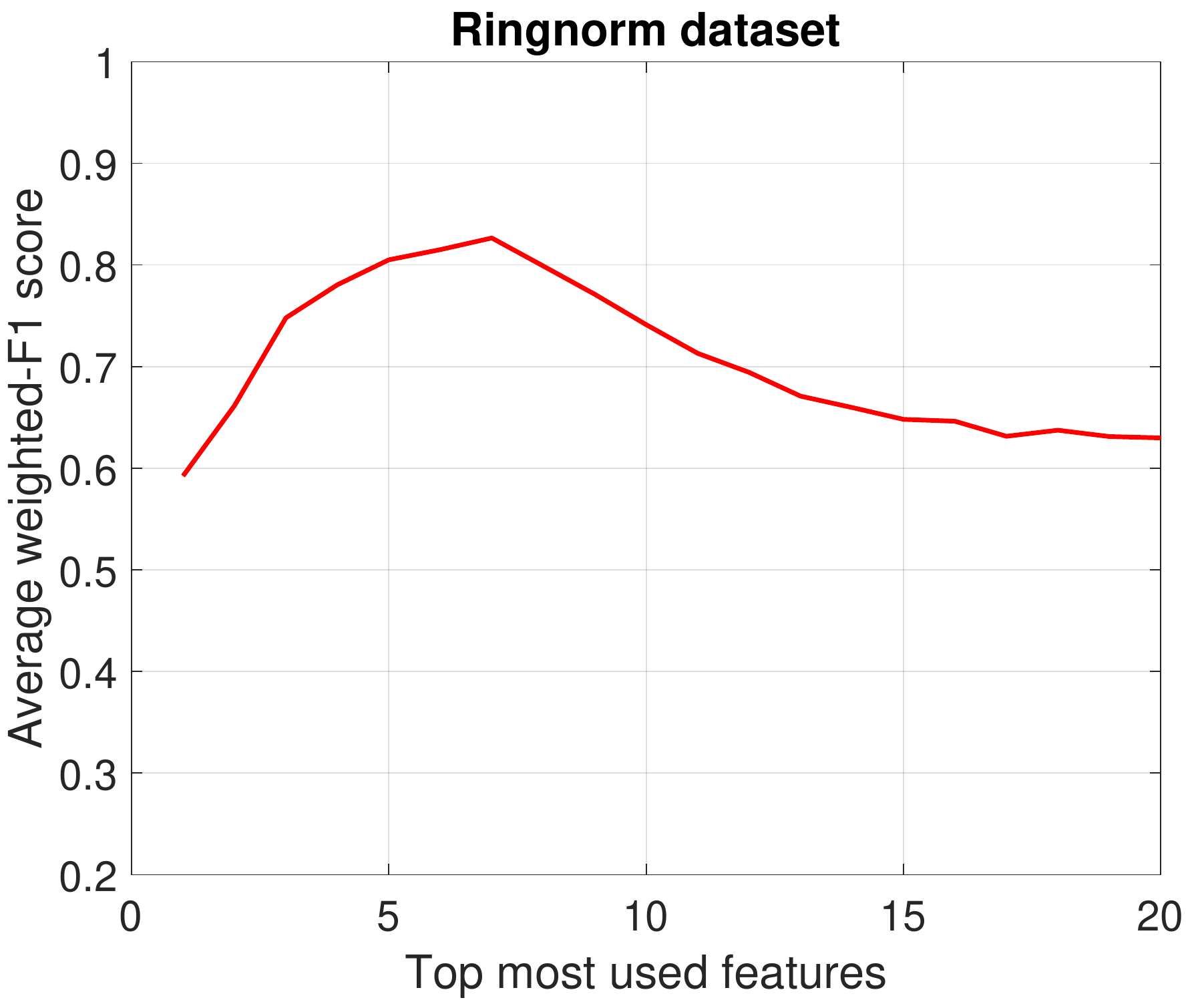}}
	\end{subfloat}
	\hfill%
	\begin{subfloat}[Connectionist\_bench\_sonar]{
			\includegraphics[width=0.32\textwidth, height=0.19\textheight]{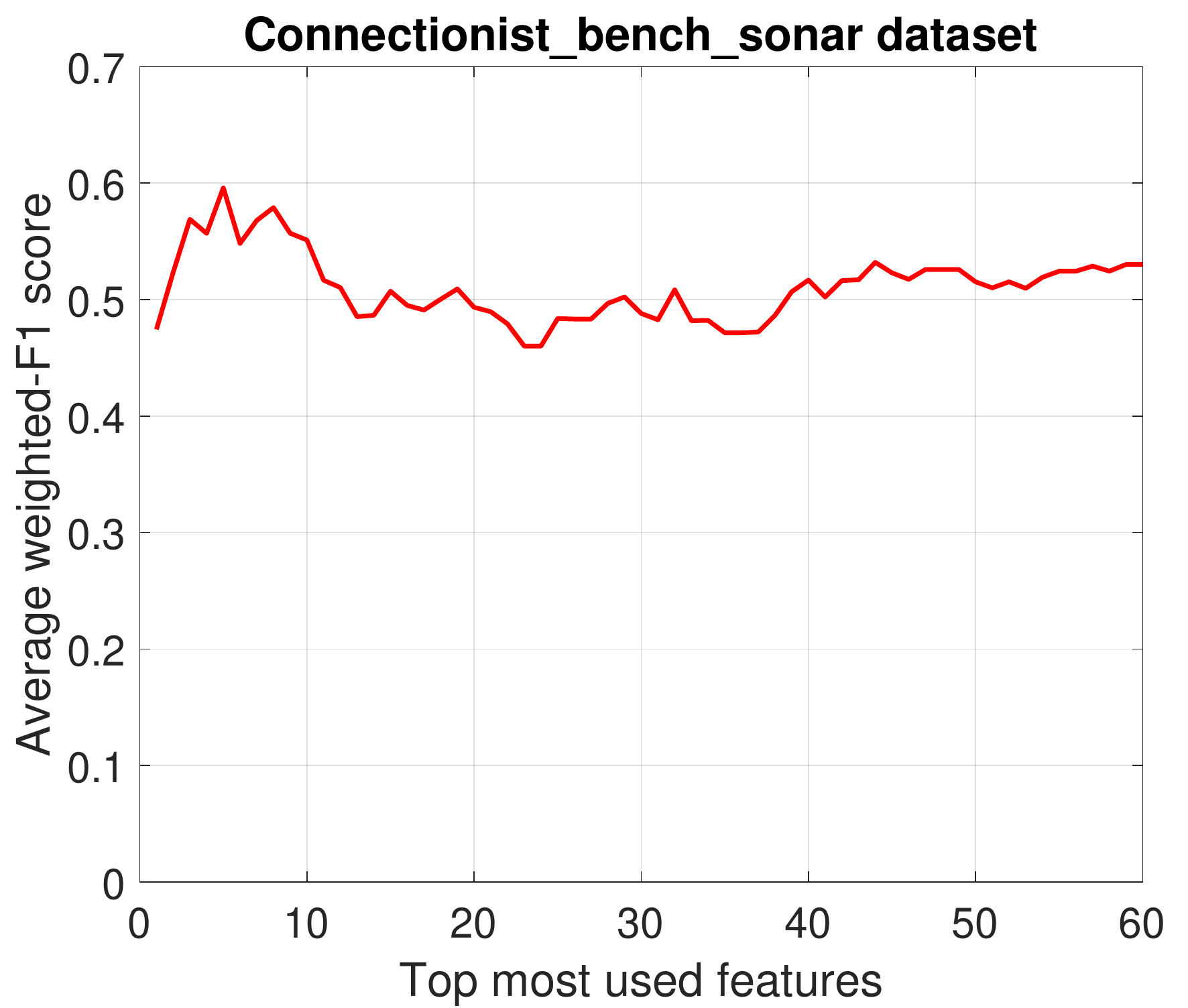}}
	\end{subfloat}
	\hfill
	\caption{Average weighted-F1 scores over 40 testing folds of a single model using training sets with top-k most used features over different datasets.}
	\label{Fig_S4}
\end{figure}

We can also identify the used probability of each feature over 4000 base learners to find the importance scores of features with respect to the performance of the ensemble model. This information is given in Fig. \ref{Fig_S3}. From the importance scores of features, we built a single model using top-K of the most important features to assess the performance of the random hyperboxes and the use of single models. We can observe that in many datasets, the single model often achieves better performance when it is trained on more features. However, in several cases such as in \textit{ringnorm} and \textit{connectionist\_bench\_sonar} datasets, the best performance of the single model is obtained if it is trained on a subset of the most important features. From Figs. \ref{Fig_S1} and \ref{Fig_S4}, it is easily seen that the random hyperboxes model trained using a subset of features usually achieves higher classification accuracy than the single model trained on the same dataset using all of the available features.

\subsection{Analyzing the Effectiveness of the Random Hyperboxes on High Dimensional Data}\label{sup_high_dim}
When building predictive models for problems with very high dimensional data, the performance of models is negatively influenced by the redundancy of features. This problem is known as the Curse of Dimensionality \cite{Friedman1997}. This experiment is to assess the robustness of the random hyperboxes classifier for high dimensional data in comparison to the single IOL-GFMM model. We used two very high dimensional dataset, i.e., \textit{PEMS database} \cite{Cuturi11} and \textit{Complex Hydraulic System} \cite{Helwig15}. 80\% of samples in each dataset were used as training data and the remaining 20\% of samples were testing data. The summaries of these datasets are shown in Table \ref{high_dim_data}.

\begin{table}[!ht]
	\centering
	\caption{Summarize Information of High Dimensional Datasets}\label{high_dim_data}
	\scriptsize{
		\begin{tabular}{|L{1.1cm}|c|c|c|c|c|}
			\hline
			\textbf{Dataset}         & \textbf{\#samples} & \textbf{\#features} & \textbf{\#classes} & \textbf{\#training} & \textbf{\#testing} \\ \hline \hline
			PEMS database            & 440                & 138 672             & 7                  & 352                 & 88                 \\ \hline
			Complex Hydraulic System & 2205               & 43 680              & 2                  & 1764                & 441                \\ \hline
		\end{tabular}
	}
\end{table}

\begin{figure}[!ht]
	\centering
	\includegraphics[width=0.4\textwidth]{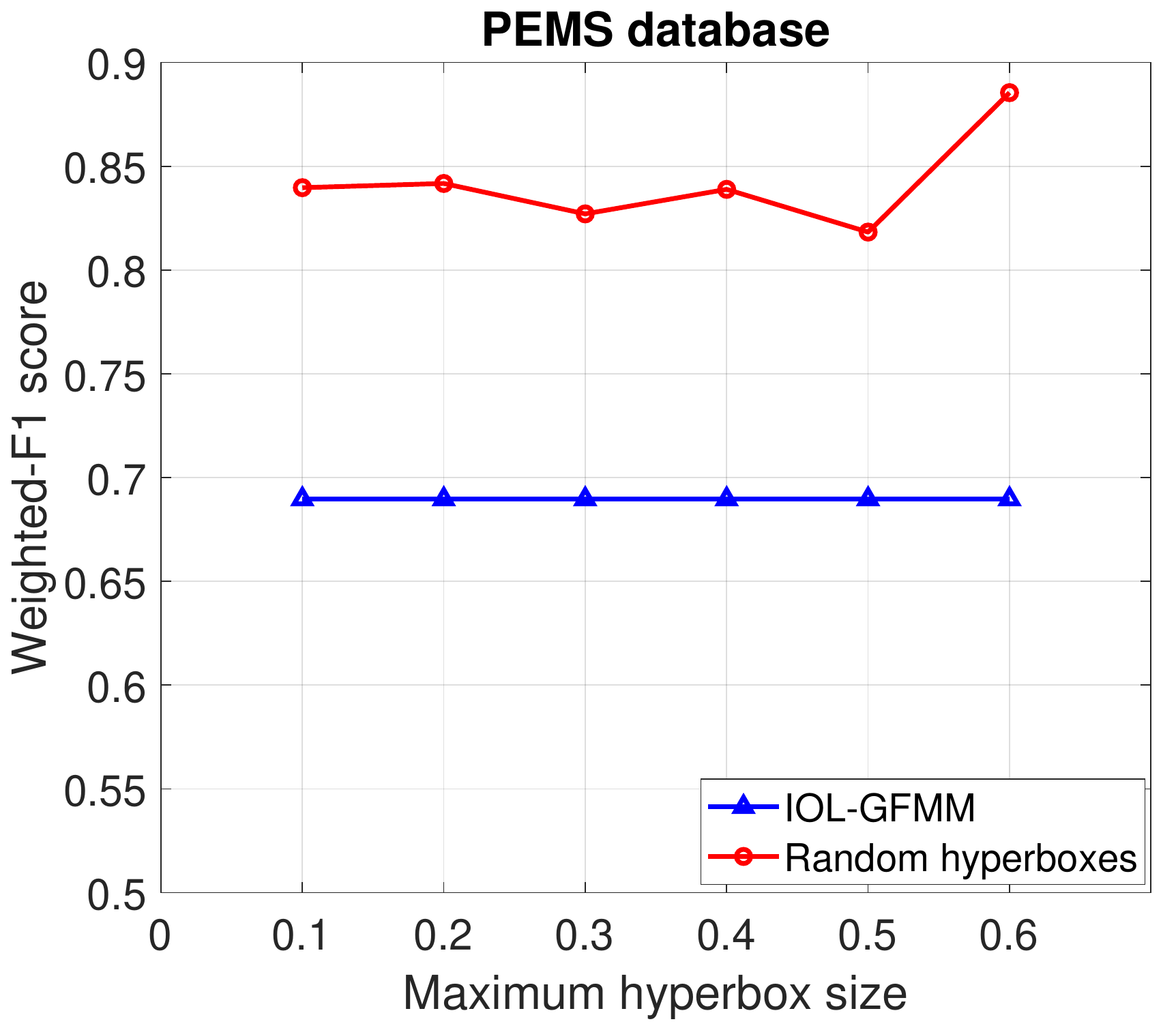}
	\caption{Weighted-F1 score of the random hyperboxes and IOL-GFMM for the \textit{PEMS database} dataset}
	\label{Fig_S5}
\end{figure}

\begin{figure}[!ht]
	\centering
	\includegraphics[width=0.4\textwidth]{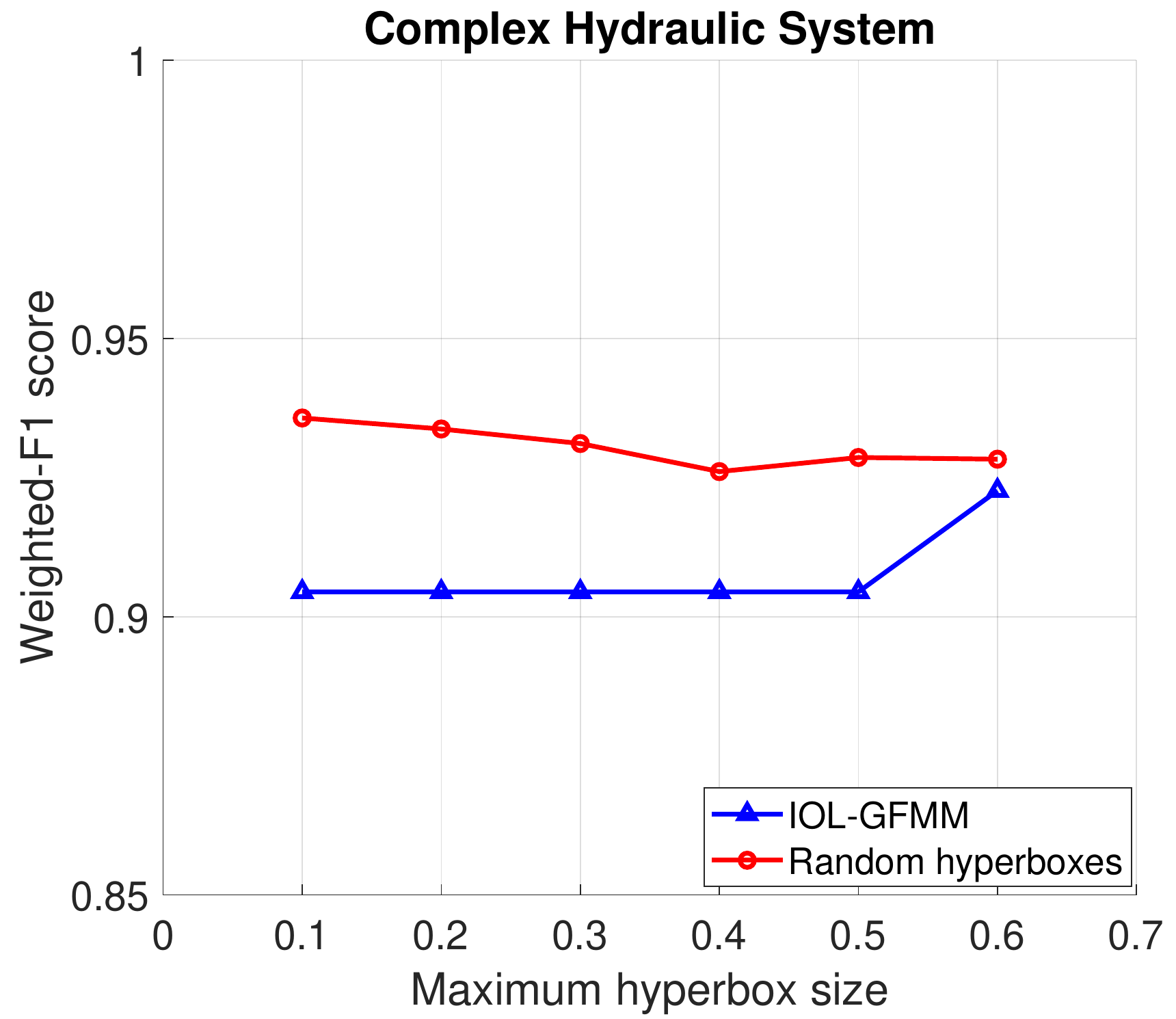}
	\caption{Weighted-F1 score of the random hyperboxes and IOL-GFMM for the \textit{Complex Hydraulic System} dataset}
	\label{Fig_S6}
\end{figure}

In this experiment, each base learner in the random hyperboxes model is trained on 50\% of samples randomly selected from the training data. The maximum number of used features for each base learner is set to $2 \sqrt{p}$, where $p$ is the number of dimensions of the dataset. The number of base learners for each random hyperboxes model is $ m = 100$. The weighted-F1 scores of the random hyperboxes and single IOL-GFMM model through different values of $\theta$ are given in Fig. \ref{Fig_S5} for the \textit{PEMS database} dataset and in Fig. \ref{Fig_S6} for the \textit{Complex Hydraulic System} dataset.

It can be observed that the IOL-GFMM has consistently lower performance than RH with the very high dimensional data. In contrast, the random hyperboxes can achieve high accuracy using only $2 \sqrt{p}$ random features at most for each base learner. The diversity in the base learners and the use of a low number of features allow the random hyperboxes to obtain better performance across the maximum hyperbox size values. Because each base learner in the random hyperboxes model uses a much smaller number of features compared to the IOL-GFMM model trained using all features, training time and testing time of the random hyperboxes is faster than that of the IOL-GFMM model. The training and testing time of each classifier is given in Tables \ref{training_time} and \ref{testing_time}. Fast training and testing time along with better accuracy confirm the efficiency of the ensemble model in comparison to the single model using the same learning algorithm. 

\begin{table*}[!ht]
	\centering
	\scriptsize{
		\caption{Training Time (s) of the IOL-GFMM and Random Hyperboxes Model on the High Dimensional Datasets} \label{training_time}
		\begin{tabular}{|l|l|c|c|c|c|c|c|}
			\hline
			\textbf{Dataset} & \textbf{Algorithm} & \textbf{$\theta = 0.1$} & \textbf{$\theta = 0.2$} & \textbf{$\theta = 0.3$} & \textbf{$\theta = 0.4$} & \textbf{$\theta = 0.5$} & \textbf{$\theta = 0.6$} \\ \hline \hline
			\multirow{2}{*}{PEMS database}            & IOL-GFMM           & 51.3784                                      & 56.5849                                      & 52.6432                                      & 52.12905                                     & 56.7359                                      & 57.1392                                      \\ \cline{2-8} 
			& Random hyperboxes  & 26.2364                                      & 26.4292                                      & 27.0474                                      & 27.3853                                      & 29.3139                                      & 28.7593                                      \\ \hline
			\multirow{2}{*}{Complex Hydraulic System} & IOL-GFMM           & 2093.5169                                    & 2235.3104                                    & 2045.8519                                    & 1914.7439                                    & 1987.5575                                    & 1785.5609                                    \\ \cline{2-8} 
			& Random hyperboxes  & 154.9104                                     & 125.8966                                     & 100.0234                                     & 84.0987                                      & 75.5298                                      & 66.7039                                      \\ \hline
		\end{tabular}
	}
\end{table*}

\begin{table*}[!ht]
	\centering
	\scriptsize{
		\caption{Testing Time (s) of the IOL-GFMM and Random Hyperboxes Model on the High Dimensional Datasets} \label{testing_time}
		\begin{tabular}{|l|l|c|c|c|c|c|c|}
			\hline
			\textbf{Dataset} & \textbf{Algorithm} & \textbf{$\theta = 0.1$} & \textbf{$\theta = 0.2$} & \textbf{$\theta = 0.3$} & \textbf{$\theta = 0.4$} & \textbf{$\theta = 0.5$} & \textbf{$\theta = 0.6$} \\ \hline \hline
			\multirow{2}{*}{PEMS database}            & IOL-GFMM           & 121.2674                                     & 126.1965                                     & 121.4517                                     & 122.0169                                     & 126.4106                                     & 126.3136                                     \\ \cline{2-8} 
			& Random hyperboxes  & 11.3272                                      & 11.3308                                      & 12.8158                                      & 11.6774                                      & 12.3205                                      & 10.8228                                      \\ \hline
			\multirow{2}{*}{Complex Hydraulic System} & IOL-GFMM           & 1440.4623                                    & 1506.2449                                    & 1467.3662                                    & 1357.6034                                    & 1277.8380                                    & 1083.6029                                    \\ \cline{2-8} 
			& Random hyperboxes  & 118.4271                                     & 69.8559                                      & 44.8562                                      & 29.9445                                      & 23.3218                                      & 17.1749                                      \\ \hline
		\end{tabular}
	}
\end{table*}

\subsection{Supplementary Part for Analyzing the Roles of the Number of Base Learners and Maximum Number of Used Features in the Random Hyperboxes models}\label{sup_sen_params}
This part provides some supplementary figures for subsection IV.A.2 from the main paper. This experiment was performed on eight different datasets with diversity in the numbers of samples, features, and classes, i.e., \textit{plant\_species\_leaves\_margin}, \textit{plant\_species\_leaves\_shape}, \textit{movement\_libras}, \textit{connectionist\_bench\_sonar}, \textit{vehicle\_sihouettes}, \textit{breast\_cancer\_wisconsin}, \textit{heart}, and \textit{vowel}. The purpose of this experiment is to study the impacts of the number of base learners and the maximum number of used features on the classification performance of the random hyperboxes model.

\begin{figure}[!ht]
	\centering
	\begin{subfloat}[Plant\_species\_leaves\_margin]{
			\includegraphics[width=0.4\textwidth, height=0.2\textheight]{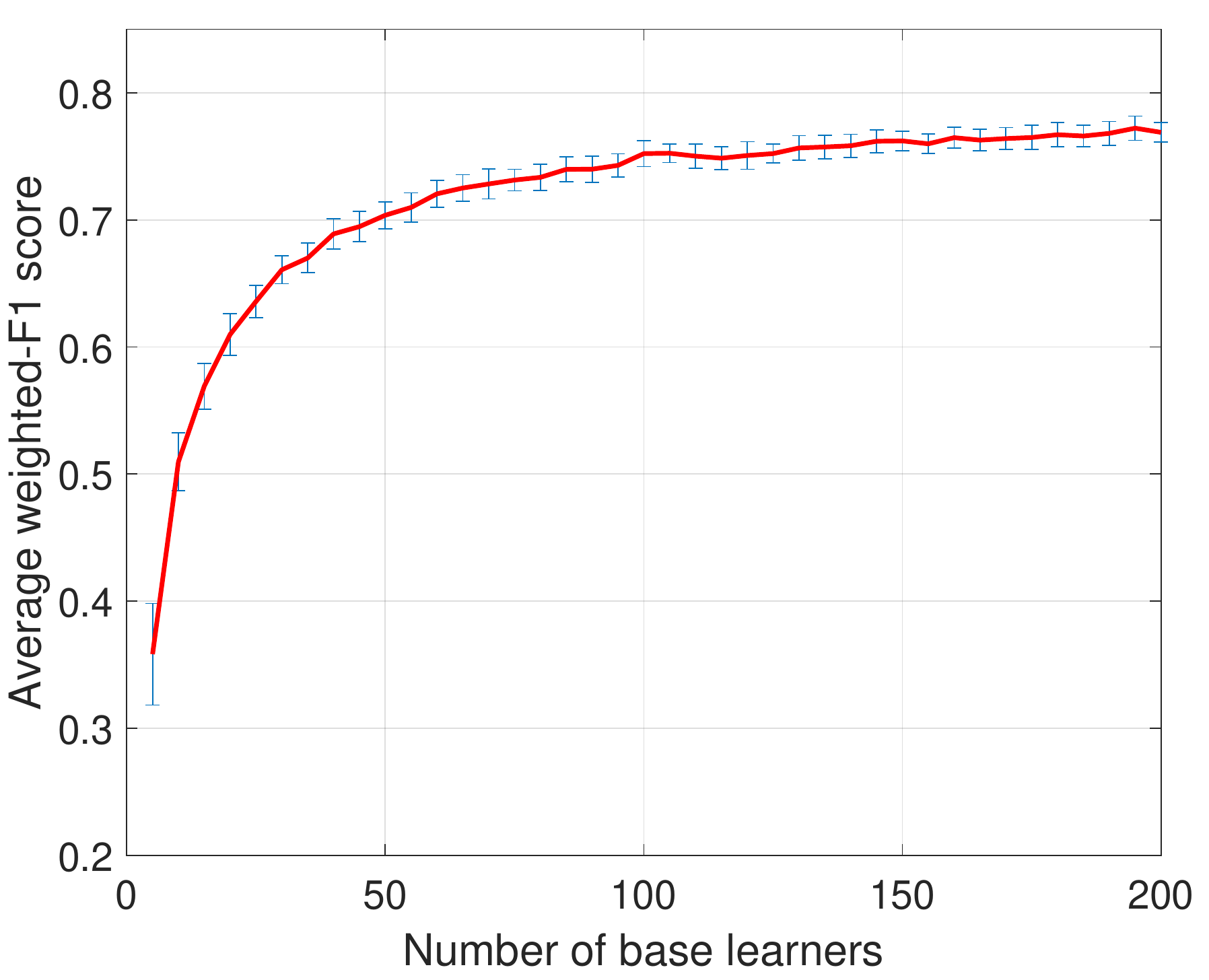}}
	\end{subfloat}
	\begin{subfloat}[Plant\_species\_leaves\_shape]{
			\includegraphics[width=0.4\textwidth, height=0.2\textheight]{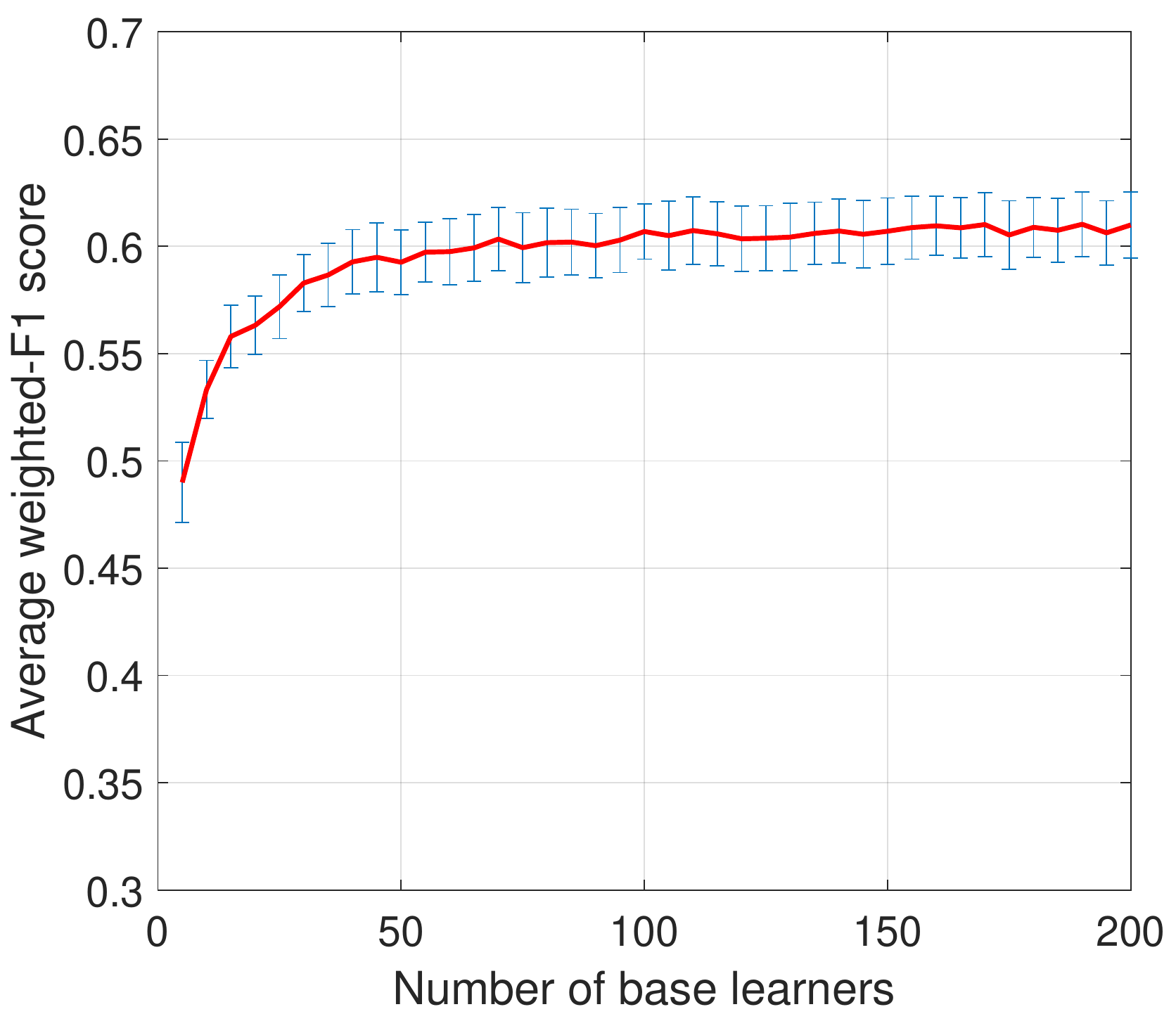}}
	\end{subfloat}
	\begin{subfloat}[Heart]{
			\includegraphics[width=0.4\textwidth, height=0.2\textheight]{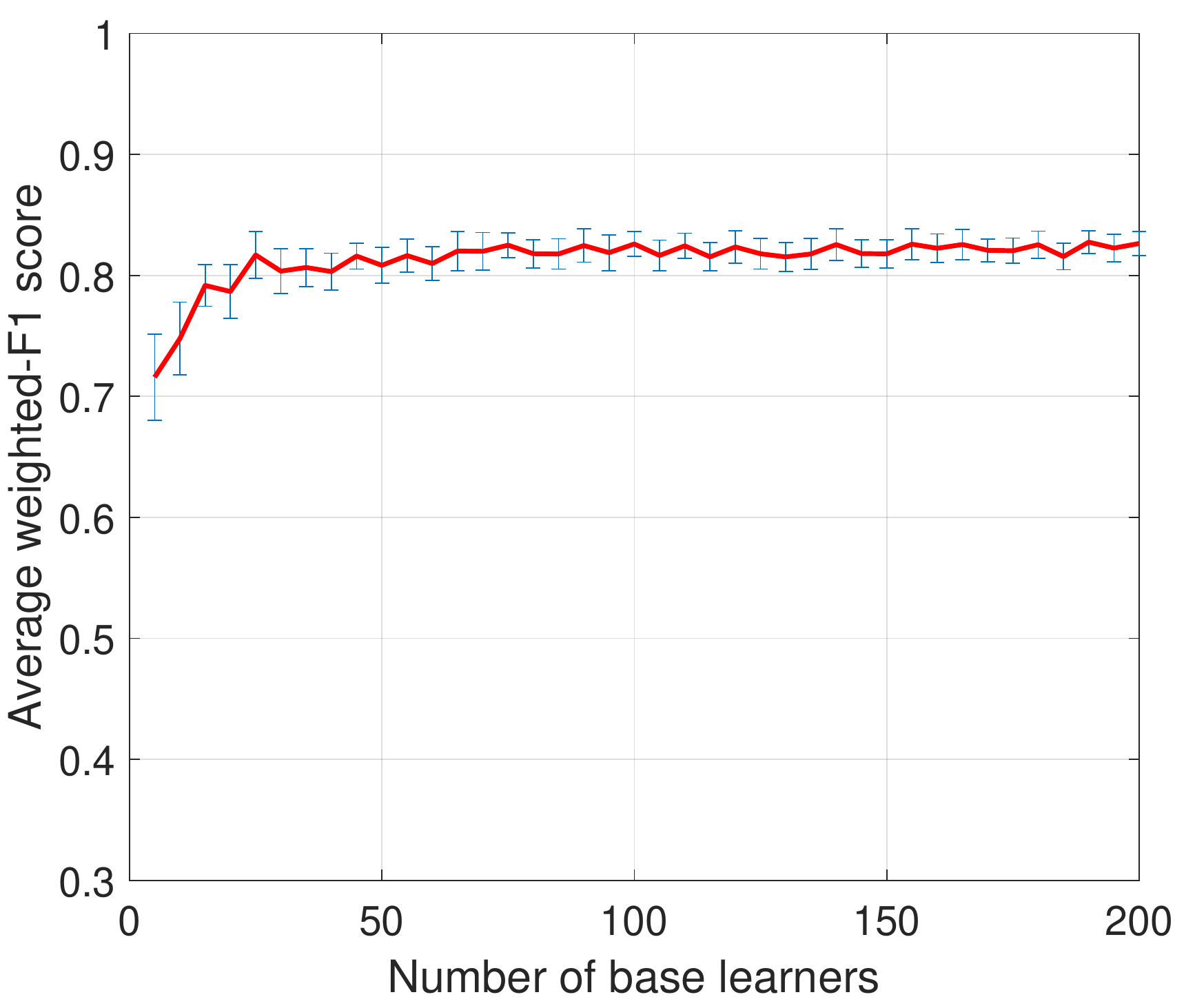}}
	\end{subfloat}
	\begin{subfloat}[Vowel]{
			\includegraphics[width=0.4\textwidth, height=0.2\textheight]{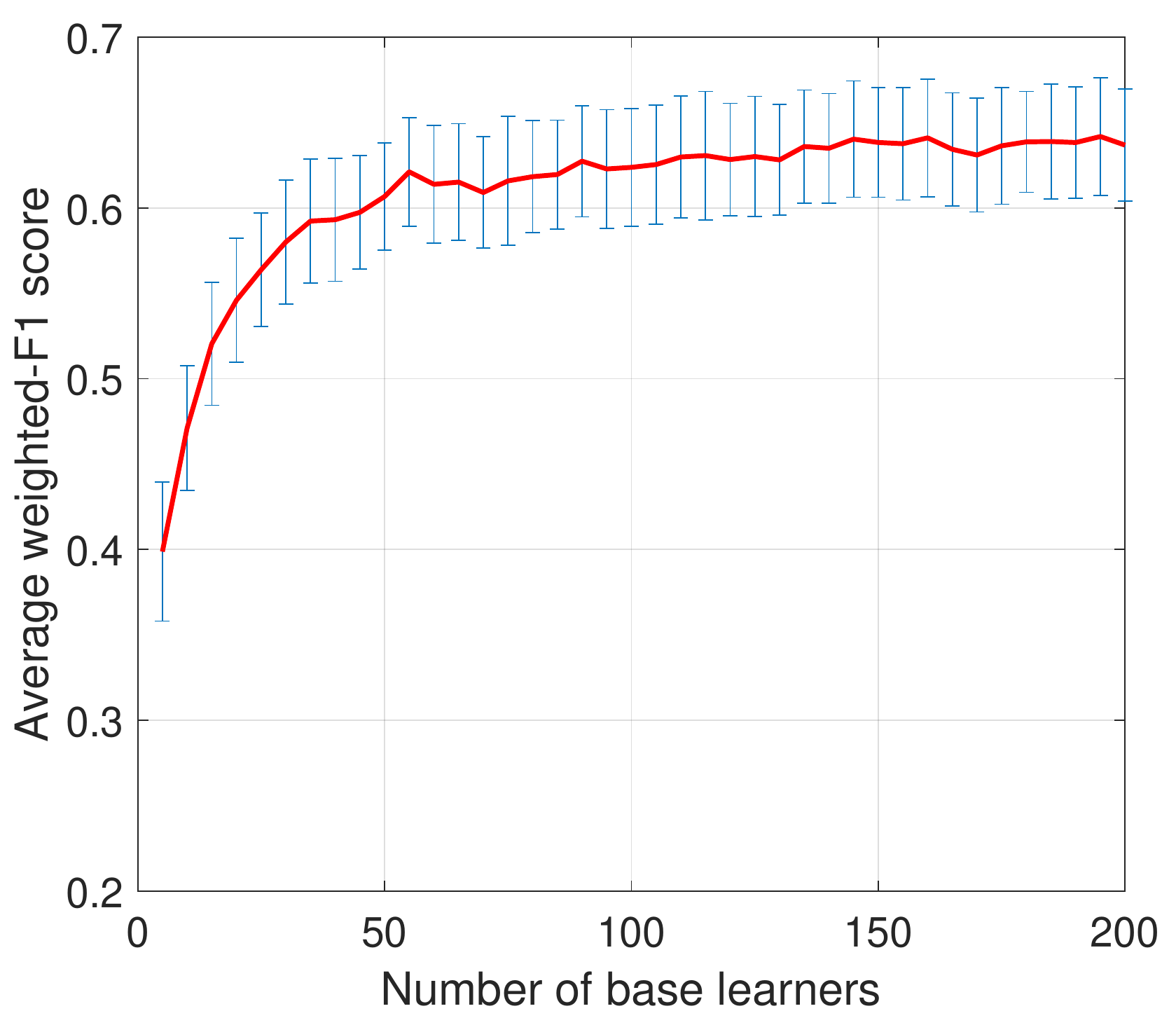}}
	\end{subfloat}
	\begin{subfloat}[Movement\_libras]{
			\includegraphics[width=0.4\textwidth, height=0.2\textheight]{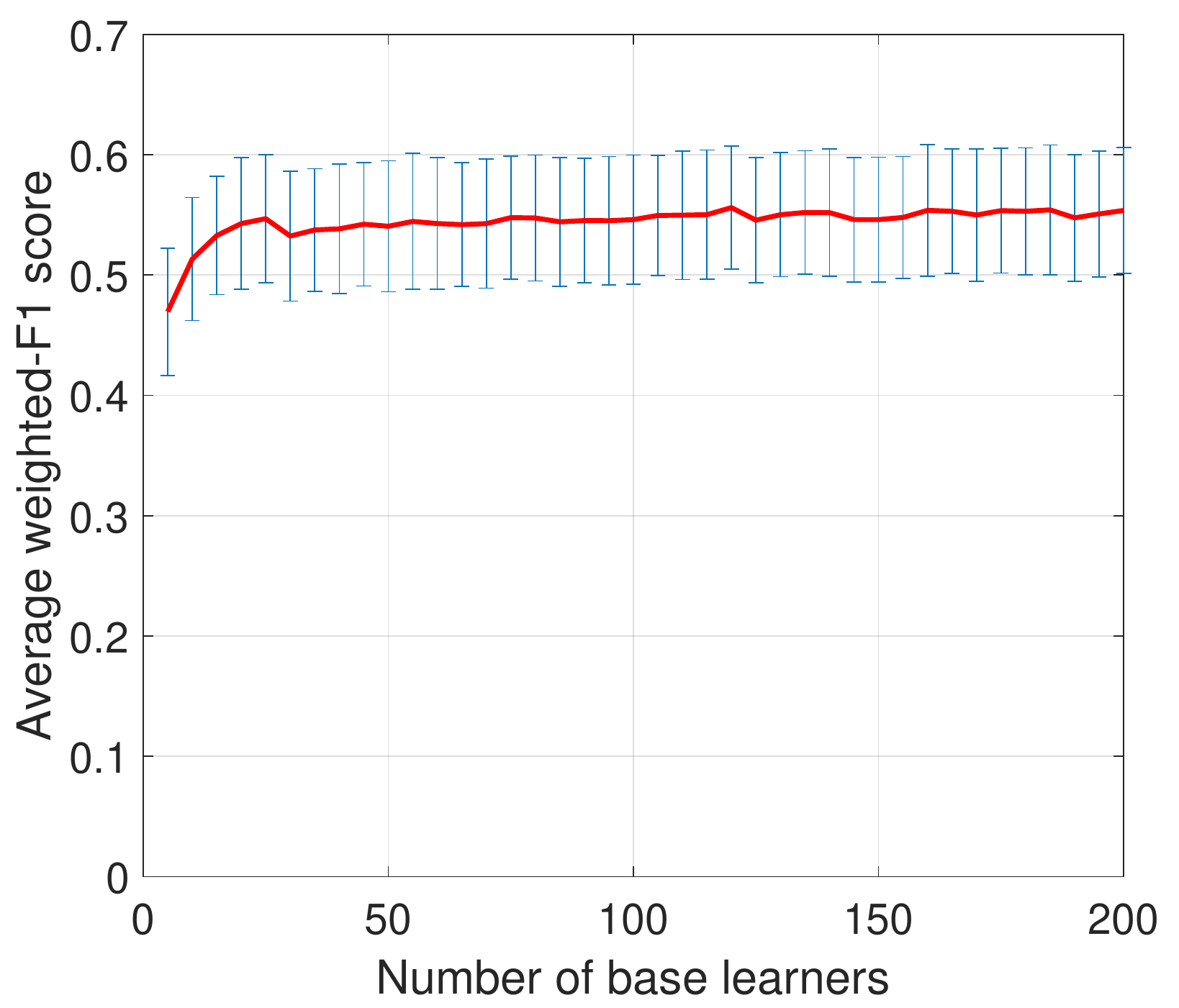}}
	\end{subfloat}
	\begin{subfloat}[Connectionist\_bench\_sonar]{
			\includegraphics[width=0.4\textwidth, height=0.2\textheight]{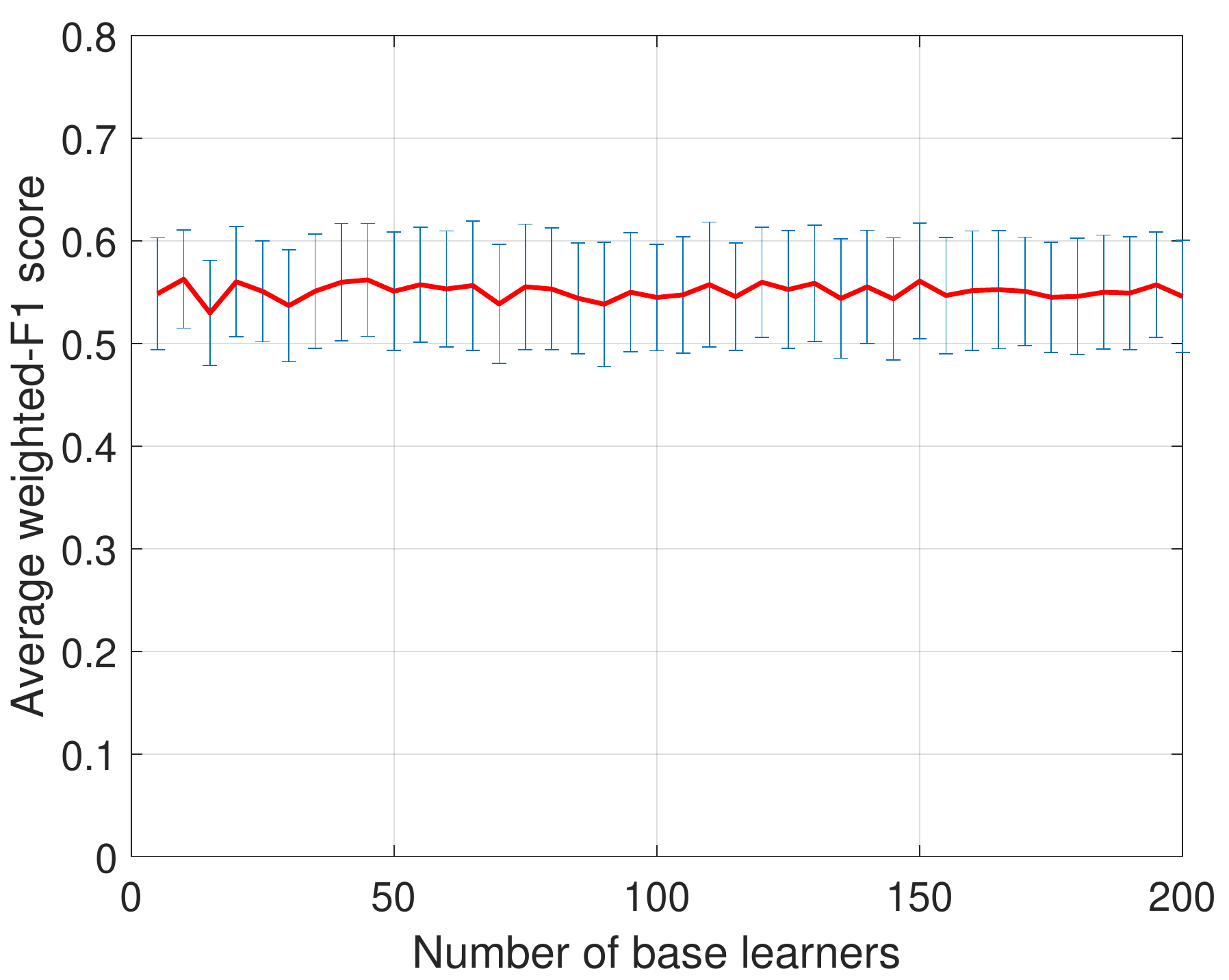}}
	\end{subfloat}
	\begin{subfloat}[Vehicle\_silhouettes]{
			\includegraphics[width=0.4\textwidth, height=0.2\textheight]{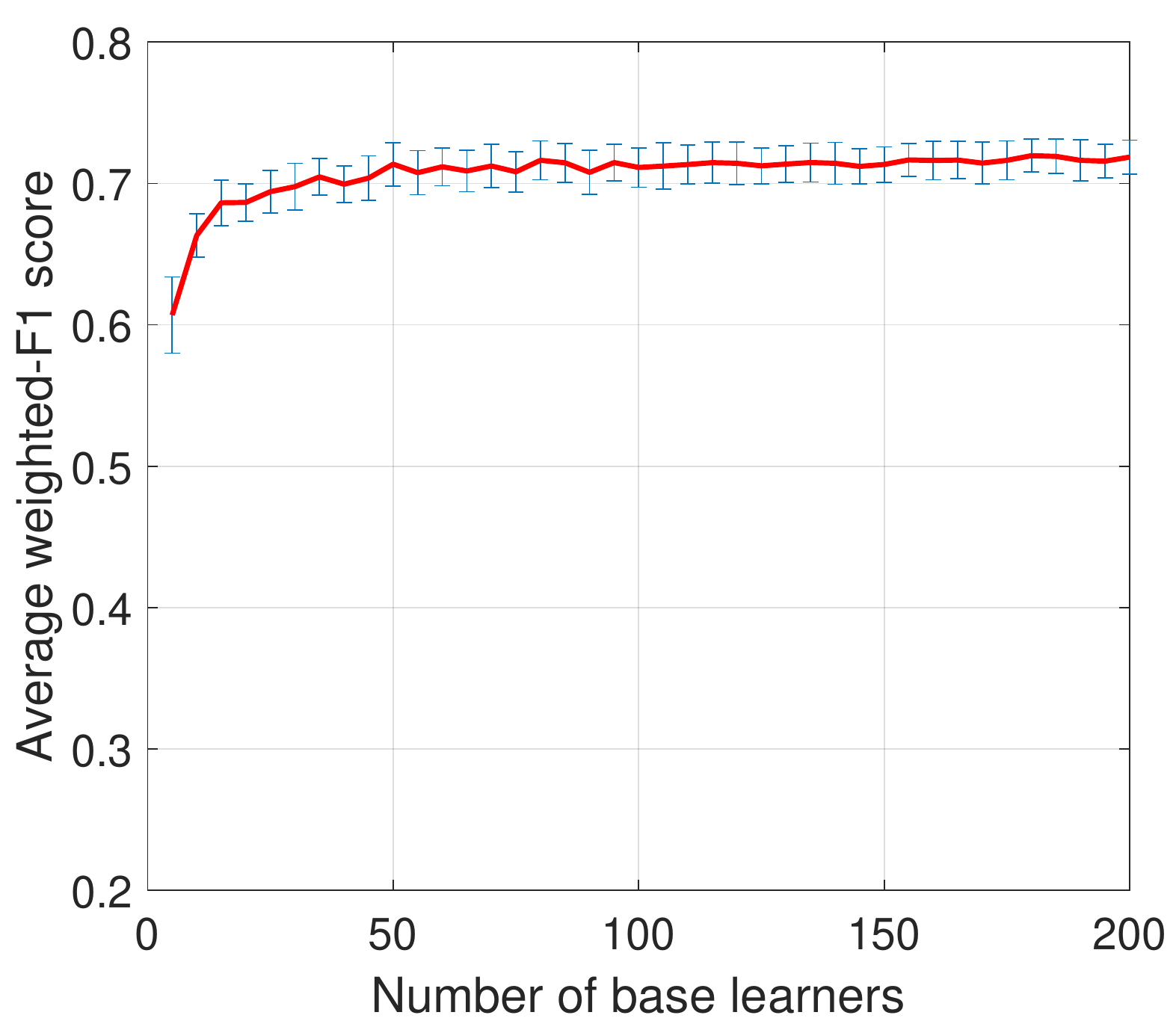}}
	\end{subfloat}
	\begin{subfloat}[Breast\_cancer\_wisconsin]{
			\includegraphics[width=0.4\textwidth, height=0.2\textheight]{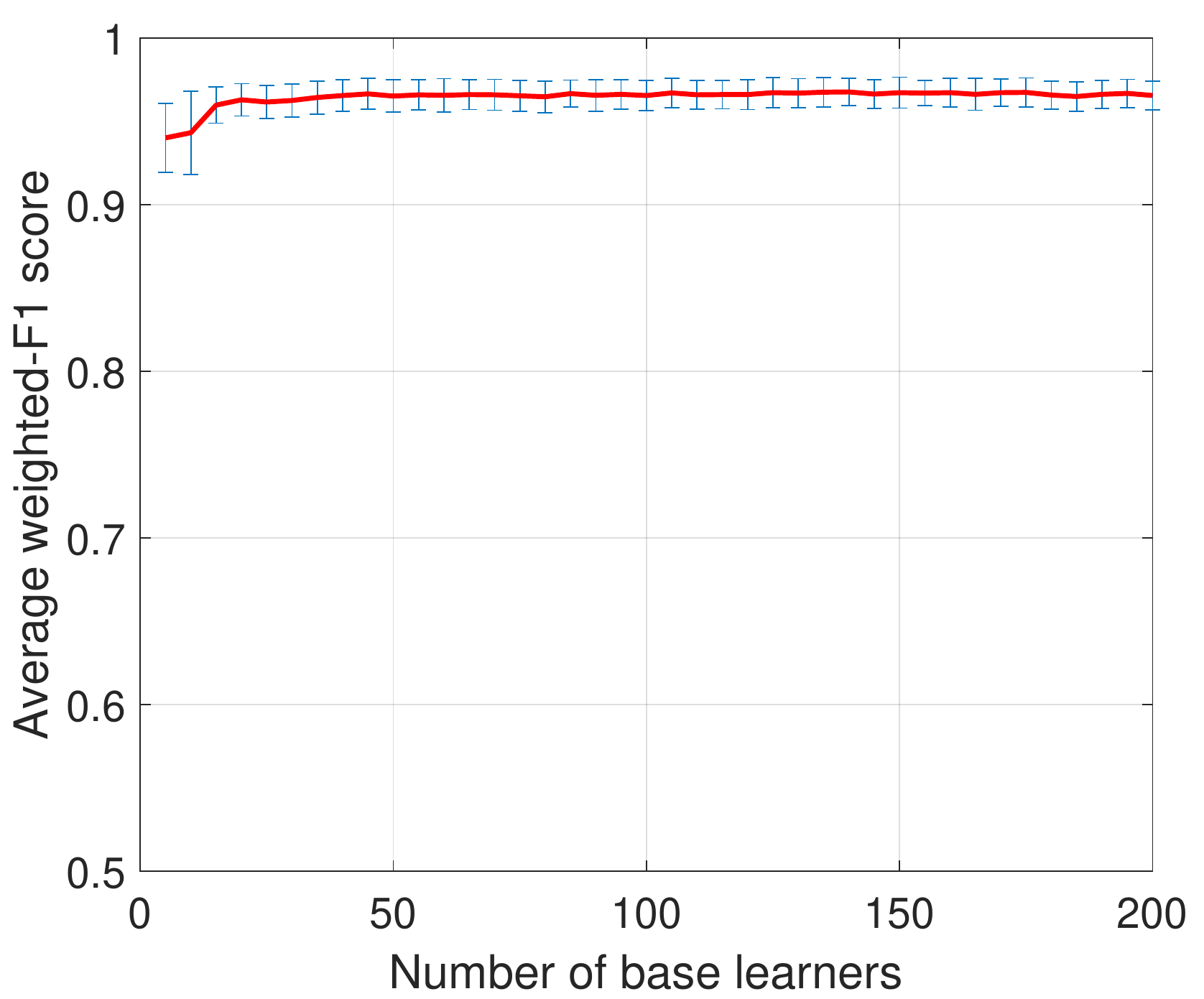}}
	\end{subfloat}
	\caption{The change in the average weighted-F1 scores when increasing the number of base learners for different datasets.}
	\label{Fig_S7}
\end{figure}

Fig. \ref{Fig_S7} shows the change in the average weighted-F1 score when we increase the number of base estimators. We can observe a general trend over all experimental datasets which is that the increase in the number of base learners does not lead to the decrease in the classification accuracy. These empirical results are consistent with the statements in the theoretical part (section III.C.1) from the main paper.

\begin{figure}[!ht]
	\centering
	\begin{subfloat}[Plant\_species\_leaves\_margin]{
			\includegraphics[width=0.4\textwidth, height=0.2\textheight]{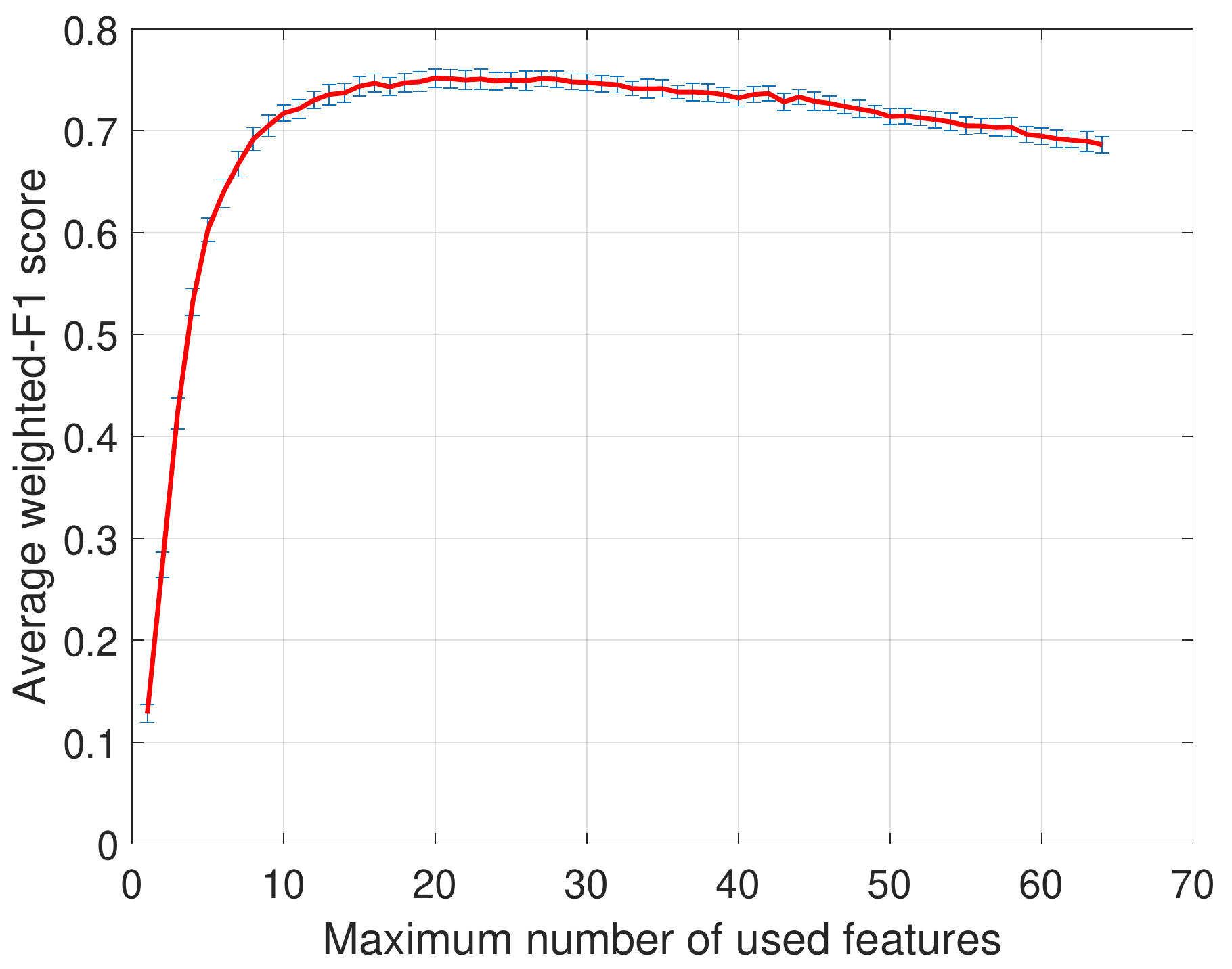}}
	\end{subfloat}
	\begin{subfloat}[Plant\_species\_leaves\_shape]{
			\includegraphics[width=0.4\textwidth, height=0.2\textheight]{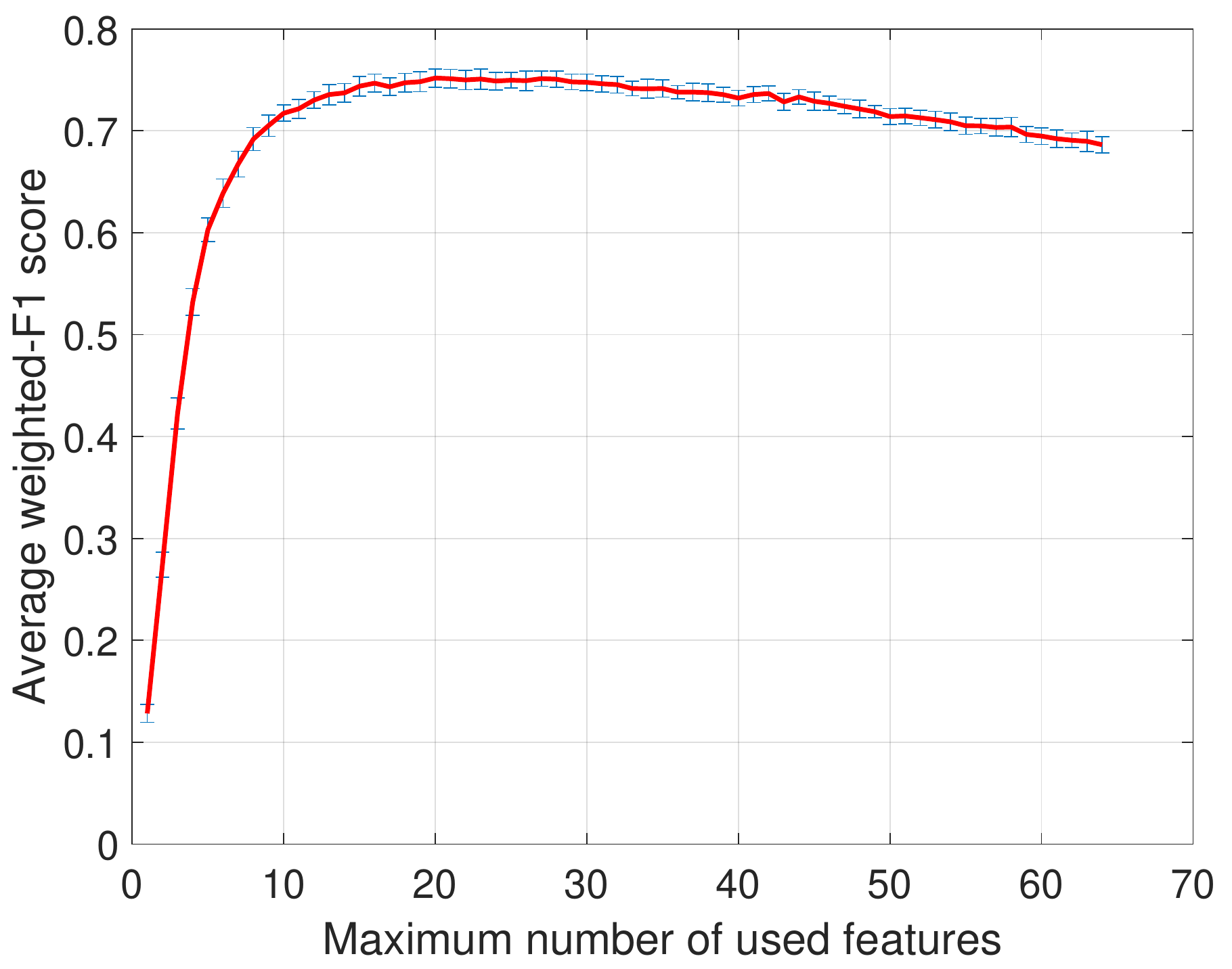}}
	\end{subfloat}
	\begin{subfloat}[Heart]{
			\includegraphics[width=0.4\textwidth, height=0.2\textheight]{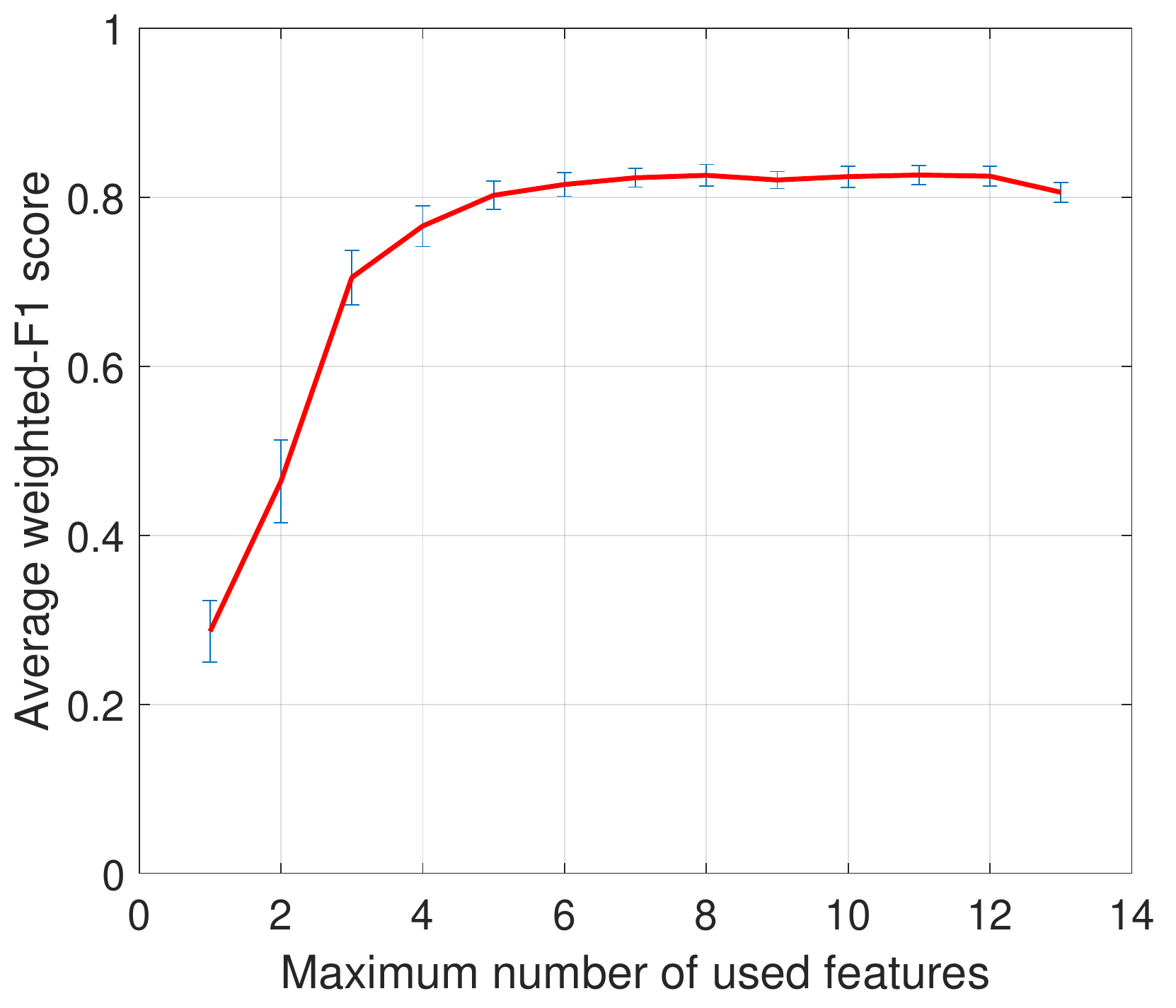}}
	\end{subfloat}
	\begin{subfloat}[Vowel]{
			\includegraphics[width=0.4\textwidth, height=0.2\textheight]{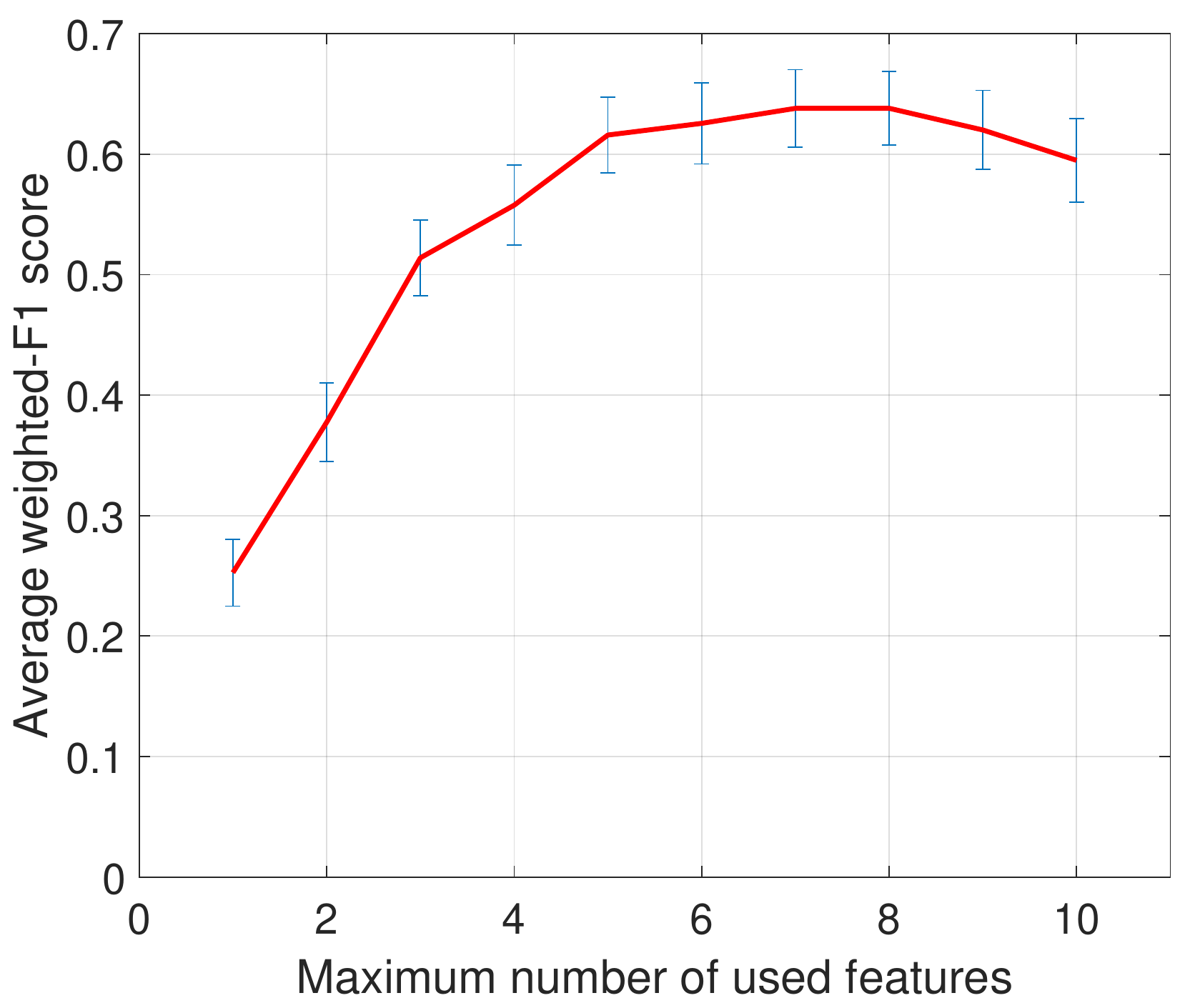}}
	\end{subfloat}
	\begin{subfloat}[Movement\_libras]{
			\includegraphics[width=0.4\textwidth, height=0.2\textheight]{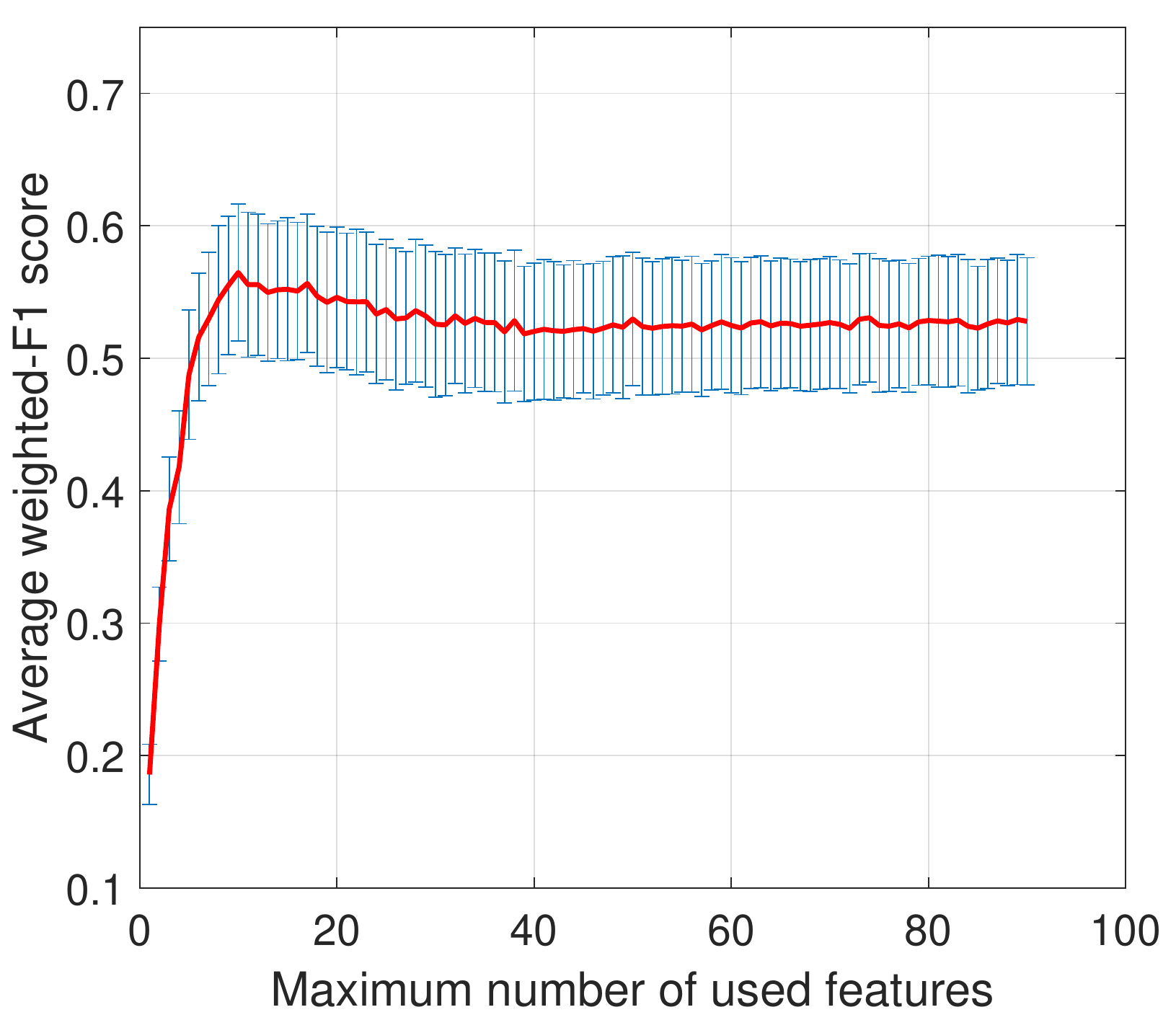}}
	\end{subfloat}
	\begin{subfloat}[Connectionist\_bench\_sonar]{
			\includegraphics[width=0.4\textwidth, height=0.2\textheight]{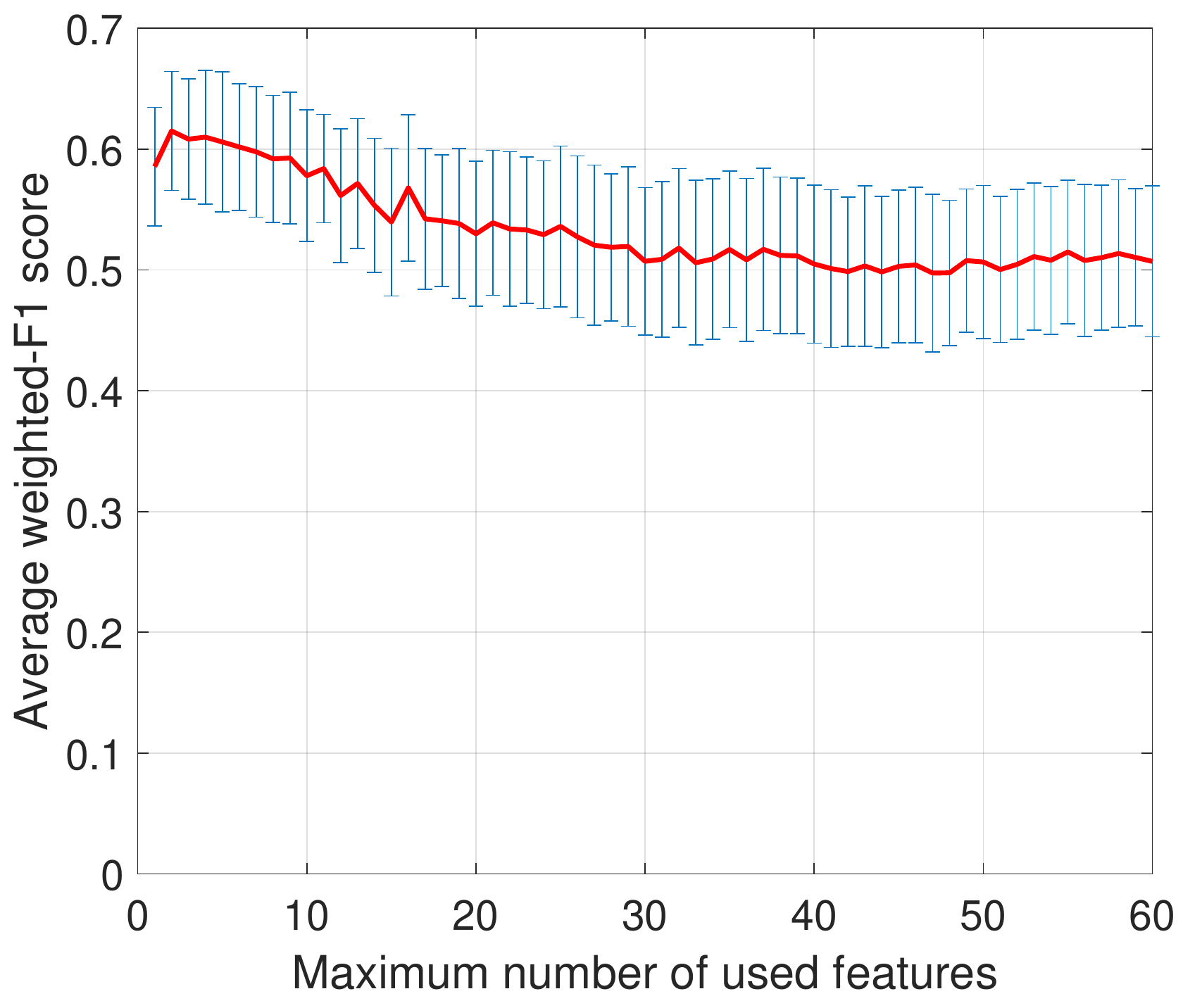}}
	\end{subfloat}
	\begin{subfloat}[Vehicle\_silhouettes]{
			\includegraphics[width=0.4\textwidth, height=0.2\textheight]{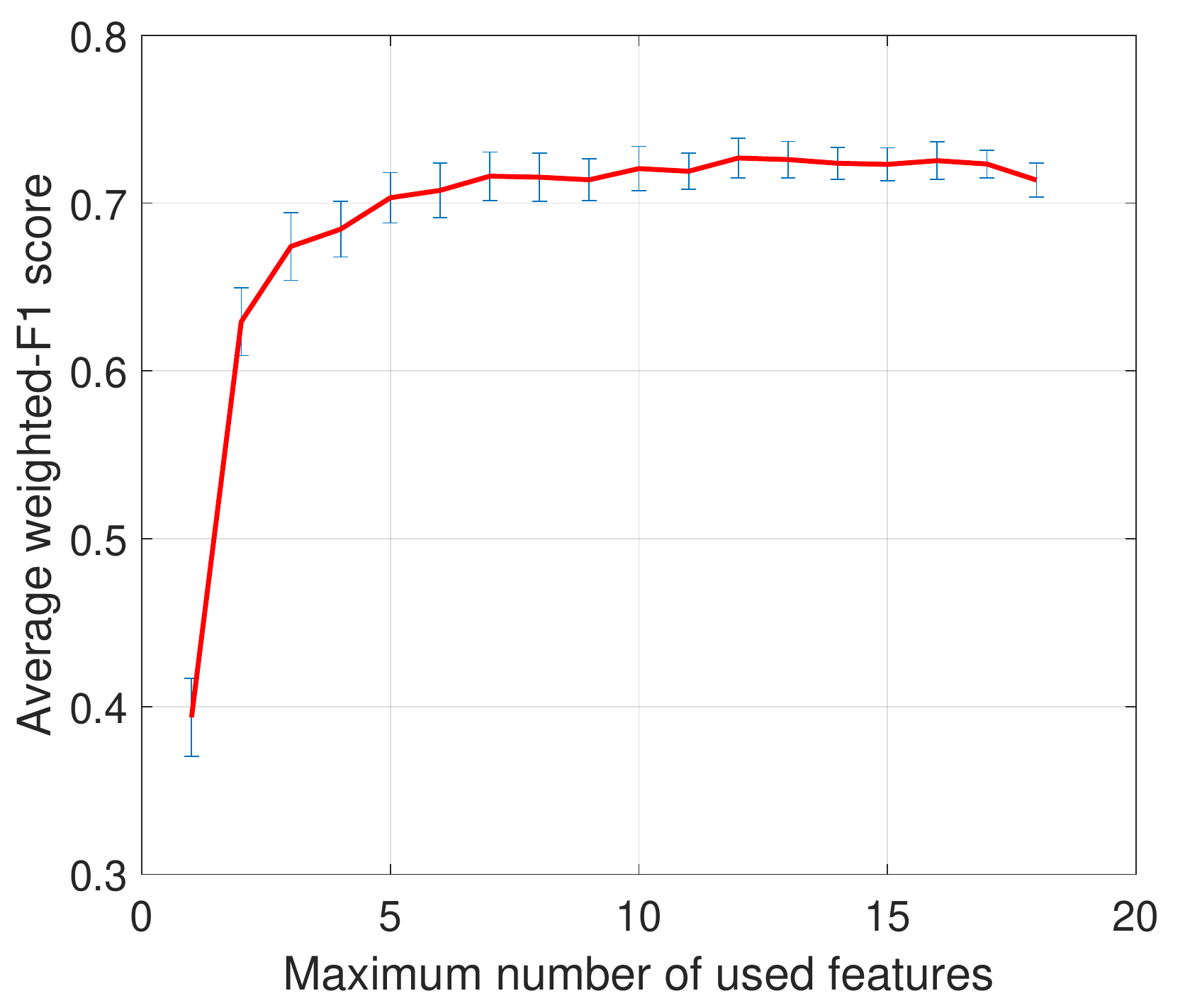}}
	\end{subfloat}
	\begin{subfloat}[Breast\_cancer\_wisconsin]{
			\includegraphics[width=0.4\textwidth, height=0.2\textheight]{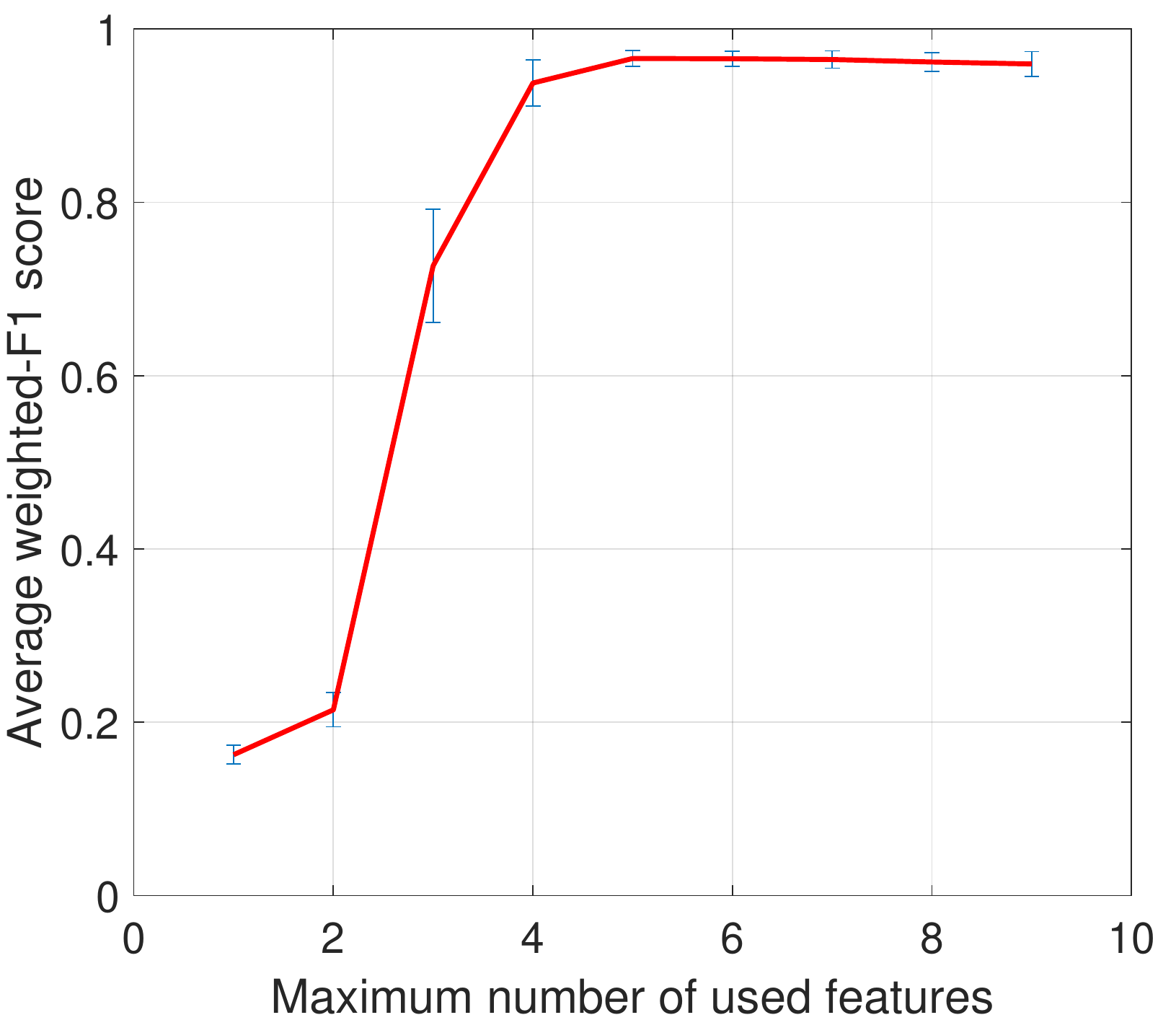}}
	\end{subfloat}
	\caption{The change in the average weighted-F1 scores when increasing the maximum number of used dimensions for different datasets.}
	\label{Fig_S8}
\end{figure}

Fig. \ref{Fig_S8} presents the change in the classification performance when the maximum number of used features increases. A general trend can be observed in which the classification accuracy only increases up to a certain value of the maximum number of used features, and then decreases if the maximum number of features available for the base classifiers is increased. The reason for this trend is explained by the correlation between base learners as shown in subsection IV.A.2 from the main paper.

\subsection{Comparing the Performance of the Random Hyperboxes to Other Classifiers}\label{sup_cmp_other}
\subsubsection{Datasets and Parameter Settings}\label{sup_dataset} \hfill

In this paper, we used 20 datasets with diversity in the numbers of samples, features, and classes taken from the UCI repository \cite{Dua2019}. Table \ref{dataset_tab} summarizes the information of these datasets. Each dataset is normalized to the range of [0, 1] according to the requirement of the fuzzy min-max neural networks. The experiments were executed on the computer using Red Hat Enterprise Linux 7.5 with Intel Xeon Gold 6150 2.7GHz CPU and 64GB RAM.

\begin{table}[!ht]
	\centering
	\caption{The Descriptions of the Used Datasets}
	\label{dataset_tab}
	\scriptsize{
		\begin{tabular}{|l|l|c|c|c|}
			\hline
			\textbf{ID} & \textbf{Dataset}               & \textbf{\# samples} & \textbf{\# features} & \textbf{\# classes} \\ \hline \hline
			1           & Balance\_scale                 & 625                 & 4                    & 3                   \\ \hline
			2           & banknote\_authentication       & 1372                & 4                    & 2                   \\ \hline
			3           & blood\_transfusion             & 748                 & 4                    & 2                   \\ \hline
			4           & breast\_cancer\_wisconsin      & 699                 & 9                    & 2                   \\ \hline
			5           & BreastCancerCoimbra            & 116                 & 9                    & 2                   \\ \hline
			6           & connectionist\_bench\_sonar    & 208                 & 60                   & 2                   \\ \hline
			7           & haberman                       & 306                 & 3                    & 2                   \\ \hline
			8           & heart                          & 270                 & 13                   & 2                   \\ \hline
			9           & movement\_libras               & 360                 & 90                   & 15                  \\ \hline
			10          & pima\_diabetes                 & 768                 & 8                    & 2                   \\ \hline
			11          & plant\_species\_leaves\_margin & 1600                & 64                   & 100                 \\ \hline
			12          & plant\_species\_leaves\_shape  & 1600                & 64                   & 100                 \\ \hline
			13          & ringnorm                       & 7400                & 20                   & 2                   \\ \hline
			14          & landsat\_satellite                        & 6435                & 36                   & 6                   \\ \hline
			15          & twonorm                        & 7400                & 20                   & 2                   \\ \hline
			16          & vehicle\_silhouettes                        & 846                 & 18                   & 4                   \\ \hline
			17          & vertebral\_column              & 310                 & 6                    & 3                   \\ \hline
			18          & vowel                          & 990                 & 10                   & 11                  \\ \hline
			19          & waveform                       & 5000                & 21                   & 3                   \\ \hline
			20          & wireless\_indoor\_localization & 2000                & 7                    & 4                   \\ \hline
		\end{tabular}
	}
\end{table}

For experiments, the maximum hyperbox size of based learners in the random hyperboxes model, as well as different types of FMNNs, is set to $\theta = 0.1$ and the sensitivity parameter of the membership function is fixed at $\gamma = 1$. To compare to other ensemble methods, this study deployed the threshold $2 \sqrt{p}$ for the maximum number of used features and 50\% of training samples were randomly sampled to train base learners ($r_s = 0.5$). As common settings in the random forest and ensemble classifiers literature, we set the number of base learners $m = 100$. For other parameters of classifiers, we used default settings of libraries such as scikit-learn \cite{Pedregosa2011}, XGBoost \cite{Chen2016}, LightGBM \cite{Ke2017} apart from the maximum tree depth of decision trees and tree-based ensemble methods is set to the value of 10 to prevent overfitting \cite{Bertsimas2017}. For models using a threshold value for nearest neighbors, we used $K = 5$.

We did not adjust the values of hyperparameters for models in these experiments, although we are aware that a thorough experimental comparison among approaches should tune their hyperparameters to their best for every data set. Our reasons for using the standard implementations in libraries are three-fold. First, our goal in this paper was to achieve initial analyses of the effectiveness of the random hyperboxes classifier. If it is worse than other methods with standard implementations, no further studies would be worthwhile to improve and exploit the proposed method. In the opposite case, we need a comparative study to evaluate the impacts of hyperparameters on the predictive performance of methods. Second, standard implementations in libraries are general enough to perform quite well across many problems \cite{Rodriguez2006}. The lack of fine-tuning is compensated by the diversity in the number of features, samples, and classes of the used data sets. These datasets are quite common and were randomly chosen without intentionally favoring any learning algorithms. Moreover, the performance of models is also assessed using 10 times repeated 4-fold cross-validation along with statistical testing methods. Third, the use of default parameters without tuning will be easily reproducible by other studies.

\subsubsection{A Comparison of the Random Hyperboxes With Other FMNNs} \label{sup_cmp_other_fmnn} \hfill

This subsection provides the results of the average weighted-F1 scores of fuzzy min-max classifiers mentioned in subsection IV.B.1 from the main paper. Among different types of fuzzy min-max neural networks, IOL-GFMM and RFMNN have mechanisms to make the decision when there are at least two winning hyperboxes representing different classes (in this case, the sample is located on the decision boundary). Therefore, to make a fair comparison, other fuzzy min-max classifiers used the Manhattan distance from the input pattern to central points of winning hyperboxes to find the predictive class instead of randomly selecting a class. We have implemented all of these fuzzy min-max neural networks in Python.

The average weighted-F1 scores of classifiers using 10 times repeated 4-fold cross-validation are shown in Table \ref{cmp_other_fmnn_01} for the maximum hyperbox size $\theta = 0.1$ and Table \ref{cmp_other_fmnn_07} for $\theta = 0.7$. To facilitate the process of evaluating the performance and performing statistical testing, the performance of classifiers on each dataset is ranked, in which the best classifier with the highest average weighted-F1 score is ranked first, and the second-best classifier is ranked two and so on. The classifiers with the same average weighted-F1 scores are assigned the average value of their ranks. Table \ref{cmp_other_fmnn_01_ranking} shows the ranks of classifiers using $\theta = 0.1$, while Table \ref{cmp_other_fmnn_07_ranking} presents the ranks of classifiers with $\theta = 0.7$.

It can be seen that the random hyperboxes classifier achieves the lowest rank for both high and low values of $\theta$. Its average ranks are twice as low as those of the second-best classifiers. In addition, the random hyperboxes classifier obtains the highest average weighted-F1 scores on almost all considered datasets. These figures show the superior performance of the random hyperboxes classifier in comparison to other types of fuzzy min-max neural networks.

\begin{table*}[!ht]
	\centering
	\caption{The Average Weighted-F1 Scores and Standard Deviation of the Random Hyperboxes (RH) and Other Fuzzy Min-Max Neural Networks ($\theta=0.1$)} \label{cmp_other_fmnn_01}
	\scriptsize{
		\begin{tabular}{|c|l|R{1cm}|R{1cm}|R{1cm}|R{1cm}|R{1cm}|R{1cm}|R{1cm}|R{1.2cm}|R{1.6cm}|}
			\hline
			\textbf{ID} & \textbf{Dataset}               & \textbf{IOL-GFMM} & \textbf{Onln-GFMM} & \textbf{RH} & \textbf{FMNN}    & \textbf{EFMNN}   & \textbf{KNEFMNN} & \textbf{RFMNN}   & \textbf{AGGLO-2} & \textbf{Onln-GFMM + AGGLO-2} \\ \hline \hline
			1                             & Balance\_scale                    & \textbf{0.79383 $\pm$ 0.0558} & 0.73358 $\pm$ 0.0857                & 0.73821 $\pm$ 0.0894                           & 0.73823 $\pm$ 0.0635                           & 0.75990 $\pm$ 0.0800                           & 0.75990 \break$\pm$ 0.0800                           & 0.75990 $\pm$ 0.0800                           & \textbf{0.79383 \break$\pm$ 0.0558} & \textbf{0.79383 \break $\pm$ 0.0558} \\ \hline
			2                            & banknote\_authentication          & 0.99782 $\pm$ 0.0013                           & 0.99709 $\pm$ 0.0021                & 0.99818 $\pm$ 0.0018                & 0.99854 $\pm$ 0.0015                           & \textbf{0.99927 $\pm$ 0.0013} & 0.99854 \break$\pm$ 0.0015                           & \textbf{0.99927 $\pm$ 0.0013} & 0.99854 \break$\pm$ 0.0015                           & 0.99854 \break$\pm$ 0.0015                           \\ \hline
			3                            & blood\_transfusion                & 0.56827 $\pm$ 0.1417                           & 0.60796 $\pm$ 0.0623                & \textbf{0.68264 $\pm$ 0.0190} & 0.62517 $\pm$ 0.0596                           & 0.63917 $\pm$ 0.0443                           & 0.64600 \break$\pm$ 0.0520                             & 0.63505 $\pm$ 0.0639                           & 0.64685 \break$\pm$ 0.0646                           & 0.58389 \break$\pm$ 0.1132                           \\ \hline
			4                            & breast\_cancer\_wisconsin         & 0.94383 $\pm$ 0.0329                           & 0.94827 $\pm$ 0.0306                & \textbf{0.96685 $\pm$ 0.0180} & 0.95840 $\pm$ 0.0229                           & 0.95976 $\pm$ 0.0244                           & 0.95976 \break$\pm$ 0.0244                           & 0.95976 $\pm$ 0.0243                           & 0.94383 \break$\pm$ 0.0329                           & 0.94383 \break$\pm$ 0.0329                           \\ \hline
			5                            & BreastCancerCoimbra               & 0.62798 $\pm$ 0.0381                           & 0.62798 $\pm$ 0.0381                & \textbf{0.66561 $\pm$ 0.0939} & 0.64246 $\pm$ 0.1193                           & 0.60216 $\pm$ 0.0401                           & 0.60216 \break$\pm$ 0.0401                           & 0.60216 $\pm$ 0.0401                           & 0.62798 \break$\pm$ 0.0381                           & 0.62798 \break$\pm$ 0.0381                           \\ \hline
			6                            & connectionist\_bench\_sonar       & 0.53013 $\pm$ 0.0816                           & 0.53013 $\pm$ 0.0816                & 0.55964 $\pm$ 0.1085                           & \textbf{0.58747 $\pm$ 0.1217} & 0.55768 $\pm$ 0.1160                           & 0.55768 \break$\pm$ 0.1160                           & 0.55768 $\pm$ 0.1160                           & 0.53013 \break$\pm$ 0.0816                           & 0.53013 \break$\pm$ 0.0816                           \\ \hline
			7                            & haberman                          & 0.61185 $\pm$ 0.0418                           & 0.61711 $\pm$ 0.0387                & \textbf{0.64211 $\pm$ 0.0294} & 0.60223 $\pm$ 0.0168                           & 0.63494 $\pm$ 0.0291                           & 0.63076 \break$\pm$ 0.0172                           & 0.62178 $\pm$ 0.0450                           & 0.60629 \break$\pm$ 0.0601                           & 0.58827 \break$\pm$ 0.0710                           \\ \hline
			8                            & heart                             & 0.76101 $\pm$ 0.0272                           & 0.75470 $\pm$ 0.0125                & \textbf{0.82643 $\pm$ 0.0252} & 0.79486 $\pm$ 0.0112                           & 0.77308 $\pm$ 0.0295                           & 0.77308 \break$\pm$ 0.0295                           & 0.77308 $\pm$ 0.0295                           & 0.76101 \break$\pm$ 0.0272                           & 0.76101 \break$\pm$ 0.0272                           \\ \hline
			9                            & movement\_libras                  & 0.53032 $\pm$ 0.0796                           & 0.53032 $\pm$ 0.0796                & \textbf{0.54465 $\pm$ 0.1064} & 0.47268 $\pm$ 0.1048                           & 0.49732 $\pm$ 0.0879                           & 0.49732 \break$\pm$ 0.0879                           & 0.49732 $\pm$ 0.0879                           & 0.53255 \break$\pm$ 0.0796                           & 0.53255 \break$\pm$ 0.0796                           \\ \hline
			10                           & pima\_diabetes                    & 0.70322 $\pm$ 0.0113                           & 0.69864 $\pm$ 0.0118                & \textbf{0.72760 $\pm$ 0.0339} & 0.71264 $\pm$ 0.0182                           & 0.71634 $\pm$ 0.0356                           & 0.72053 \break$\pm$ 0.0284                           & 0.71634 $\pm$ 0.0356                           & 0.70372 \break$\pm$ 0.0164                           & 0.70216 \break$\pm$ 0.0151                           \\ \hline
			11                           & plant\_species\_leaves\_margin    & 0.57408 $\pm$ 0.0217                           & 0.58113 $\pm$ 0.0251                & 0.74748 $\pm$ 0.0195                           & 0.67553 $\pm$ 0.0107                           & \textbf{0.76291 \break$\pm$ 0.0090} & \textbf{0.76291 \break$\pm$ 0.0090} & \textbf{0.76291 $\pm$ 0.0090} & 0.57408 \break$\pm$ 0.0217                           & 0.57408 \break$\pm$ 0.0217                           \\ \hline
			12                           & plant\_species\_leaves\_shape     & 0.53546 $\pm$ 0.0218                           & 0.55296 $\pm$ 0.0193                & \textbf{0.60695 $\pm$ 0.0306} & 0.49222 $\pm$ 0.0242                           & 0.48698 $\pm$ 0.0324                           & 0.51801 \break$\pm$ 0.0426                           & 0.48154 $\pm$ 0.0340                           & 0.55600 \break$\pm$ 0.0395                           & 0.56172 \break$\pm$ 0.0323                           \\ \hline
			13                           & ringnorm                          & 0.62981 $\pm$ 0.0070                           & 0.62981 $\pm$ 0.0070                & \textbf{0.94237 $\pm$ 0.0057} & 0.79121 $\pm$ 0.0067                           & 0.66059 $\pm$ 0.0156                           & 0.60626 \break$\pm$ 0.0135                           & 0.66059 \break$\pm$ 0.0156                           & 0.63184 \break$\pm$ 0.0045                           & 0.63184 \break$\pm$ 0.0045                           \\ \hline
			14                           & landsat\_satellite                & 0.86259 $\pm$ 0.0179                           & 0.86207 $\pm$ 0.0191                & \textbf{0.87875 $\pm$ 0.0076} & 0.80895 $\pm$ 0.0405                           & 0.86752 $\pm$ 0.0140                           & 0.86749 \break$\pm$ 0.0165                           & 0.86792 $\pm$ 0.0117                           & 0.86631 \break$\pm$ 0.0193                           & 0.86464 \break$\pm$ 0.0163                           \\ \hline
			15                           & twonorm                           & 0.94283 $\pm$ 0.0054                           & 0.94284 $\pm$ 0.0054                & \textbf{0.97211 $\pm$ 0.0032} & 0.94824 $\pm$ 0.0060                           & 0.94932 $\pm$ 0.0062                           & 0.94932 \break$\pm$ 0.0062                           & 0.94932 $\pm$ 0.0062                           & 0.94284 \break$\pm$ 0.0054                           & 0.94284 \break$\pm$ 0.0054                           \\ \hline
			16                           & vehicle\_silhouettes              & 0.65535 $\pm$ 0.0195                           & 0.65993 $\pm$ 0.0183                & \textbf{0.72044 $\pm$ 0.0309} & 0.66694 $\pm$ 0.0112                           & 0.66552 $\pm$ 0.0246                           & 0.66715 \break$\pm$ 0.0200                           & 0.66552 $\pm$ 0.0246                           & 0.65307 \break$\pm$ 0.0197                           & 0.65432 \break$\pm$ 0.0210                           \\ \hline
			17                           & vertebral\_column                 & 0.74496 $\pm$ 0.0405                           & 0.74136 $\pm$ 0.0184                & \textbf{0.76542 $\pm$ 0.0411} & 0.69182 $\pm$ 0.0385                           & 0.74947 $\pm$ 0.0528                           & 0.76296 \break$\pm$ 0.0417                           & 0.75850 $\pm$ 0.0518                           & 0.73614 \break$\pm$ 0.0372                           & 0.72441 \break$\pm$ 0.0473                           \\ \hline
			18                           & vowel                             & 0.55279 $\pm$ 0.0491                           & 0.55279 $\pm$ 0.0491                & \textbf{0.63059 $\pm$ 0.0676} & 0.59758 $\pm$ 0.0387                           & 0.58589 $\pm$ 0.0400                           & 0.58907 \break$\pm$ 0.0387                           & 0.58589 $\pm$ 0.0400                           & 0.55507 \break$\pm$ 0.0525                           & 0.55953 \break$\pm$ 0.0547                           \\ \hline
			19                           & waveform                          & 0.78002 $\pm$ 0.0107                           & 0.78002 $\pm$ 0.0107                & \textbf{0.83795 $\pm$ 0.0068} & 0.76572 $\pm$ 0.0136                           & 0.78752 $\pm$ 0.0115                           & 0.78752 \break$\pm$ 0.0115                           & 0.78752 $\pm$ 0.0115                           & 0.78002 \break$\pm$ 0.0107                           & 0.78002 \break$\pm$ 0.0107                           \\ \hline
			20                           & wireless\_indoor\_localization    & 0.96511  $\pm$ 0.0048                          & 0.96561 $\pm$ 0.0061                & \textbf{0.97726 $\pm$ 0.0086} & 0.96266 $\pm$ 0.0181                           & 0.97252 $\pm$ 0.0058                           & 0.97403 \break$\pm$ 0.0110                           & 0.97252 $\pm$ 0.0058                           & 0.96364 \break$\pm$ 0.0087                           & 0.96565 \break$\pm$ 0.0068                           \\ \hline
		\end{tabular}
	}
\end{table*}

\begin{table*}[!ht]
	\centering
	\caption{The Ranking of the Random Hyperboxes and Other Fuzzy Min-Max Classfiers ($\theta=0.1$)} \label{cmp_other_fmnn_01_ranking}
	\scriptsize{
		\begin{tabular}{|c|l|c|c|c|c|c|c|c|c|C{1.5cm}|}
			\hline
			\textbf{ID} & \textbf{Dataset}               & \textbf{IOL-GFMM} & \textbf{Onln-GFMM} & \textbf{RH} & \textbf{FMNN}    & \textbf{EFMNN}   & \textbf{KNEFMNN} & \textbf{RFMNN}   & \textbf{AGGLO-2} & \textbf{Onln-GFMM + AGGLO-2} \\ \hline \hline
			1           & Balance\_scale                 & \textbf{2}        & 9                  & 8                          & 7             & 5              & 5                & 5              & \textbf{2}       & \textbf{2}                   \\ \hline
			2           & banknote\_authentication       & 8                 & 9                  & 7                          & 4.5           & \textbf{1.5}   & 4.5              & \textbf{1.5}   & 4.5              & 4.5                          \\ \hline
			3           & blood\_transfusion             & 9                 & 7                  & \textbf{1}                 & 6             & 4              & 3                & 5              & 2                & 8                            \\ \hline
			4           & breast\_cancer\_wisconsin      & 8                 & 6                  & \textbf{1}                 & 5             & 3              & 3                & 3              & 8                & 8                            \\ \hline
			5           & BreastCancerCoimbra            & 4.5               & 4.5                & \textbf{1}                 & 2             & 8              & 8                & 8              & 4.5              & 4.5                          \\ \hline
			6           & connectionist\_bench\_sonar    & 7.5               & 7.5                & 2                          & \textbf{1}    & 4              & 4                & 4              & 7.5              & 7.5                          \\ \hline
			7           & haberman                       & 6                 & 5                  & \textbf{1}                 & 8             & 2              & 3                & 4              & 7                & 9                            \\ \hline
			8           & heart                          & 7                 & 9                  & \textbf{1}                 & 2             & 4              & 4                & 4              & 7                & 7                            \\ \hline
			9           & movement\_libras               & 4.5               & 4.5                & \textbf{1}                 & 9             & 7              & 7                & 7              & 2.5              & 2.5                          \\ \hline
			10          & pima\_diabetes                 & 7                 & 9                  & \textbf{1}                 & 5             & 3.5            & 2                & 3.5            & 6                & 8                            \\ \hline
			11          & plant\_species\_leaves\_margin & 8                 & 6                  & 4                          & 5             & \textbf{2}     & \textbf{2}       & \textbf{2}     & 8                & 8                            \\ \hline
			12          & plant\_species\_leaves\_shape  & 5                 & 4                  & \textbf{1}                 & 7             & 8              & 6                & 9              & 3                & 2                            \\ \hline
			13          & ringnorm                       & 7.5               & 7.5                & \textbf{1}                 & 2             & 3.5            & 9                & 3.5            & 5.5              & 5.5                          \\ \hline
			14          & landsat\_satellite             & 7                 & 8                  & \textbf{1}                 & 9             & 3              & 4                & 2              & 5                & 6                            \\ \hline
			15          & twonorm                        & 7.5               & 7.5                & \textbf{1}                 & 5             & 3              & 3                & 3              & 7.5              & 7.5                          \\ \hline
			16          & vehicle\_silhouettes           & 7                 & 6                  & \textbf{1}                 & 3             & 4.5            & 2                & 4.5            & 9                & 8                            \\ \hline
			17          & vertebral\_column              & 5                 & 6                  & \textbf{1}                 & 9             & 4              & 2                & 3              & 7                & 8                            \\ \hline
			18          & vowel                          & 8.5               & 8.5                & \textbf{1}                 & 2             & 4.5            & 3                & 4.5            & 7                & 6                            \\ \hline
			19          & waveform                       & 6.5               & 6.5                & \textbf{1}                 & 9             & 3              & 3                & 3              & 6.5              & 6.5                          \\ \hline
			20          & wireless\_indoor\_localization & 7                 & 6                  & \textbf{1}                 & 9             & 3.5            & 2                & 3.5            & 8                & 5                            \\ \hline
			\multicolumn{2}{|c|}{\textbf{Average rank}}  & 6.625             & 6.825              & \textbf{1.85}              & 5.475         & 4.05           & 3.975            & 4.15           & 5.875            & 6.175                        \\ \hline
		\end{tabular}
	}
\end{table*}

\begin{table*}[!ht]
	\centering
	\caption{The Average Weighted-F1 Scores and Standard Deviation of the Random Hyperboxes (RH) and Other Fuzzy Min-Max Neural Networks ($\theta=0.7$)} \label{cmp_other_fmnn_07}
	\scriptsize{
		\begin{tabular}{|c|l|R{1cm}|R{1cm}|R{1cm}|R{1cm}|R{1cm}|R{1cm}|R{1cm}|R{1.2cm}|R{1.6cm}|}
			\hline
			\textbf{ID} & \textbf{Dataset}               & \textbf{IOL-GFMM} & \textbf{Onln-GFMM} & \textbf{RH} & \textbf{FMNN}    & \textbf{EFMNN}   & \textbf{KNEFMNN} & \textbf{RFMNN}   & \textbf{AGGLO-2} & \textbf{Onln-GFMM + AGGLO-2} \\ \hline \hline
			1                          &  Balance\_scale                  & 0.59351 $\pm$ 0.0956               & 0.54762 $\pm$ 0.1334                & 0.66276 $\pm$ 0.0915                           & 0.41057 $\pm$ 0.1928           & 0.60536 $\pm$ 0.1306            & 0.61254 \break$\pm$ 0.0863              & \textbf{0.70701 $\pm$ 0.0942} & 0.67825 \break$\pm$ 0.0438                           & 0.67825 \break$\pm$ 0.0438                           \\ \hline
			2                            & banknote\_authentication          & 0.70285 $\pm$ 0.0311              & 0.80833 $\pm$ 0.0184                & 0.95539 $\pm$ 0.0086                           & 0.86049 $\pm$ 0.0423           & 0.76974 $\pm$ 0.0901            & 0.82069 \break$\pm$ 0.0202              & 0.75250 $\pm$ 0.0816                            & \textbf{0.99562 \break$\pm$ 0.0026} & 0.9949 \break$\pm$ 0.0013                            \\ \hline
			3                            & blood\_transfusion                & 0.53269 $\pm$ 0.1765               & 0.46882 $\pm$ 0.1652                & \textbf{0.68779 $\pm$ 0.0292} & 0.55976 $\pm$ 0.0831           & 0.62575 $\pm$ 0.1050            & 0.60291 \break$\pm$ 0.0905              & 0.62544 $\pm$ 0.0675                           & 0.65971 \break$\pm$ 0.0840                           & 0.62368 \break$\pm$ 0.1017                           \\ \hline
			4                            & breast\_cancer\_wisconsin         & 0.95315 $\pm$ 0.0313               & 0.95995 $\pm$ 0.0208                & \textbf{0.96801 $\pm$ 0.0164} & 0.88336 $\pm$ 0.0623           & 0.92016 $\pm$ 0.0505            & 0.94232 \break$\pm$ 0.0166              & 0.95007 $\pm$ 0.0307                           & 0.95118 \break$\pm$ 0.0292                           & 0.95111 \break$\pm$ 0.0316                           \\ \hline
			5                            & BreastCancerCoimbra               & 0.51162 $\pm$ 0.1013               & 0.59101 $\pm$ 0.0692                & \textbf{0.64862 $\pm$ 0.1000} & 0.49436 $\pm$ 0.1081           & 0.57430 $\pm$ 0.0463             & 0.61823 \break$\pm$ 0.0204              & 0.51317 $\pm$ 0.0934                           & 0.54535 \break$\pm$ 0.0868                           & 0.54535 \break$\pm$ 0.0868                           \\ \hline
			6                            & connectionist\_bench\_sonar       & 0.49895 $\pm$ 0.0870               & 0.48042 $\pm$ 0.1324                & \textbf{0.56346 $\pm$ 0.0870} & 0.41787 $\pm$ 0.0573           & 0.51797 $\pm$ 0.0904            & 0.46899 \break$\pm$ 0.1189              & 0.51250 $\pm$ 0.0920                           & 0.48416 \break$\pm$ 0.1419                           & 0.48416 \break$\pm$ 0.1419                           \\ \hline
			7                            & haberman                          & 0.51121 $\pm$ 0.2094               & 0.49672 $\pm$ 0.1993                & 0.59068 $\pm$ 0.1456                           & 0.29147 $\pm$ 0.2110           & 0.58733 $\pm$ 0.0958            & 0.50931 \break$\pm$ 0.1729              & 0.61214 $\pm$ 0.0605                           & 0.62306 \break$\pm$ 0.1450                           & \textbf{0.64127 \break$\pm$ 0.1016} \\ \hline
			8                            & heart                             & 0.77523 $\pm$ 0.0512               & 0.77319 $\pm$ 0.0226                & \textbf{0.81720 $\pm$ 0.0253}  & 0.69788 $\pm$ 0.0263           & 0.78555 $\pm$ 0.0494            & 0.79202 \break$\pm$ 0.0282              & 0.80062 $\pm$ 0.0392                           & 0.76551 \break$\pm$ 0.0317                           & 0.76551 \break$\pm$ 0.0317                           \\ \hline
			9                            & movement\_libras                  & 0.49320 $\pm$ 0.0878                & 0.51492 $\pm$ 0.0921                & \textbf{0.55832 $\pm$ 0.0966} & 0.31186 $\pm$ 0.1207           & 0.45040 $\pm$ 0.1039             & 0.47365 \break$\pm$ 0.1292              & 0.36403 $\pm$ 0.0916                           & 0.55273 \break$\pm$ 0.1417                           & 0.51082 \break$\pm$ 0.1144                           \\ \hline
			10                           & pima\_diabetes                    & 0.68428 $\pm$ 0.0452               & 0.64318 $\pm$ 0.0831                & \textbf{0.74040 $\pm$ 0.0272}  & 0.58610 $\pm$ 0.0589            & 0.63445 $\pm$ 0.0180            & 0.64443 \break$\pm$ 0.0471              & 0.63759 $\pm$ 0.0174                           & 0.69013 \break$\pm$ 0.0226                           & 0.69448 \break$\pm$ 0.0328                           \\ \hline
			11                           & plant\_species\_leaves\_margin    & 0.62682 $\pm$ 0.0101               & 0.63295 $\pm$ 0.0097                & \textbf{0.79507 $\pm$ 0.0251} & 0.78283 $\pm$ 0.0166           & 0.77767 $\pm$ 0.0150            & 0.78105 \break$\pm$ 0.0186              & 0.69359 $\pm$ 0.0199                           & 0.61894 \break$\pm$ 0.0108                           & 0.61894 \break$\pm$ 0.0108                           \\ \hline
			12                           & plant\_species\_leaves\_shape     & 0.52069 $\pm$ 0.0166               & 0.45639 $\pm$ 0.0259                & \textbf{0.61318 $\pm$ 0.0251} & 0.43395 $\pm$ 0.0382           & 0.43569 \break$\pm$ 0.0355            & 0.43625 \break$\pm$ 0.0355              & 0.44809 $\pm$ 0.0283                           & 0.56043 \break$\pm$ 0.0257                           & 0.55659 \break$\pm$ 0.0211                           \\ \hline
			13                           & ringnorm                          & 0.65641 $\pm$ 0.1616               & 0.77946 $\pm$ 0.0323                & 0.86803 $\pm$ 0.0250                           & 0.82055 $\pm$ 0.0090           & 0.65747 $\pm$ 0.0431            & 0.67500 \break$\pm$ 0.0705              & 0.66250 $\pm$ 0.0818                            & \textbf{0.87453 \break$\pm$ 0.0018} & \textbf{0.87453 \break$\pm$ 0.0018} \\ \hline
			14                           & landsat\_satellite                & 0.83192 $\pm$ 0.0269               & 0.58411 $\pm$ 0.0145                & 0.86226 $\pm$ 0.0165                           & 0.52566 $\pm$ 0.1310          & 0.54733 $\pm$ 0.1926            & 0.66287 \break$\pm$ 0.0617              & 0.75732 $\pm$ 0.0439                           & 0.87039 \break$\pm$ 0.0090                           & \textbf{0.87299 \break$\pm$ 0.0148} \\ \hline
			15                           & twonorm                           & 0.91658 $\pm$ 0.0088               & 0.76649 $\pm$ 0.0176                & \textbf{0.97122 $\pm$ 0.0029} & 0.80136 $\pm$ 0.0387           & 0.79747 $\pm$ 0.0167            & 0.73673 \break$\pm$ 0.0057              & 0.69279 $\pm$ 0.1192                           & 0.95973 \break$\pm$ 0.0016                           & 0.95973 \break$\pm$ 0.0016                           \\ \hline
			16                           & vehicle\_silhouettes              & 0.64533 $\pm$ 0.0184               & 0.48170 $\pm$ 0.0668                 & \textbf{0.70332 $\pm$ 0.0255} & 0.28571 $\pm$ 0.0317           & 0.48748 $\pm$ 0.0196            & 0.49573 \break$\pm$ 0.0408              & 0.55892 $\pm$ 0.0488                           & 0.65988 \break$\pm$ 0.0328                           & 0.66057 \break$\pm$ 0.0266                           \\ \hline
			17                           & vertebral\_column                 & 0.61791 $\pm$ 0.0341               & 0.74575 $\pm$ 0.0193                & \textbf{0.77698 $\pm$ 0.0396} & 0.74546 $\pm$ 0.0223           & 0.73233 $\pm$ 0.0256            & 0.75136 \break$\pm$ 0.0213              & 0.74130 $\pm$ 0.0138                           & 0.73627 \break$\pm$ 0.0705                           & 0.75678 \break$\pm$ 0.0547                           \\ \hline
			18                           & vowel                             & 0.53239 $\pm$ 0.0370               & 0.43198 $\pm$ 0.0434                & \textbf{0.58876 $\pm$ 0.0698} & 0.33920 $\pm$ 0.0476            & 0.40335 $\pm$ 0.0485            & 0.40713 \break$\pm$ 0.0409              & 0.46578 $\pm$ 0.0521                           & 0.52614 \break$\pm$ 0.0633                           & 0.53066 \break$\pm$ 0.0401                           \\ \hline
			19                           & waveform                          & 0.79168 $\pm$ 0.0150               & 0.76606 $\pm$ 0.0090                & \textbf{0.83201 $\pm$ 0.0098} & 0.70826 $\pm$ 0.0275           & 0.73494 $\pm$ 0.0188            & 0.75625 \break$\pm$ 0.0151              & 0.51829 $\pm$ 0.0275                           & 0.81138 \break$\pm$ 0.0126                           & 0.81138 \break$\pm$ 0.0126                           \\ \hline
			20                           & wireless\_indoor\_localization    & 0.85201 $\pm$ 0.0450               & 0.93139 $\pm$ 0.0240                & \textbf{0.97437 $\pm$ 0.0093} & 0.92680 $\pm$ 0.0280            & 0.86827 $\pm$ 0.0470            & 0.86771 \break$\pm$ 0.0484              & 0.84572 $\pm$ 0.0483                           & 0.97389 \break$\pm$ 0.0053                           & 0.97398 \break$\pm$ 0.0042                           \\ \hline
			
		\end{tabular}
	}
\end{table*}

\begin{table*}[!ht]
	\centering
	\caption{The Ranking of the Random Hyperboxes and Other Fuzzy Min-Max Classfiers ($\theta=0.7$)} \label{cmp_other_fmnn_07_ranking}
	\scriptsize{
		\begin{tabular}{|c|l|c|c|c|c|c|c|c|c|C{1.5cm}|}
			\hline
			\textbf{ID} & \textbf{Dataset}               & \textbf{IOL-GFMM} & \textbf{Onln-GFMM} & \textbf{RH} & \textbf{FMNN}    & \textbf{EFMNN}   & \textbf{KNEFMNN} & \textbf{RFMNN}   & \textbf{AGGLO-2} & \textbf{Onln-GFMM + AGGLO-2} \\ \hline \hline
			1           & Balance\_scale                 & 7                 & 8                  & 4                          & 9             & 6              & 5                & \textbf{1}     & 2.5              & 2.5                          \\ \hline
			2           & banknote\_authentication       & 9                 & 6                  & 3                          & 4             & 7              & 5                & 8              & \textbf{1}       & 2                            \\ \hline
			3           & blood\_transfusion             & 8                 & 9                  & \textbf{1}                 & 7             & 3              & 6                & 4              & 2                & 5                            \\ \hline
			4           & breast\_cancer\_wisconsin      & 3                 & 2                  & \textbf{1}                 & 9             & 8              & 7                & 6              & 4                & 5                            \\ \hline
			5           & BreastCancerCoimbra            & 8                 & 3                  & \textbf{1}                 & 9             & 4              & 2                & 7              & 5.5              & 5.5                          \\ \hline
			6           & connectionist\_bench\_sonar    & 4                 & 7                  & \textbf{1}                 & 9             & 2              & 8                & 3              & 5.5              & 5.5                          \\ \hline
			7           & haberman                       & 7                 & 8                  & 4                          & 9             & 5              & 6                & 3              & 2                & \textbf{1}                   \\ \hline
			8           & heart                          & 5                 & 6                  & \textbf{1}                 & 9             & 4              & 3                & 2              & 7.5              & 7.5                          \\ \hline
			9           & movement\_libras               & 5                 & 3                  & \textbf{1}                 & 9             & 7              & 6                & 8              & 2                & 4                            \\ \hline
			10          & pima\_diabetes                 & 4                 & 6                  & \textbf{1}                 & 9             & 8              & 5                & 7              & 3                & 2                            \\ \hline
			11          & plant\_species\_leaves\_margin & 7                 & 6                  & \textbf{1}                 & 2             & 4              & 3                & 5              & 8.5              & 8.5                          \\ \hline
			12          & plant\_species\_leaves\_shape  & 4                 & 5                  & \textbf{1}                 & 9             & 8              & 7                & 6              & 2                & 3                            \\ \hline
			13          & ringnorm                       & 9                 & 5                  & 3                          & 4             & 8              & 6                & 7              & \textbf{1.5}     & \textbf{1.5}                 \\ \hline
			14          & landsat\_satellite             & 4                 & 7                  & 3                          & 9             & 8              & 6                & 5              & 2                & \textbf{1}                   \\ \hline
			15          & twonorm                        & 4                 & 7                  & \textbf{1}                 & 5             & 6              & 8                & 9              & 2.5              & 2.5                          \\ \hline
			16          & vehicle\_silhouettes           & 4                 & 8                  & \textbf{1}                 & 9             & 7              & 6                & 5              & 3                & 2                            \\ \hline
			17          & vertebral\_column              & 9                 & 4                  & \textbf{1}                 & 5             & 8              & 3                & 6              & 7                & 2                            \\ \hline
			18          & vowel                          & 2                 & 6                  & \textbf{1}                 & 9             & 8              & 7                & 5              & 4                & 3                            \\ \hline
			19          & waveform                       & 4                 & 5                  & \textbf{1}                 & 8             & 7              & 6                & 9              & 2.5              & 2.5                          \\ \hline
			20          & wireless\_indoor\_localization & 8                 & 4                  & \textbf{1}                 & 5             & 6              & 7                & 9              & 3                & 2                            \\ \hline
			\multicolumn{2}{|c|}{\textbf{Average rank}}  & 5.75              & 5.75               & \textbf{1.6}               & 7.4           & 6.2            & 5.6              & 5.75           & 3.55             & 3.4                          \\ \hline
		\end{tabular}
	}
\end{table*}

\subsubsection{A Comparison of the Random Hyperboxes With Other Ensemble Classifiers} \label{sup_cmp_other_ensemble} \hfill

This subsection presents the experimental results of the random hyperboxes and other popular ensemble classifiers mentioned in subsection IV.B.2 from the main paper. The base learners of the random hyperboxes model used the threshold $\theta = 0.1$ for the maximum hyperbox size.

Table \ref{cmp_other_ensemble} shows the average weighted-F1 scores of classifiers through 40 iterations with different testing folds (10 times repeated 4-fold cross-validation). The ranks of these classifiers are shown in Table \ref{cmp_other_ensemble_ranking}.

\begin{table*}[!ht]
	\centering
	\caption{The Average Weighted-F1 Scores and Standard Deviation of the Random Hyperbox Model and Other Ensemble Models} \label{cmp_other_ensemble}
	\scriptsize{
		\begin{tabular}{|c|l|R{1cm}|R{1cm}|R{1cm}|R{1cm}|R{1cm}|R{1cm}|R{1.9cm}|R{1.9cm}|}
			\hline
			\textbf{ID} & \textbf{Dataset}               & \textbf{RH} & \textbf{Random Forest} & \textbf{Rotation Forest} & \textbf{XGBoost} & \textbf{LightGBM} & \textbf{Gradient Boosting} & \textbf{Ens-IOL-GFMM (DL)} & \textbf{Ens-IOL-GFMM (ML)} \\ \hline \hline
			1                            & Balance\_scale                    & 0.73821 $\pm$ 0.0894          & 0.59325 $\pm$ 0.1943                    & 0.71405 $\pm$ 0.1039                      & 0.69423 $\pm$ 0.1589              & 0.59721 $\pm$ 0.2134               & 0.58660 $\pm$ 0.1947                        & \textbf{0.79987 \break$\pm$ 0.0476}               & 0.79383 \break$\pm$ 0.0558                        \\ \hline
			2                            & banknote\_authentication          & 0.99818 $\pm$ 0.0018          & 0.99054 $\pm$ 0.0043                    & 0.99052 $\pm$ 0.0053                      & 0.99563 $\pm$ 0.0025              & 0.99636 $\pm$ 0.0032               & 0.99709 $\pm$ 0.0021                        & 0.9984 \break$\pm$ 0.0017                         & \textbf{0.99869 \break$\pm$ 0.0015}               \\ \hline
			3                            & blood\_transfusion                & 0.68264 $\pm$ 0.0190          & 0.66193 $\pm$ 0.0331                    & 0.63456 $\pm$ 0.0630                      & 0.68826 $\pm$ 0.0368              & 0.66894 $\pm$ 0.0418               & 0.64810 $\pm$ 0.0448                        & \textbf{0.68847 \break$\pm$ 0.0268}               & 0.65257 \break$\pm$ 0.0706                        \\ \hline
			4                            & breast\_cancer\_wisconsin         & \textbf{0.96685 $\pm$ 0.0180} & 0.95570 $\pm$ 0.0253                     & 0.95426 $\pm$ 0.0140                      & 0.96139 $\pm$ 0.0174              & 0.94973 $\pm$ 0.0181               & 0.95848 $\pm$ 0.0214                        & 0.94721 \break$\pm$ 0.0332                        & 0.94383 \break$\pm$ 0.0329                        \\ \hline
			5                            & BreastCancerCoimbra               & 0.66561 $\pm$ 0.0939          & 0.66808 $\pm$ 0.0917                    & 0.61706 $\pm$ 0.0808                      & 0.70412 $\pm$ 0.1285              & \textbf{0.74284 $\pm$ 0.0879}      & 0.65378 $\pm$ 0.1094                        & 0.62096 \break$\pm$ 0.0631                        & 0.62798 \break$\pm$ 0.0381                        \\ \hline
			6                            & connectionist\_bench\_sonar       & 0.55964 $\pm$ 0.1085          & 0.55992 $\pm$ 0.0453                    & 0.60441 $\pm$ 0.0425                      & \textbf{0.63616 $\pm$ 0.0802}     & 0.61462 $\pm$ 0.0408               & 0.55653 $\pm$ 0.0603                        & 0.52737 \break$\pm$ 0.1244                        & 0.53013 \break$\pm$ 0.0816                        \\ \hline
			7                            & haberman                          & 0.64211 $\pm$ 0.0294          & 0.55050 $\pm$ 0.1330                     & 0.53359 $\pm$ 0.0789                      & 0.63966 $\pm$ 0.0488              & 0.65125 $\pm$ 0.0478               & 0.63137 $\pm$ 0.0214                        & \textbf{0.65373 \break$\pm$ 0.0428}               & 0.59270 \break$\pm$ 0.0865                        \\ \hline
			8                            & heart                             & \textbf{0.82643 $\pm$ 0.0252} & 0.80593 $\pm$ 0.0534                    & 0.81462 $\pm$ 0.0298                      & 0.81757 $\pm$ 0.0430              & 0.79135 $\pm$ 0.0463               & 0.80947 $\pm$ 0.0393                        & 0.75458 \break$\pm$ 0.0185                        & 0.76101 \break$\pm$ 0.0272                        \\ \hline
			9                            & movement\_libras                  & \textbf{0.54465 $\pm$ 0.1064} & 0.48745 $\pm$ 0.1456                    & 0.47701 $\pm$ 0.1957                      & 0.42539 $\pm$ 0.1558              & 0.46981 $\pm$ 0.1835               & 0.39863 $\pm$ 0.1370                        & 0.5254 \break$\pm$ 0.0858                         & 0.52819 \break$\pm$ 0.0818                        \\ \hline
			10                           & pima\_diabetes                    & 0.72760 $\pm$ 0.0339          & 0.76566 $\pm$ 0.0455                    & \textbf{0.77200 $\pm$ 0.0474}             & 0.75014 $\pm$ 0.0364              & 0.72647 $\pm$ 0.0228               & 0.75678 $\pm$ 0.0357                        & 0.72359 \break$\pm$ 0.0230                        & 0.70680 \break$\pm$ 0.0148                        \\ \hline
			11                           & plant\_species\_leaves\_margin    & 0.74748 $\pm$ 0.0195          & 0.72687 $\pm$ 0.0246                    & 0.66113 $\pm$ 0.0356                      & \textbf{0.78722 $\pm$ 0.0096}     & 0.78549 $\pm$ 0.0139               & 0.32808 $\pm$ 0.0174                        & 0.59427 \break$\pm$ 0.0167                        & 0.57408 \break$\pm$ 0.0217                        \\ \hline
			12                           & plant\_species\_leaves\_shape     & \textbf{0.60695 $\pm$ 0.0306} & 0.55444 $\pm$ 0.0146                    & 0.52707 $\pm$ 0.0139                      & 0.55204 $\pm$ 0.0201              & 0.50336 $\pm$ 0.0148               & 0.31396 $\pm$ 0.0311                        & 0.56949 \break$\pm$ 0.0293                        & 0.58314 \break$\pm$ 0.0103                        \\ \hline
			13                           & ringnorm                          & 0.94237 $\pm$ 0.0057          & 0.94594 $\pm$ 0.0063                    & 0.92605 $\pm$ 0.0021                      & 0.97851 $\pm$ 0.0030              & \textbf{0.98094 $\pm$ 0.0008}      & 0.97810 $\pm$ 0.0041                        & 0.58982 \break$\pm$ 0.0059                        & 0.59580 \break$\pm$ 0.0058                        \\ \hline
			14                           & landsat\_satellite                & 0.87875 $\pm$ 0.0076          & 0.88374 $\pm$ 0.0082                    & 0.88817 $\pm$ 0.0092                      & 0.89463 $\pm$ 0.0060              & \textbf{0.89942 $\pm$ 0.0059}      & 0.89217 $\pm$ 0.0059                        & 0.86874 \break$\pm$ 0.0110                        & 0.86412 \break$\pm$ 0.0165                        \\ \hline
			15                           & twonorm                           & 0.97211 $\pm$ 0.0032          & 0.97081 $\pm$ 0.0009                    & 0.96554 $\pm$ 0.0033                      & 0.97230 $\pm$ 0.0019               & 0.97311 $\pm$ 0.0016               & \textbf{0.97365 $\pm$ 0.0010}               & 0.96477 \break$\pm$ 0.0039                        & 0.94284 \break$\pm$ 0.0054                        \\ \hline
			16                           & vehicle\_silhouettes              & 0.72044 $\pm$ 0.0309          & 0.74634 $\pm$ 0.0251                    & 0.73402 $\pm$ 0.0173                      & 0.76474 $\pm$ 0.0136              & \textbf{0.76918 $\pm$ 0.0086}      & 0.76615 $\pm$ 0.0064                        & 0.66757 \break$\pm$ 0.0214                        & 0.65051 \break$\pm$ 0.0206                        \\ \hline
			17                           & vertebral\_column                 & 0.76542 $\pm$ 0.0411          & \textbf{0.85617 $\pm$ 0.0329}           & 0.74431 $\pm$ 0.0451                      & 0.80543 $\pm$ 0.0398              & 0.82058 $\pm$ 0.0497               & 0.81368 $\pm$ 0.0391                        & 0.73812 \break$\pm$ 0.0572                        & 0.76917 \break$\pm$ 0.0320                        \\ \hline
			18                           & vowel                             & \textbf{0.63059 $\pm$ 0.0676} & 0.59992 $\pm$ 0.0470                    & 0.57678 $\pm$ 0.1026                      & 0.60286 $\pm$ 0.0685              & 0.60048 $\pm$ 0.0465               & 0.60462 $\pm$ 0.0529                        & 0.54363 \break$\pm$ 0.0684                        & 0.56073 \break$\pm$ 0.0485                        \\ \hline
			19                           & waveform                          & 0.83795 $\pm$ 0.0068          & 0.84593 $\pm$ 0.0120                    & 0.85010 $\pm$ 0.0093                      & \textbf{0.85878 $\pm$ 0.0065}     & 0.85587 $\pm$ 0.0037               & 0.85855 $\pm$ 0.0086                        & 0.81709 \break$\pm$ 0.0082                        & 0.78002 \break$\pm$ 0.0107                        \\ \hline
			20                           & wireless\_indoor\_localization    & \textbf{0.97726 $\pm$ 0.0086} & 0.97348 $\pm$ 0.0159                    & 0.95229 $\pm$ 0.0112                      & 0.97599 $\pm$ 0.0112              & 0.97554 $\pm$ 0.0088               & 0.97344 $\pm$ 0.0185                        & 0.97566 \break$\pm$ 0.0037                        & 0.97112 \break$\pm$ 0.0058                        \\ \hline
		\end{tabular}
	}
\end{table*}

\begin{table}[!ht]
	\centering
	\caption{The Ranking of the Random Hyperbox Model and Other Ensemble Models} \label{cmp_other_ensemble_ranking}
	\scriptsize{
		\begin{tabular}{|c|l|c|C{1cm}|C{1cm}|c|c|C{1.3cm}|C{2cm}|C{2cm}|}
			\hline
			\textbf{ID} & \textbf{Dataset}               & \textbf{RH} & \textbf{Random Forest} & \textbf{Rotation Forest} & \textbf{XGBoost} & \textbf{LightGBM} & \textbf{Gradient Boosting} & \textbf{Ens-IOL-GFMM (DL)} & \textbf{Ens-IOL-GFMM (ML)} \\ \hline \hline
			1           & Balance\_scale                 & 3                          & 7                      & 4                        & 5                & 6                 & 8                          & \textbf{1}                 & 2                          \\ \hline
			2           & banknote\_authentication       & 3                          & 7                      & 8                        & 6                & 5                 & 4                          & 2                          & \textbf{1}                 \\ \hline
			3           & blood\_transfusion             & 3                          & 5                      & 8                        & 2                & 4                 & 7                          & \textbf{1}                 & 6                          \\ \hline
			4           & breast\_cancer\_wisconsin      & \textbf{1}                 & 4                      & 5                        & 2                & 6                 & 3                          & 7                          & 8                          \\ \hline
			5           & BreastCancerCoimbra            & 4                          & 3                      & 8                        & 2                & \textbf{1}        & 5                          & 7                          & 6                          \\ \hline
			6           & connectionist\_bench\_sonar    & 5                          & 4                      & 3                        & \textbf{1}       & 2                 & 6                          & 8                          & 7                          \\ \hline
			7           & haberman                       & 3                          & 7                      & 8                        & 4                & 2                 & 5                          & \textbf{1}                 & 6                          \\ \hline
			8           & heart                          & \textbf{1}                 & 5                      & 3                        & 2                & 6                 & 4                          & 8                          & 7                          \\ \hline
			9           & movement\_libras               & \textbf{1}                 & 4                      & 5                        & 7                & 6                 & 8                          & 3                          & 2                          \\ \hline
			10          & pima\_diabetes                 & 5                          & 2                      & \textbf{1}               & 4                & 6                 & 3                          & 7                          & 8                          \\ \hline
			11          & plant\_species\_leaves\_margin & 3                          & 4                      & 5                        & \textbf{1}       & 2                 & 8                          & 6                          & 7                          \\ \hline
			12          & plant\_species\_leaves\_shape  & \textbf{1}                 & 4                      & 6                        & 5                & 7                 & 8                          & 3                          & 2                          \\ \hline
			13          & ringnorm                       & 5                          & 4                      & 6                        & 2                & \textbf{1}        & 3                          & 8                          & 7                          \\ \hline
			14          & landsat\_satellite             & 6                          & 5                      & 4                        & 2                & \textbf{1}        & 3                          & 7                          & 8                          \\ \hline
			15          & twonorm                        & 4                          & 5                      & 6                        & 3                & 2                 & \textbf{1}                 & 7                          & 8                          \\ \hline
			16          & vehicle\_silhouettes           & 6                          & 4                      & 5                        & 3                & \textbf{1}        & 2                          & 7                          & 8                          \\ \hline
			17          & vertebral\_column              & 6                          & \textbf{1}             & 7                        & 4                & 2                 & 3                          & 8                          & 5                          \\ \hline
			18          & vowel                          & \textbf{1}                 & 5                      & 6                        & 3                & 4                 & 2                          & 8                          & 7                          \\ \hline
			19          & waveform                       & 6                          & 5                      & 4                        & \textbf{1}       & 3                 & 2                          & 7                          & 8                          \\ \hline
			20          & wireless\_indoor\_localization & \textbf{1}                 & 5                      & 8                        & 2                & 4                 & 6                          & 3                          & 7                          \\ \hline
			\multicolumn{2}{|c|}{\textbf{Average rank}}  & 3.4                        & 4.5                    & 5.5                      & \textbf{3.05}    & 3.55              & 4.55                       & 5.45                       & 6                          \\ \hline
		\end{tabular}
	}
\end{table}

It can be observed that the average performance of the random hyperboxes is much better than the results of Random Forest, Rotation Forest, Gradient Boosting, and the ensemble models of IOL-GFMM learners with full features. It is also slightly better than LightGBM, but the random hyperboxes classifier cannot outperform the XGBoost model on 20 considered datasets. In spite of using the same base learners and sampling method, the random hyperboxes classifier is much better than the Ens-IOL-GFMM with decision and model combination levels. It is due to the fact that the random hyperboxes classifier uses only a subset of features to train each base learner. This method reduces the correlation between base learners, and so it leads to the reduction of generalization errors. These empirical results are consistent with the theoretical results presented in the main paper. However, it is also noted that the correlation is linked with variance, so achieving a low correlation but high variance will not decrease the prediction error. In addition, when reducing correlation by using a smaller number of features, it will also increase the variance of each base learner. Therefore, to achieve the reduction of prediction error, the correlation between base learners has to decrease faster than the growth of the variance. This issue needs to be analyzed in more details in the future study, especially the relationship between the maximum number of used features and the number of base learners.

\subsubsection{A Comparison of the Random Hyperboxes With Other Machine Learning Algorithms} \label{sup_cmp_other_ml} \hfill

This part presents the empirical results of the random hyperboxes classifier and other popular machine learning algorithms shown in subsection IV.B.3 from the main paper.

Table \ref{cmp_other_ml} shows the average weighted-F1 scores of the random hyperboxes and other classifiers for the 20 datasets. The ranks of these models are presented in Table \ref{cmp_other_ml_rank}. In this experiment, the random hyperboxes achieved the best performance among considered classifiers.

\begin{table*}[!ht]
	\centering
	\caption{The Average Weighted-F1 Scores and Standard Deviation of the Random Hyperboxes and Other Machine Learning Algorithms} \label{cmp_other_ml}
	\scriptsize{
		\begin{tabular}{|c|l|C{2cm}|C{2cm}|C{2cm}|C{2cm}|C{2cm}|C{2cm}|}
			\hline
			\textbf{ID} & \multicolumn{1}{c|}{\textbf{Dataset}} & \textbf{Random hyperboxes} & \textbf{Decision tree} & \textbf{SVM}     & \textbf{KNN}     & \textbf{Naive Bayes} & \textbf{LDA}     \\ \hline \hline
			1                            & Balance\_scale                 & 0.73821 $\pm$ 0.0894                        & 0.57709 $\pm$ 0.1822                    & \textbf{0.80213 $\pm$ 0.0505} & 0.74637 $\pm$ 0.0566          & 0.64495 $\pm$ 0.1950                  & 0.76648 $\pm$ 0.0743          \\ \hline
			2                            & banknote\_authentication       & 0.99818 $\pm$ 0.0018                        & 0.98471 $\pm$ 0.0052                    & 0.97891 $\pm$ 0.0082          & \textbf{0.99854 $\pm$ 0.0015} & 0.83836 $\pm$ 0.0232                  & 0.97674 $\pm$ 0.0080          \\ \hline
			3                            & blood\_transfusion             & 0.68264 $\pm$ 0.0190                        & 0.64101 $\pm$ 0.0670                    & 0.65913 $\pm$ 0.0019          & 0.64196 $\pm$ 0.1086          & 0.70667 $\pm$ 0.0386                  & \textbf{0.72411 $\pm$ 0.0748} \\ \hline
			4                            & breast\_cancer\_wisconsin      & \textbf{0.96685 $\pm$ 0.0180}               & 0.92186 $\pm$ 0.0306                    & 0.95989 $\pm$ 0.0267          & 0.96287 $\pm$ 0.0222          & 0.96035 $\pm$ 0.0154                  & 0.95521 $\pm$ 0.0356          \\ \hline
			5                            & BreastCancerCoimbra            & 0.66561 $\pm$ 0.0939                        & 0.63368 $\pm$ 0.1102                    & 0.38407 $\pm$ 0.0145          & \textbf{0.68939 $\pm$ 0.1169} & 0.56916 $\pm$ 0.1305                  & 0.58265 $\pm$ 0.1505          \\ \hline
			6                            & connectionist\_bench\_sonar    & 0.55964 $\pm$ 0.1085                        & 0.53636 $\pm$ 0.0475                    & \textbf{0.56390 $\pm$ 0.1128} & 0.47834 $\pm$ 0.1083          & 0.53318 $\pm$ 0.1559                  & 0.55007 $\pm$ 0.0890          \\ \hline
			7                            & haberman                       & 0.64211 $\pm$ 0.0294                        & 0.56815 $\pm$ 0.0639                    & 0.62317 $\pm$ 0.0036          & 0.68248 $\pm$ 0.0334          & 0.69737 $\pm$ 0.0264                  & \textbf{0.70811 $\pm$ 0.0171} \\ \hline
			8                            & heart                          & 0.82643 $\pm$ 0.0252                        & 0.76268 $\pm$ 0.0588                    & 0.83253 $\pm$ 0.0339          & 0.80648 $\pm$ 0.0171          & \textbf{0.84803 $\pm$ 0.0122}         & 0.83265 $\pm$ 0.0335          \\ \hline
			9                            & movement\_libras               & \textbf{0.54465 $\pm$ 0.1064}               & 0.34075 $\pm$ 0.1197                    & 0.41919 $\pm$ 0.1311          & 0.50787 $\pm$ 0.1029          & 0.42762 $\pm$ 0.1902                  & 0.50048 $\pm$ 0.0515          \\ \hline
			10                           & pima\_diabetes                 & 0.72760 $\pm$ 0.0339                        & 0.70859 $\pm$ 0.0347                    & 0.73475 $\pm$ 0.0259          & 0.74955 $\pm$ 0.0224          & 0.74796 $\pm$ 0.0232                  & \textbf{0.76027 $\pm$ 0.0327} \\ \hline
			11                           & plant\_species\_leaves\_margin & 0.74748 $\pm$ 0.0195                        & 0.17446 $\pm$ 0.0127                    & 0.72261 $\pm$ 0.0118          & 0.74603 $\pm$ 0.0092          & 0.71911 $\pm$ 0.0191                  & \textbf{0.78220 $\pm$ 0.0189} \\ \hline
			12                           & plant\_species\_leaves\_shape  & \textbf{0.60695 $\pm$ 0.0306}               & 0.32775 $\pm$ 0.0372                    & 0.40432 $\pm$ 0.0226          & 0.54375 $\pm$ 0.0349          & 0.51460 $\pm$ 0.0173                  & 0.47546 $\pm$ 0.0369          \\ \hline
			13                           & ringnorm                       & 0.94237 $\pm$ 0.0057                        & 0.86909 $\pm$ 0.0040                    & 0.83666 $\pm$ 0.0038          & 0.66874 $\pm$ 0.0097          & \textbf{0.98662 $\pm$ 0.0016}         & 0.77029 $\pm$ 0.0015          \\ \hline
			14                           & landsat\_satellite             & 0.87875 $\pm$ 0.0076                        & 0.83136 $\pm$ 0.0183                    & 0.82357 $\pm$ 0.0107          & \textbf{0.88190 $\pm$ 0.0130} & 0.79565 $\pm$ 0.0370                  & 0.82012 $\pm$ 0.0116          \\ \hline
			15                           & twonorm                        & 0.97211 $\pm$ 0.0032                        & 0.84553 $\pm$ 0.0066                    & 0.97824 $\pm$ 0.0017          & 0.97284 $\pm$ 0.0022          & \textbf{0.97892 $\pm$ 0.0020}         & 0.97838 $\pm$ 0.0025          \\ \hline
			16                           & vehicle\_silhouettes           & 0.72044 $\pm$ 0.0309                        & 0.71707 $\pm$ 0.0262                    & 0.55849 $\pm$ 0.0273          & 0.69223 $\pm$ 0.0067          & 0.44138 $\pm$ 0.0210                  & \textbf{0.78081 $\pm$ 0.0304} \\ \hline
			17                           & vertebral\_column              & 0.76542 $\pm$ 0.0411                        & 0.81931 $\pm$ 0.0454                    & 0.61393 $\pm$ 0.0375          & 0.73993 $\pm$ 0.0351          & \textbf{0.82177 $\pm$ 0.0346}         & 0.80953 $\pm$ 0.0673          \\ \hline
			18                           & vowel                          & \textbf{0.63059 $\pm$ 0.0676}               & 0.43943 $\pm$ 0.0552                    & 0.36361 $\pm$ 0.0883          & 0.57885 $\pm$ 0.0586          & 0.52917 $\pm$ 0.0814                  & 0.46100 $\pm$ 0.0763          \\ \hline
			19                           & waveform                       & 0.83795 $\pm$ 0.0068                        & 0.76834 $\pm$ 0.0049                    & \textbf{0.86992 $\pm$ 0.0025} & 0.84700 $\pm$ 0.0080          & 0.79820 $\pm$ 0.0045                  & 0.86255 $\pm$ 0.0074          \\ \hline
			20                           & wireless\_indoor\_localization & 0.97726 $\pm$ 0.0086                        & 0.96088 $\pm$ 0.0175                    & 0.97405 $\pm$ 0.0052          & 0.97556 $\pm$ 0.0040          & \textbf{0.98055 $\pm$ 0.0096}         & 0.97074 $\pm$ 0.0062          \\ \hline
		\end{tabular}
	}
\end{table*}

\begin{table*}[!ht]
	\centering
	\caption{The Ranking of the Random Hyperboxes and Other Machine Learning Algorithms} \label{cmp_other_ml_rank}
	\scriptsize{
		\begin{tabular}{|c|l|c|c|c|c|c|c|}
			\hline
			\textbf{ID} & \multicolumn{1}{c|}{\textbf{Dataset}} & \textbf{Random hyperboxes} & \textbf{Decision tree} & \textbf{SVM}     & \textbf{KNN}     & \textbf{Naive Bayes} & \textbf{LDA}     \\ \hline \hline
			1           & Balance\_scale                        & 4                          & 6                      & \textbf{1}   & 3             & 5                    & 2            \\ \hline
			2           & banknote\_authentication              & 2                          & 3                      & 4            & \textbf{1}    & 6                    & 5            \\ \hline
			3           & blood\_transfusion                    & 3                          & 6                      & 4            & 5             & 2                    & \textbf{1}   \\ \hline
			4           & breast\_cancer\_wisconsin             & \textbf{1}                 & 6                      & 4            & 2             & 3                    & 5            \\ \hline
			5           & BreastCancerCoimbra                   & 2                          & 3                      & 6            & \textbf{1}    & 5                    & 4            \\ \hline
			6           & connectionist\_bench\_sonar           & 2                          & 4                      & \textbf{1}   & 6             & 5                    & 3            \\ \hline
			7           & haberman                              & 4                          & 6                      & 5            & 3             & 2                    & \textbf{1}   \\ \hline
			8           & heart                                 & 4                          & 6                      & 3            & 5             & \textbf{1}           & 2            \\ \hline
			9           & movement\_libras                      & \textbf{1}                 & 6                      & 5            & 2             & 4                    & 3            \\ \hline
			10          & pima\_diabetes                        & 5                          & 6                      & 4            & 2             & 3                    & \textbf{1}   \\ \hline
			11          & plant\_species\_leaves\_margin        & 2                          & 6                      & 4            & 3             & 5                    & \textbf{1}   \\ \hline
			12          & plant\_species\_leaves\_shape         & \textbf{1}                 & 6                      & 5            & 2             & 3                    & 4            \\ \hline
			13          & ringnorm                              & 2                          & 3                      & 4            & 6             & \textbf{1}           & 5            \\ \hline
			14          & landsat\_satellite                    & 2                          & 3                      & 4            & \textbf{1}    & 6                    & 5            \\ \hline
			15          & twonorm                               & 5                          & 6                      & 3            & 4             & \textbf{1}           & 2            \\ \hline
			16          & vehicle\_silhouettes                  & 2                          & 3                      & 5            & 4             & 6                    & \textbf{1}   \\ \hline
			17          & vertebral\_column                     & 4                          & 2                      & 6            & 5             & \textbf{1}           & 3            \\ \hline
			18          & vowel                                 & \textbf{1}                 & 5                      & 6            & 2             & 3                    & 4            \\ \hline
			19          & waveform                              & 4                          & 6                      & \textbf{1}   & 3             & 5                    & 2            \\ \hline
			20          & wireless\_indoor\_localization        & 2                          & 6                      & 4            & 3             & \textbf{1}           & 5            \\ \hline
			\multicolumn{2}{|c|}{\textbf{Average rank}}         & \textbf{2.65}                       & 4.9                    & 3.95         & 3.15 & 3.4                  & 2.95         \\ \hline
		\end{tabular}
	}
\end{table*}



\twocolumn
\bibliographystyle{IEEEtran}
\bibliography{reference}
%

%








\end{document}